\documentclass[10pt,twocolumn,letterpaper]{article}
\usepackage[pagenumbers]{cvpr}
\usepackage[utf8]{inputenc}
\usepackage[T1]{fontenc}
\usepackage{amsmath,amssymb,amsfonts,mathtools}
\usepackage{graphicx,booktabs,tabularx,array,adjustbox}
\usepackage{microtype,placeins,needspace,fvextra}
\definecolor{cvprblue}{rgb}{0.21,0.49,0.74}
\usepackage[pagebackref,breaklinks,colorlinks,allcolors=cvprblue]{hyperref}
\hypersetup{pdftitle={CHARM: Certified Homology-Aware Reverse-scale Mapping}}
\def\confName{CVPR}
\def\confYear{2027}

\graphicspath{{figures/}}
\title{CHARM: Certified Homology-Aware Reverse-scale Mapping via Diffusion-Guided Quotient Towers}
\author{Zhe Li \\
  Peking Union Medical College Hospital \\
  Beijing, China \\
  \texttt{dzztlz@163.com}
}

\begin{document}

\maketitle
\begin{abstract}
Conventional dimensionality-reduction methods optimize geometric fidelity but do not return auditable homology certificates or identify the input observations associated with nontrivial classes. 
Direct high-dimensional persistent-homology computation is costly, 
and its representatives are rarely linked to a shared planar view. We introduce Certified Homology-Aware Reverse-scale Mapping (CHARM), 
which augments two-dimensional coordinates with algebraically certified, 
sample-traceable structural records. Diffusion-guided cover proposals and 
homotopy-safe contractions compress the input neighborhood flag complex. 
A Cover Quotient Tower supports cross-scale representative tracking and 
exact $H_k$ certification when algebraic tests complete. 
Reverse-scale layout connects quotient-chain preimages to the same coordinates used for geometric exploration. 
Contraction preserves the input neighborhood flag-complex homotopy type; 
the quotient tower inherits a multiplicative 3-interleaving with coarse-hop Rips, and conditional distortion bounds connect it to original Euclidean Rips. 
In a fixed-seed benchmark of 22 datasets (41,637 samples), CHARM achieves the highest share of first- or second-place scores across 15 metrics. On nonsynthetic data, it ranks first or second on every metric and first on all four $H_0/H_1$ persistence-diagram metrics. 
CHARM couples strong nonsynthetic geometric performance with an auditable structural layer.
\end{abstract}

\par

\section{Introduction}
Dimensionality reduction (DR) maps high-dimensional observations to planar coordinates for visual exploration. Methods such as t-SNE, UMAP, and Isomap optimize geometric relationships, including neighborhoods and graph distances \citep{ref12,ref13,ref14}. These objectives describe the geometry of the projection, but neighborhood fidelity does not establish nontrivial homology classes or identify the input observations associated with their representatives. Geometry and topological certification therefore answer different questions.\par
Topology-aware DR methods add persistent-topology objectives or preserve selected features \citep{ref15,ref16,ref21,ref27,ref30}; many focus on $H_0$ guarantees or topology-sensitive optimization rather than auditable, sample-localized higher-order representatives. Standalone persistent homology provides algebraic answers, yet growing high-dimensional simplicial complexes increase certification cost \citep{ref5,ref6}, and recovered chains are not inherently connected to a planar layout. The expression challenge is equally fundamental: a continuous closed surface such as $S^2$ has nontrivial $H_2$ but cannot be continuously embedded injectively into $\mathbb R^2$ \citep{ref25}. This motivates a planar view accompanied by a queryable structural layer, rather than relying on coordinates alone to convey higher-order topology.

The resulting task has three linked requirements: \emph{compress} the complex while retaining sound structural information; \emph{certify and track} quotient-level homology across scales; and \emph{localize} representatives to original observations in the displayed coordinate system. CHARM addresses them with a certificate-to-visualization pipeline.

\textbf{Approach.} Diffusion-guided cover proposals control the scope of homotopy-safe graph contractions, reducing the subsequent algebraic workload. A Cover Quotient Tower supports cross-scale representative transport, while complete-incidence closures enable boundary and independence tests that produce auditable certificates. Reverse-scale layout uses the retained block hierarchy and quotient-chain preimages to connect the certified records to original samples in one shared two-dimensional embedding.

\textbf{Contributions.} (i) An auditable structural layer that attaches certified quotient-chain records, cross-scale provenance, support overlap, and censoring information to shared planar coordinates.
 (ii) Diffusion-guided, safety-tested homotopy-preserving compression, combining the Cover Quotient 3-interleaving \citep{ref1} with new conditional Euclidean-to-quotient approximation bounds and their composite form.
 (iii) Exact quotient-level $H_0$--$H_2$ cycle and independence decisions on complete incidence closures when reduction completes.
 (iv) Results on 22 datasets:  achieving the highest proportion of top-two rankings across the 15 metrics. 
 (v) Results on 9 nonsynthetic benchmarks: ranking all first on 4 $H_0/H_1$ diagram metrics and all first or second on the other 11 metrics. 

\begin{figure}[htbp]

\centering

\includegraphics[width=\linewidth]{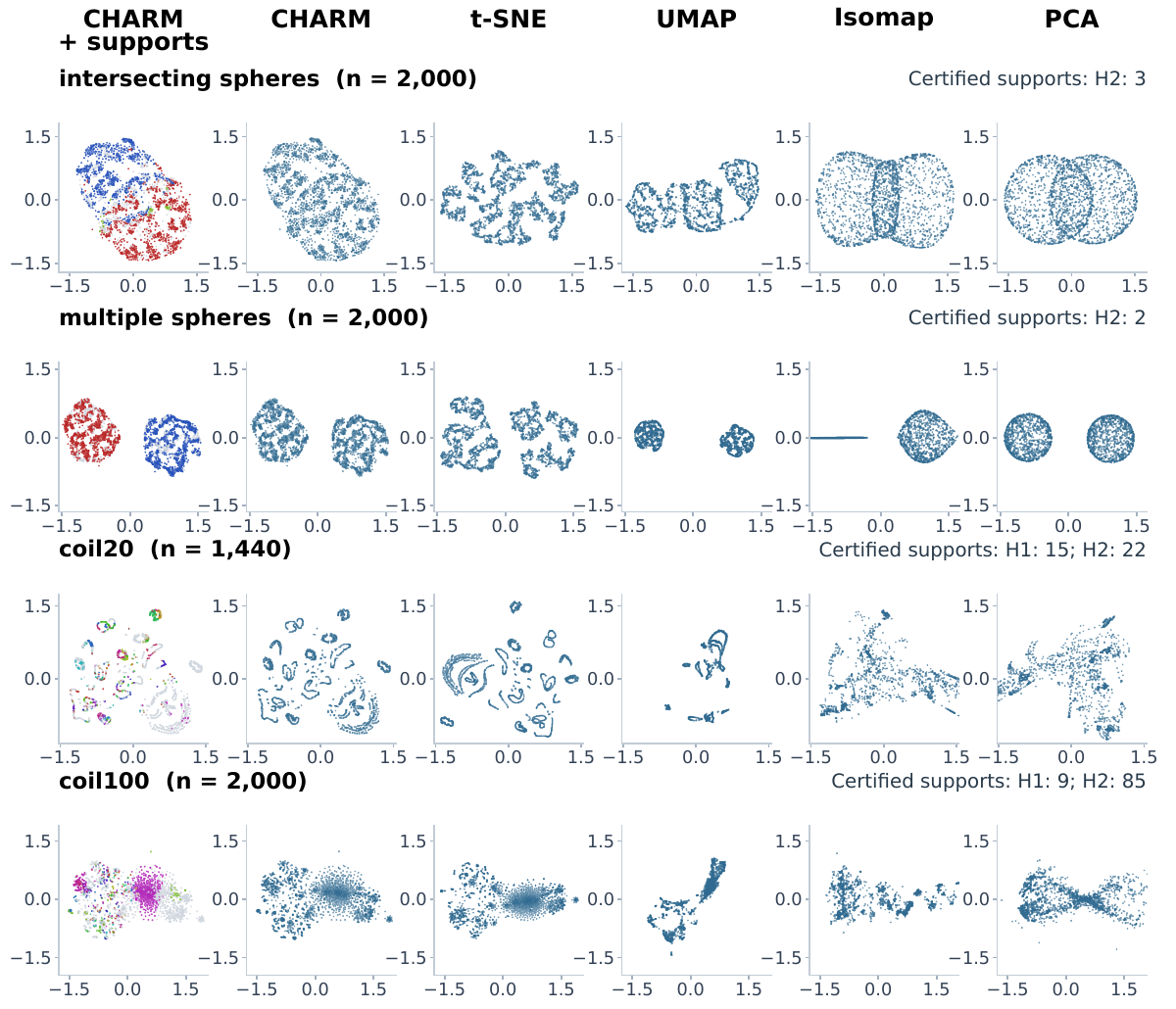}

\caption{One shared planar coordinate system, with certified sample-support annotations. Rows: intersecting spheres, multiple spheres, COIL-20, and COIL-100; columns: CHARM + supports, CHARM, t-SNE, UMAP, Isomap, and PCA. CHARM clearly identifies the respective original-sample supports of the two intersecting shells, including overlapping memberships---a capability absent from t-SNE and UMAP, which obscure their circular contours, and Isomap and PCA, which show two circles without support attribution (Supplement F.4--F.5).} 
\label{fig:showcase}

\end{figure}

\section{Related Work}
Geometric DR methods emphasize local or global relationships: t-SNE and UMAP focus on neighborhoods, while PCA, Isomap, and GLoMAP offer complementary geometric criteria \citep{ref12,ref13,ref14,ref19,ref23,ref24}. Topological Autoencoders and homology-aware landmarking optimize topological structure \citep{ref15,ref27,ref30}; TopoMap and TopoMap++ provide $H_0$-oriented projection guarantees \citep{ref16,ref21}. These approaches primarily target properties of the projected coordinates, whereas CHARM also exposes algebraically certified, sample-traceable higher-order representatives.

Diffusion geometry informs neighborhood summaries \citep{ref2,ref22,ref26}; witness and graph-induced complexes, sparse Rips filtrations, and simplicial towers compress topology computations \citep{ref6,ref7,ref8,ref9,ref28}. Cover Quotient provides nested-cover towers and coarse-metric interleaving \citep{ref1}, while representative-chain algorithms recover explicit cycles \citep{ref29}. These lines either optimize topology in a projected representation without necessarily returning localized $H_k$ certificates, or compute homology apart from the displayed coordinates. CHARM joins safe compression, complete-incidence certification, and representative provenance to a shared reverse-scale planar view. Supplement H details the distinctions.

\section{Method: CHARM}

\subsection{Task and Overview}

Given $X=\{x_i\}_{i=1}^n\subset\mathbb R^D$, CHARM outputs coordinates $Y\in\mathbb R^{n\times2}$ and structural records. Each certified record stores a quotient-level representative, its dimension, first certification scale $b_{\mathrm{cert}}$, tracking status, and the input samples associated with its quotient vertices. Its representative is certified at $b_{\mathrm{cert}}$; right-censored tracks retain their last confirmed survival scale. We call this representative-dependent preimage its \textbf{homology support}. We write $k_{\mathrm{nn}}$ for graph neighborhood size and $k$ for homological dimension, using $\mathbb F_2$ coefficients for certification.\par

A shared membership hierarchy connects diffusion-guided proposals, compressed quotient towers, certified homology records, and reverse-scale coordinates to the original samples. Supplement A--C give implementation details and pseudocode.\par

\textbf{Execution order.} At each hop event, CHARM updates the partition and quotient map, transports earlier certified representatives, then tests new independent classes. Each accepted class receives its first certification event $b_{\mathrm{cert}}$. Reverse traversal of the saved block forest initializes shared coordinates for geometric and persistent-homology (PH) optimization; retained memberships produce sample-level annotations. Supplement B.11 and C specify the event logic.\par

\subsection{Diffusion Cover Proposals and Safe Contraction}

Diffusion geometry aggregates multistep connectivity into distances that guide compression regions \citep{ref2,ref22,ref26}; graph tests certify the subsequent contractions. Construct the symmetric union $k_{\mathrm{nn}}$-nearest-neighbor graph $G_0$ by connecting pairs selected by either endpoint. Let $h_i$ be the distance to the $k_{\mathrm{nn}}$th positive-distance neighbor. Define an adaptive Gaussian kernel on graph edges and a lazy Markov operator:
\begin{align*}
   K_{ij}&=\exp\!\left(-\frac{\|x_i-x_j\|^2}{2h_ih_j}\right),\\
   P_L&=\tfrac12\left(I+\operatorname{diag}(K\mathbf1)^{-1}K\right).
\end{align*}
Nonedges have zero weight. Compute truncated diffusion coordinates (capped at the number of nontrivial component modes) at fixed positive integer time $t_{\mathrm{prop}}$ separately for each connected component. Let $1=\lambda_0>\lambda_1\ge\cdots\ge0$ be the lazy operator's eigenvalues, with eigenvectors $\psi_j$ orthonormal under its stationary measure $\pi$. The full diffusion distance is \citep{ref2,ref22}
\begin{align*}
    D_t^2(i,j)&=\sum_{a\ge1}\lambda_a^{2t}(\psi_a(i)-\psi_a(j))^2, \\
    \Phi_t^{(s_D)}(i)&=\bigl(\lambda_a^t\psi_a(i)\bigr)_{a=1}^{s_D}.
\end{align*}

The factors $\lambda_a^t$ suppress rapidly varying modes. For an exact spectrum, the fixed-graph squared-distance truncation error is bounded by $\lambda_{s_D+1}^{2t}(\pi_i^{-1}+\pi_j^{-1})$; it vanishes when all modes are retained (Supplement D.13).\par

In the diffusion coordinates, we propose overlapping regions covering all samples. For each region, we retain the portion connected to its center in $G_0$. These regions specify candidate contractions, which we evaluate with combinatorial safety tests on the current graph.\par

For each candidate edge $uv$ whose endpoints lie within one proposed region, we inspect its global adjacency. Excluding $u,v$, let $L$ contain vertices adjacent to $u$ but not $v$, and let $R$ contain vertices adjacent to $v$ but not $u$. We accept the contraction when no edge connects $L$ to $R$, equivalently when no chordless four-cycle contains $uv$. This edgewise test supplies the homotopy-preservation guarantee stated below.\par

All tests use the current global graph. Write $\operatorname{Flag}(G)$ for the \textbf{flag complex} whose simplices are all cliques of $G$. The final coarse graph $G_c$ satisfies
\begin{equation*}
\operatorname{Flag}(G_c)\simeq\operatorname{Flag}(G_0).
\end{equation*}
Each coarse vertex $a$ stores its original members $B_a$, Euclidean centroid $c_a$, and mass $\mu_a=|B_a|/n$ for certification and layout. Homotopy preservation follows from the edgewise safety tests; full conditions and proofs appear in Supplement B.1--B.2 and Supplement Theorem D.1.\par

\subsection{Certification and Support Localization on a Quotient Tower}

\textbf{Scales from coarse-graph hop distances.} The nested-cover construction of Cover Quotient \citep{ref1} expresses growing Rips structures through relations among active quotient vertices. After fixing the diffusion time, CHARM defines topological scales using within-component shortest-hop distances $d_H$ on $G_c$. At scale $r$, activate coarse pairs satisfying $d_H(a,b)\le r$. For current partition blocks $C,D$, the complete-linkage graph has an edge precisely when $\max_{a\in C,b\in D}d_H(a,b)\le r$. CHARM merges a connected component of this graph when it is a clique, updating $\mathcal P(r)$ conservatively and consistently at tied events. By induction, each resulting block has $d_H$-diameter at most $r$. Quotient adjacency uses the distinct minimum-distance condition:
\begin{gather*}
b_{ab}=d_H(a,b),\\
Q(r)=\operatorname{Flag}(\mathcal P(r),E_r),\\
E_r=\{\{C,D\}:C\ne D,\min_{a\in C,b\in D}d_H(a,b)\le r\}.
\end{gather*}
Edge insertions, block mergers, and their induced simplicial maps form a \textbf{quotient tower}, with connected components processed separately. The tower records structural evolution, and block parent-child relations provide the layout hierarchy. Preprocessing contractions preserve the base homotopy type; subsequent scale evolution permits homology classes to die.\par

Boundary reductions operate on active quotient simplices with reusable local algebra. Section~\ref{sec:bounds} gives approximation bounds; Supplement E gives complexity.\par

\textbf{From candidates to homology certificates.} Representative chains turn homology classes into localizable geometric supports \citep{ref29}, so CHARM stores both certification results and their chain-level provenance. For $H_0,H_1,H_2$, completed enumeration and reduction yield exact quotient-level decisions. For requested $k\ge3$, an optional resource-aware mode uses mass-weighted local rank to select search regions and simplex-incidence constraints to generate candidates (Supplement B.6--B.8); the present experiments evaluate $H_0$--$H_2$. Every certified representative $z$ satisfies
\begin{equation*}
\partial_kz=0,\qquad
z\notin\operatorname{im}\partial_{k+1}
\end{equation*}
and independence from the existing homology directions. To certify these conditions, we construct complete \textbf{incidence blocks} from the relations between simplices and their cofaces one dimension higher. We then close the relevant blocks and historical representatives under these relations. Algebra on the resulting blocks gives the same boundary and independence decisions as the corresponding global tests (Supplement Theorem D.8); Supplement B.9--B.10 specify the construction.\par

Certified representatives are pushed along tower maps, recording their first certification scale $b_{\mathrm{cert}}$ and confirmed track termination. A boundary image certifies class vanishing; dependence on earlier representatives only terminates an independent direction. If survival is confirmed up to a scale and testing subsequently stops at the terminal scale or resource budget, the record is right-censored at its last confirmed survival. Supplement B.11 specifies tracking.\par

We display certified classes that are independent of existing directions and have traceable survival records, giving priority to longer confirmed survival.\par

\textbf{Early certificate and exact fallback.} For an incidence-two support $Z$, every $(k-1)$-face occurs twice, so $z_Z=\sum_{\sigma\in Z}\sigma$ is a cycle over $\mathbb F_2$. A fast sufficient non-boundary test checks the common neighbors of every $k$-simplex $\sigma$ in its support:
\begin{equation*}
\exists\sigma\in\operatorname{supp}(z_Z):\quad
\bigcap_{v\in\sigma}N_Q(v;r)=\varnothing.
\end{equation*}
Such a simplex has no $(k+1)$-coface and therefore cannot occur in a $(k+1)$-boundary; hence $[z_Z]\neq0$. Failure of this \emph{sufficient} test does not imply that the cycle is a boundary. CHARM instead reduces the candidate against the complete $(k,k+1)$-incidence block closure containing its simplices. If $\mathcal B_J=\operatorname{im}(\partial_{k+1}|_J)$ denotes the local boundary subspace, block completeness gives the exact global boundary verdict. For independence, the closure is enlarged through all accepted representatives sharing incident blocks, and the new chain is tested modulo the sum of those representatives and boundary spaces. This transitive representative--block step is necessary: testing only the directly incident blocks can miss a linear dependence mediated by other historical representatives (Supplement B.9--B.11, D.4--D.8).\par

\textbf{Localizing original samples.} For a quotient block $C$, define $\Pi(C)=\bigcup_{a\in C}B_a$. If $V(z)$ denotes the quotient vertices in the representative, then
\begin{equation*}
\operatorname{Supp}_X(z)=\bigcup_{C\in V(z)}\Pi(C).
\end{equation*}
This map identifies the original observations attached to the quotient-chain vertices, enabling auditable sample annotations and overlap analysis in the planar view. Algebraic certification applies to the quotient chain; the preimage supplies its traceable original-sample membership.\par

\subsection{Reverse-scale Layout and Shared Coordinate Optimization}

The two-dimensional view shows sample relationships together with homology locations. CHARM uses one coordinate matrix $Y$ for all original samples and all scales. Within each frame, a block is positioned at its members' centroid. For frame $t$, define the membership-averaging matrix\par

\begin{equation*}
(A_t)_{Ci}=\frac{\mathbf1\{i\in\Pi(C)\}}{|\Pi(C)|},\qquad Y_t=A_tY.
\end{equation*}

Thus skeleton constraints at every scale act on the same $Y$. The shared geometric objective is
\begin{align*}
\mathcal L_{\mathrm{geom}}(Y)&=D_{\mathrm{KL}}(P^{\mathrm{aff}}\|Q^{\mathrm{aff}}(Y))\\
&+\lambda_{\mathrm{metric}}\mathcal L_{\mathrm{metric}}(Y)\\
&+\lambda_{\mathrm{rev}}\mathcal L_{\mathrm{reverse}}(Y).
\end{align*}
Here $P^{\mathrm{aff}}$ contains original-sample neighborhood affinity probabilities, and $Q^{\mathrm{aff}}(Y)$ uses a heavy-tailed Student-$t$ kernel. The neighborhood term matches local similarities, with the heavy tail alleviating two-dimensional crowding \citep{ref12}. The metric term constrains pairwise geometric distortion, and the reverse-scale term aligns member centroids $A_tY$ with skeleton edge-length relations.\par

We initialize the layout at the coarsest recorded scale that retains the target homology supports, then follow the block refinements toward finer scales. Unsplit blocks inherit their positions. Children start near their parents and adjust locally according to quotient adjacency, while parent constraints suppress cross-scale drift. When no certified structure is available for display, we initialize from the graph skeleton. This reverse-scale procedure uses the forward tracking records to initialize point-level optimization and define skeleton constraints.\par

Final points, support colors, and frame centroids all derive from $Y$. Each displayed certified record has the form $(k,z,b_{\mathrm{cert}},\text{tracking status},\operatorname{Supp}_X(z))$, joining a quotient-level certificate to a selectable original-sample preimage. Edge-chain overlays for $H_1$ and triangle-chain overlays for $H_2$ display representative geometry; record labels specify certified homological dimension and provenance. The same coordinates support inspection of overlapping records and comparisons across scales. Supplement B.12, D, and E give the layout, theory, and complexity analysis.\par

\subsection{Computational Structure and Certification Status}

Coarse-vertex contraction reduces the input size of
quotient-tower computation, while complete incidence
blocks and reusable echelon bases enable localized
certification. Let $n_c,m_c$ denote coarse-graph
vertices and edges. Radius-truncated BFS from all
coarse vertices costs
\begin{equation*}
O\!\left(
\sum_{a\in V_c}
(|V_a^{(R)}|+|E_a^{(R)}|)
\right),
\end{equation*}
where $V_a^{(R)},E_a^{(R)}$ are visited vertices
and scanned edges. The worst-case time is
$O(n_c(n_c+m_c))$; storing $m_R$ returned hop
pairs requires $O(m_R)$ space, reaching
$\Theta(n_c^2)$ in the worst case.

For $H_2$, a quotient clique on $v$ vertices
contains $\binom{v}{3}$ triangles and
$\binom{v}{4}$ tetrahedra, exposing the
combinatorial cost of full simplex enumeration.
Naive dense elimination on an $R_J\times C_J$
boundary block costs
$O(R_JC_J\min\{R_J,C_J\})$ time and
$O(R_JC_J)$ space. Across scales, algebraic
cost accumulates over processed blocks and
independence tests; sparse reductions, complete
incidence localization, and valid cached bases
reduce repeated work.

Higher-dimensional incidence-two discovery can
branch exponentially under ambiguous coface
continuations. Completed algebraic decisions
yield certified records; right-censored tracks
retain their last confirmed survival scales.
Higher-dimensional discovery remains
seed-dependent for $k\ge3$
(Supplement D.9 and E).\par

\subsection{Error Guarantees from Original Rips to Coarse Graphs and Quotient Towers}
\label{sec:bounds}

\textbf{Quotient tower relative to the coarse graph.} Let $\mathcal R_H(r)=\operatorname{VR}(V_c,d_H;r)$, using the convention that Rips edges have length at most $r$. Since $\mathcal P(r)$ is nested and each block has $d_H$-diameter at most $r$, projection and within-block representative selection induce
\begin{align*}
H_k(\mathcal R_H(r))&\longrightarrow H_k(Q(r)),\\
H_k(Q(r))&\longrightarrow H_k(\mathcal R_H(3r)).
\end{align*}
The factor 3 accounts for two within-block distances and one interblock witness edge. The scale-compatible homology maps give the established multiplicative \textbf{3-interleaving} with the coarse-graph filtration, or $d_I,d_B\le\log3$ on logarithmic scales (Supplement Theorem D.3; \citep{ref1,ref11}).\par

\textbf{Proof sketch.} Choose a representative $a_i\in C_i$ per quotient block. For each quotient edge, witnesses $u\in C_i,v\in C_j$ have $d_H(u,v)\le r$; the block diameter bound yields $d_H(a_i,a_j)\le r+r+r=3r$. Hence representatives of a quotient simplex span a Rips simplex at $3r$. The reverse projection is simplicial at $r$, and the induced maps commute across scales.\par

\textbf{Coarse graph relative to original Euclidean Rips.} First assume $G_0$ is connected. Let $q:X\to V_c$ be the membership projection retained by safe contraction. Calibrate hop distances with a length unit $\ell>0$, setting $\widetilde d_H=\ell d_H$, and let
\begin{align*}
\mathcal R_X(r)&=\operatorname{VR}(X,\|\cdot\|;r),\\
\widetilde{\mathcal R}_H(r)&=\mathcal R_H(r/\ell),\quad
\widetilde Q(r)=Q(r/\ell).
\end{align*}
If all original sample pairs satisfy the verified bound
\begin{equation*}
\left|\|x_i-x_j\|-\ell d_H(q(i),q(j))\right|\le\varepsilon,
\tag{1}
\end{equation*}
then $H_k(\mathcal R_X)$ and $H_k(\widetilde{\mathcal R}_H)$ are additively $\varepsilon$-interleaved, and their bottleneck distance on the original scale is at most $\varepsilon$ (Supplement Theorem D.11). A sufficient bound computable without enumerating all original sample pairs is
\begin{align*}
\varepsilon&=2\rho+\eta_\ell,\quad
\rho=\max_a\max_{i\in B_a}\|x_i-c_a\|,\\
\eta_\ell&=\max_{a,b}\bigl|\|c_a-c_b\|-\ell d_H(a,b)\bigr|.
\tag{2}
\end{align*}
Here $\rho$ measures aggregation radius, and $\eta_\ell$ measures the distortion between coarse hop distances and Euclidean centroid distances. The latter uses exact values for all coarse pairs or a proven bound covering those pairs. This separates aggregation error from graph-metric distortion.\par

\textbf{Composite guarantee.}
Under condition (1), the forward map is
$\mathcal R_X(r)\to\widetilde Q(r+\varepsilon)$,
whereas quotient-block representative selection
induces
$\widetilde Q(r)\to\mathcal R_X(3r+\varepsilon)$.
The factor $3$ acts on the coarse metric before
the single Euclidean distortion term. These maps
form a two-sided interleaving with
$\phi(r)=3r+\varepsilon$ and composites equal
to the structure maps at
$\phi^2(r)=9r+4\varepsilon$
(Supplement Theorem D.12).

For $r\ge r_{\min}>0$, define the restricted
persistence modules
\begin{equation*}
\mathbb H_X^{+}
=H_k(\mathcal R_X)|_{[r_{\min},\infty)},
\ 
\mathbb H_Q^{+}
=H_k(\widetilde Q)|_{[r_{\min},\infty)}
\end{equation*}
Their multiplicative interleaving constants obey
\begin{align*}
c_H=1+\frac{\varepsilon}{r_{\min}},\ 
c_Q&=3+\frac{\varepsilon}{r_{\min}},\\
d_B^{\log}\bigl(
\operatorname{Dgm}(\mathbb H_X^{+}),
\operatorname{Dgm}(\mathbb H_Q^{+})
\bigr)&\le\log c_Q
\tag{3}
\end{align*}
Indeed, $\phi(r)\le c_Qr$ and
$\phi(c_Qr)\le c_Q^2r$ for all
$r\ge r_{\min}$, so both maps and their
composites advance to the required
multiplicative structure maps.

A finite interval $[b,d)$ in the restricted
quotient barcode with $d/b>c_Q^2$ must match
a nonzero original-Rips interval, with
multiplicative endpoint discrepancy at most
$c_Q$ on the restricted scale. Its original
birth endpoint is also untruncated when
$b>c_Qr_{\min}$. A confirmed quotient
survival range $[b,s]$ with
$s>c_Q^2b$ likewise implies a nonzero
original-Rips persistent map from $c_Qb$
to $s/c_Q$, subject to scale availability.
Record certification and barcode birth
endpoints need not coincide.

These guarantees compare the original
Euclidean Rips filtration with the quotient
tower, rather than the final planar embedding.
Global diagram bounds require full extended
filtrations; finite runs support the
corresponding local persistence statements
(Supplement D.10--D.12).

\section{Experiments}
CHARM is evaluated as a joint visual and algebraic representation: certified homology records and their sample memberships establish the structural output, while neighborhood, graph-geodesic, Euclidean diagram, and runtime results assess the shared planar coordinates and computational trade-offs. All reported displays and scores use the same final coordinates.\par

\subsection{Data, Baselines, and Evaluation Protocol}

\textbf{Data and configuration.} We use 22 datasets with 41,637 samples: 9 nonsynthetic datasets (15,637 samples)---\texttt{digits}, \texttt{mnist}, \texttt{fashion\_mnist}, \texttt{coil20}, \texttt{olivetti}, \texttt{coil100}, \texttt{mouse\_scrna}, \texttt{googlenews}, and \texttt{flow\_cytometry}---and 13 synthetic datasets (26,000 samples). Baselines are t-SNE, UMAP, Isomap, and PCA, with random seed 0 throughout. CHARM uses 50 graph neighbors, proposal diffusion time 4, 32 spectral modes, and a region-size cap of 512. Geometric optimization uses perplexity 50, at most 600 iterations, and Euclidean training-distance constraints. Final outputs also undergo at most 3 persistent-homology (PH) correction steps and leaf restoration.\par

t-SNE uses perplexity 50 and at most 600 iterations; UMAP uses 50 neighbors, 600 epochs, and \texttt{min\_dist=0.1}; Isomap uses 50 neighbors; PCA uses full singular value decomposition. Supplement F.1 provides the complete dataset list, per-dataset sample counts, and training configurations; Supplement G reports per-dataset evaluation results.\par

\textbf{Evaluation.} Local metrics use all samples and Euclidean distances in the original and two-dimensional spaces, with neighborhood size 50. Graph-geodesic metrics use an independently constructed original-space union 15-neighbor graph with Euclidean edge weights and evaluate reachable pairs. Topological discrepancies compare Euclidean Vietoris--Rips (VR) persistence diagrams of the full original and embedded point clouds, computing $H_0,H_1$ over $\mathbb F_2$. Each cloud is centered and normalized by its root-mean-square (RMS) radius before computing bottleneck and order-2 Wasserstein distances.\par

Each dataset is scored separately; aggregate scores are weighted by sample count. The main text reports six neighborhood/distance metrics and four persistence-diagram distances. Supplement F.2--F.3 give all 15 metrics; Supplement G reports every dataset. Within each group and metric in the aggregate tables, bold and underlined values mark the best and second-best results according to unrounded scores.\par

\textbf{Experimental design.} All methods use the stated fixed settings and random seed 0. For CHARM, PH correction explicitly optimizes toward Euclidean persistence diagrams of the same input; diagram distances therefore measure the fidelity attained by the complete optimized pipeline. The 22 datasets were also used during method development.\par

\subsection{Certified Homology and Sample-level Localization}
\label{sec:localization}

CHARM augments planar coordinates with quotient-level certified classes with completed birth tests and confirmed or right-censored tracking, together with their original-sample preimages. Across 22 datasets, it produces 957 cross-scale certified records connecting algebraic classes to original observations and their track status (confirmed termination or right-censoring). Figure~\ref{fig:showcase} displays the shared coordinates with and without support annotations, highlighting the structural information contributed by certification.\par

\textbf{Intersecting spheres: identifying geometry and support membership.} Figure~\ref{fig:showcase} makes the distinction concrete. CHARM preserves a recognizable planar view of the two intersecting spherical shells and uses certified support annotations to assign original samples to their respective quotient-chain preimages, including samples belonging to more than one support. In contrast, t-SNE and UMAP fragment or distort the shells so that the two complete circular contours are not discernible. Isomap and PCA reveal two intersecting circular outlines, but their coordinates alone cannot attribute the overlap to the corresponding original-sample supports. CHARM provides both the visible two-shell organization and queryable support provenance in the same coordinates; the annotated and unannotated CHARM views use exactly the same embedding. The \texttt{intersecting\_spheres} benchmark contains three tracked $H_2$ records, covering 1,603 of 2,000 samples, with 906 samples present in multiple supports (Supplement F.4--F.5).\par

\textbf{Disjoint versus overlapping supports.} The two $H_2$ records of \texttt{multiple\_spheres} identify disjoint sets of 796 and 802 original samples, matching the visually separated shells in the second row of Fig.~\ref{fig:showcase}. For \texttt{intersecting\_spheres}, three tracked $H_2$ records expose overlapping memberships across the two visible shells rather than forcing the intersection into a single exclusive label. A further $H_2$ record on \texttt{nonuniform\_sphere} covers 1,623 samples. These examples show how support provenance distinguishes separated components from intersecting structures beyond what planar contours alone communicate (Supplement F.4--F.5).\par

\textbf{Image-data memberships.} CHARM also records shared memberships in image collections: 332 COIL-20 samples and 720 COIL-100 samples belong to multiple stored supports. Table~\ref{tab:support_members} summarizes the number of $H_1/H_2$ records, unique covered samples, and overlapping members. Record counts enumerate tracked quotient classes across scales rather than the Betti number at any single scale; coverage and overlap come directly from representative preimages.\par
\begin{table}
\centering
\caption{Original-sample membership for selected datasets (Supplement F.5). ``Overlap'' counts samples present in two or more stored support records.}
\label{tab:support_members}
\small\setlength{\tabcolsep}{6pt}
\begin{adjustbox}{max width=\linewidth}
\begin{tabular}{lcccc}
\toprule
Dataset & $H_1$ & $H_2$ & Covered & Overlap \\
\midrule
sphere & 0 & 1 & 1,578 & 0 \\
multiple\_spheres & 0 & 2 & 1,598 & 0 \\
intersecting\_spheres & 0 & 3 & 1,603 & 906 \\
coil20 & 15 & 22 & 536 & 332 \\
coil100 & 9 & 85 & 884 & 720 \\
\bottomrule
\end{tabular}
\end{adjustbox}
\end{table}
\textbf{Track status.} Each record is tracked across scales, and support membership is attached to each stored representative.\par
\subsection{Geometric Fidelity and Neighborhood Preservation}

CHARM attains both the highest continuity and the lowest graph-geodesic stress in the sample-weighted aggregates over nonsynthetic datasets and over all datasets. Table~\ref{tab:geometry} reports trustworthiness (T50), continuity (C50), neighborhood overlap (Q50), log-weighted neighborhood preservation across scales (RNX\_AUC\_log), graph-geodesic triplet accuracy (RTA\_G), and distance stress (stress\_G), covering local and global geometry.
\begin{table}
\small\setlength{\tabcolsep}{4pt}
\caption{\textbf{Neighborhood and graph-geodesic preservation.} Aggregation is weighted by sample count within each group. Higher is better for \(\uparrow\); lower is better for \(\downarrow\).}
\label{tab:geometry}
\begin{adjustbox}{max width=\linewidth}
\begin{tabular}{lrrrrrrr}
\toprule
Group & Method & T50 \(\uparrow\) & C50 \(\uparrow\) & Q50 \(\uparrow\) & RNX\_AUC\_log \(\uparrow\) & RTA\_G \(\uparrow\) & stress\_G \(\downarrow\) \\
\midrule
Nonsynthetic & CHARM & \textbf{0.932820} & \textbf{0.940222} & \underline{0.533032} & \underline{0.478862} & \underline{0.767870} & \textbf{0.301757} \\
Nonsynthetic & t-SNE & \underline{0.931321} & \underline{0.935928} & \textbf{0.534062} & \textbf{0.479139} & 0.754928 & 0.335772 \\
Nonsynthetic & UMAP & 0.927676 & 0.934815 & 0.514725 & 0.368367 & 0.749070 & 0.372863 \\
Nonsynthetic & Isomap & 0.839815 & 0.923977 & 0.298785 & 0.270208 & \textbf{0.781931} & \underline{0.321520} \\
Nonsynthetic & PCA & 0.823711 & 0.899779 & 0.274950 & 0.259412 & 0.729116 & 0.379699 \\
All & CHARM & 0.961093 & \textbf{0.955756} & \underline{0.650620} & \underline{0.561491} & \underline{0.775532} & \textbf{0.308088} \\
All & t-SNE & \textbf{0.967636} & 0.950764 & \textbf{0.664755} & \textbf{0.607781} & 0.767427 & 0.325245 \\
All & UMAP & \underline{0.962655} & \underline{0.954248} & 0.638602 & 0.516560 & 0.773681 & 0.334540 \\
All & Isomap & 0.853403 & 0.950800 & 0.385501 & 0.369980 & \textbf{0.783225} & \underline{0.312149} \\
All & PCA & 0.851939 & 0.942362 & 0.368305 & 0.361207 & 0.763374 & 0.327113 \\
Synthetic & CHARM & 0.978097 & 0.965099 & \underline{0.721340} & \underline{0.611187} & 0.780140 & 0.311896 \\
Synthetic & t-SNE & \textbf{0.989477} & 0.959687 & \textbf{0.743358} & \textbf{0.685150} & 0.774944 & 0.318913 \\
Synthetic & UMAP & \underline{0.983692} & 0.965936 & 0.713105 & 0.605687 & \textbf{0.788482} & 0.311491 \\
Synthetic & Isomap & 0.861575 & \underline{0.966932} & 0.437654 & 0.429985 & \underline{0.784003} & \underline{0.306513} \\
Synthetic & PCA & 0.868916 & \textbf{0.967973} & 0.424450 & 0.422428 & 0.783978 & \textbf{0.295487} \\
\bottomrule
\end{tabular}
\end{adjustbox}
\end{table}
On nonsynthetic data, CHARM combines the best T50 and C50 with the lowest stress\_G. Its Q50 is 0.533032, close to t-SNE's 0.534062, while the corresponding stress\_G values are 0.301757 and 0.335772. Thus the same coordinates provide similar neighborhood overlap with lower graph-geodesic distortion. On the 15 nonsynthetic sample-weighted metrics, CHARM is best on seven and second-best on eight (Supplement F.3).\par

On individual nonsynthetic datasets, CHARM reduces graph-geodesic stress relative to t-SNE by 15.0\% on MNIST, 21.3\% on mouse single-cell data, and 25.8\% on flow cytometry. On COIL-20, its $Q50=0.725486$ compares with 0.494500 for Isomap and 0.443889 for PCA; on Google News embeddings, it reaches $Q50=0.339700$ versus PCA's 0.112770. Synthetic datasets show complementary strengths across neighborhood and distance criteria (Supplement F.3 and G).\par

\subsection{Euclidean Persistence-diagram Fidelity}

On nonsynthetic data, CHARM attains the lowest values on all four Euclidean persistence-diagram distances between the original and final planar point clouds (Table~\ref{tab:pd}). This $H_0/H_1$ coordinate-fidelity assessment complements the algebraically certified higher-order quotient records in Section~\ref{sec:localization}. The table evaluates the complete PH-corrected CHARM pipeline; baselines have no analogous diagram-matching correction. It therefore measures full-pipeline fidelity rather than isolating the embedding contribution.
\begin{table}
\centering
\small\setlength{\tabcolsep}{4pt}
\caption{\textbf{Euclidean VR persistence-diagram distances.} dB denotes bottleneck distance and W2 denotes order-2 Wasserstein distance; lower is better throughout. Input and embedding are RMS-normalized separately.}
\label{tab:pd}
\begin{adjustbox}{max width=\linewidth}
\begin{tabular}{lrrrrr}
\toprule
Group & Method & H0\_dB \(\downarrow\) & H0\_W2 \(\downarrow\) & H1\_dB \(\downarrow\) & H1\_W2 \(\downarrow\) \\
\midrule
Nonsynthetic & CHARM & \textbf{0.668314} & \textbf{13.266261} & \textbf{0.102801} & \textbf{0.753020} \\
Nonsynthetic & t-SNE & 0.753144 & 13.354466 & 0.105936 & 0.776184 \\
Nonsynthetic & UMAP & 0.723239 & 13.331322 & 0.114668 & \underline{0.770062} \\
Nonsynthetic & Isomap & 0.749747 & 13.334842 & 0.105515 & 0.780641 \\
Nonsynthetic & PCA & \underline{0.713917} & \underline{13.300012} & \underline{0.104980} & 0.787859 \\
All & CHARM & \textbf{0.325450} & \textbf{6.099576} & \underline{0.195031} & \textbf{0.728950} \\
All & t-SNE & 0.379915 & 6.245110 & 0.208424 & 0.762859 \\
All & UMAP & 0.386242 & 6.262154 & 0.218139 & 0.766165 \\
All & Isomap & 0.355021 & 6.161252 & 0.198971 & 0.755962 \\
All & PCA & \underline{0.341532} & \underline{6.146908} & \textbf{0.189897} & \underline{0.753322} \\
Synthetic & CHARM & 0.119244 & \textbf{1.789366} & \underline{0.250500} & \textbf{0.714474} \\
Synthetic & t-SNE & 0.155447 & 1.969379 & 0.270063 & 0.754846 \\
Synthetic & UMAP & 0.183564 & 2.010593 & 0.280369 & 0.763822 \\
Synthetic & Isomap & \underline{0.117624} & 1.846890 & 0.255177 & 0.741119 \\
Synthetic & PCA & \textbf{0.117571} & \underline{1.844867} & \textbf{0.240968} & \underline{0.732550} \\
\bottomrule
\end{tabular}
\end{adjustbox}
\end{table}
Across all datasets, CHARM attains the smallest H0\_dB, H0\_W2, and H1\_W2, together with the second-smallest H1\_dB; PCA attains the smallest H1\_dB. On synthetic data, CHARM minimizes both W2 distances, and PCA minimizes both bottleneck distances. This distinction reflects the metrics' different emphases: W2 summarizes cumulative matching discrepancy, whereas bottleneck distance measures the largest matching deviation. CHARM's W2 results therefore complement its local-neighborhood and support-localization results.\par

Under the fixed-seed, in-sample protocol, these diagram distances quantify the Euclidean $H_0/H_1$ fidelity attained by the PH-corrected final coordinates. Together, Tables~\ref{tab:geometry}--\ref{tab:pd} show competitive neighborhood and graph-geodesic preservation alongside the four lowest nonsynthetic diagram discrepancies.\par

\subsection{Runtime}
CHARM yields structural records and planar coordinates in a mean elapsed time of 17.96 s across the 22 datasets, versus 15.94 s for t-SNE and 16.27 s for UMAP. Its measured pipeline includes diffusion-cover construction, safe contraction, quotient-tower certification and tracking, reverse-scale layout optimization, PH correction, and leaf restoration; baselines use their standard full pipelines (Supplement I).\par

\subsection{Evaluation Scope}
Tables~\ref{tab:geometry}--\ref{tab:pd} score final planar coordinates; Table~\ref{tab:support_members} describes structural preimages. All 22 datasets were used during development and evaluated with the stated fixed-seed, in-sample PH-corrected protocol. Recorded survival and right-censoring distinguish confirmed persistence from untested future events (Supplement B.11, F.5); evaluation provenance and broader discussion appear in Supplement F.7 and H.

\section{Conclusion}
CHARM makes dimensionality reduction both visually informative and structurally accountable. Diffusion-guided, homotopy-safe contraction reduces the complex before certification; the Cover Quotient Tower and complete-incidence reductions then identify and track independent $H_0$--$H_2$ classes. Representative preimages turn these algebraic records into original-sample memberships, while reverse-scale optimization brings them into a single interpretable planar view. The coarse-hop 3-interleaving and conditional Euclidean Rips bounds connect this efficient pipeline to explicit approximation guarantees.

The intersecting-spheres comparison reveals a central advantage over geometric DR. CHARM displays both shells while identifying their distinct and overlapping input-sample supports; t-SNE and UMAP fail to reveal the complete circular contours, and Isomap and PCA show the intersection without support attribution. On separated spheres, CHARM distinguishes two disjoint $H_2$ supports, demonstrating that its annotations expose structural membership rather than merely decorating a plot. Across the 22 datasets, 957 certified, traceable records make these relationships inspectable beyond the two-dimensional image.

This additional structural information comes with strong geometric results. Across nine nonsynthetic datasets, CHARM ranks first or second on all 15 sample-weighted metrics, leads trustworthiness and continuity, and has the lowest graph-geodesic stress. Its completed PH-corrected pipeline also yields the lowest values on all four $H_0/H_1$ Euclidean persistence-diagram distance metrics. CHARM thus combines neighborhood preservation competitive with t-SNE, lower nonsynthetic global distortion, and finer-grained topological fidelity with sample-level localization unavailable from the compared coordinate-only outputs. The resulting embedding supports not only visual pattern discovery but also direct inspection of which observations constitute each certified structure.
\par\smallskip\noindent\textbf{LLM Usage Statement.} Large language models were used solely to refine manuscript language and presentation. The algorithms and experimental results were developed by the authors.\par
\end{document}


\twocolumn[{\centering\Large\bfseries CHARM: Certified Homology-Aware Reverse-scale Mapping\par
\normalsize\bfseries Supplementary Material\par\vspace{1.1em}}]
\section*{Appendix Guide}

\begin{center}
\small\setlength{\tabcolsep}{4pt}
\begin{tabularx}{\linewidth}{lX}
\toprule
Appendix & Contents \\
\midrule
A & Chains, homology, flag complexes, towers, and formal output definitions \\
B & Full method details: diffusion covers, safe contraction, certification, tracking, and shared coordinates \\
C & Concise pseudocode for the pipeline and four subroutines \\
D & Safe contraction, Rips/tower bounds, exact certification, conditional support recovery, and diffusion smoothing \\
E & Complexity of graph construction, search, certification, and layout \\
F & Data and training protocols, 15 metrics, grouped tables, point-cloud figures, supports, and implementation sources \\
G & Per-dataset results for 22 datasets, 5 methods, and 15 metrics \\
H & Extended context, dataset-level interpretation, and limitations \\
I & Runtime summary with timing provenance caveats \\
\bottomrule
\end{tabularx}
\end{center}

\appendix

\section{Preliminaries and Formal Problem Definition}

\subsection{Simplicial Chains and Homology}
We work over $\mathbb F_2$, writing $k_{\mathrm{nn}}$ for graph neighborhood size and $k$ for homological dimension. For a finite simplicial complex $K$, the chain spaces and boundary maps give
\begin{align*}
\partial_k&:C_k(K)\to C_{k-1}(K),\\
H_k(K)&=\ker\partial_k/\operatorname{im}\partial_{k+1}.
\end{align*}
A cycle is nontrivial exactly when it is not a boundary. Mod-2 coefficients also give nonorientable closed manifolds, including the Klein bottle and $\mathbb{RP}^2$, a top-dimensional fundamental class.

\subsection{Flag Complexes}
The flag complex $\operatorname{Flag}(G)$ consists of the cliques of an undirected simple graph $G$; a $(k+1)$-clique is a $k$-simplex. Open-neighborhood intersections enumerate cofaces and incident simplices:
\begin{align*}
\operatorname{Cof}_{k+1}(\sigma)&=\{\sigma\cup\{w\}:\\
&\qquad w\in\bigcap_{v\in\sigma}N(v)\},\\
\operatorname{Inc}_k(f)&=\{f\cup\{w\}:w\in\bigcap_{v\in f}N(v)\}.
\end{align*}
where $\dim\sigma=k$ and $\dim f=k-1$. These queries drive lazy enumeration.

\subsection{Persistent Partitions and Simplicial Towers}
A persistent partition satisfies $\mathcal P(r)\preceq\mathcal P(s)$ for $r\le s$. Here each block also has diameter at most $r$ in the fixed componentwise metric $d_H$. Edge insertions and block-parent maps induce a simplicial tower. The theory uses these maps directly; a coning conversion can provide a barcode-equivalent filtration representation \citep{ref7,ref8}.

\subsection{Outputs and Certification Scope}
CHARM returns low-dimensional homology computations, certified candidates in requested dimensions $k\ge3$, tracks, memberships, block genealogy, and shared coordinates $Y$. Exact $H_0,H_1,H_2$ results require complete enumeration and algebra; interrupted computations receive \texttt{INDETERMINATE-*} states. For a quotient block $C$ and representative $z$,
\begin{equation*}
\Pi(C)=\bigcup_{a\in C}B_a,\qquad
\operatorname{Supp}_X(z)=\bigcup_{C\in V(z)}\Pi(C).
\end{equation*}
This preimage localizes the selected quotient chain; it does not assert that the highlighted samples form an original-space representative chain.

\subsection{Incidence-two Candidates and Exact Low-dimensional Computation}
A connected incidence-two support is a nonempty set $Z$ of $k$-simplices, connected through shared $(k-1)$-faces, with $\deg_Z(f)=2$ for every participating face. The chain $\sum_{\sigma\in Z}\sigma$ is a cycle because each boundary face occurs twice. Triangle peeling discovers such a target if free-face removal isolates it as a dual component with all edge degrees two. Exact $H_2$ computation uses complete boundary matrices and does not depend on peeling success.

\subsection{Higher-dimensional Local Discoverability and Resource Conditions}
For $k\ge3$, conditional discovery completeness requires a selected seed on the target, exact enumeration of seed-incident simplices, and completion of the target-consistent branch within the resource budget. Theorem D.9 proves this conditional recovery statement for incidence-two targets, rather than all homology classes. Exact boundary, independence, and tracking decisions require complete incidence closure and current caches (Appendices B.9--B.11).

\section{Implementation Details}

\subsection{Locally Adaptive Gaussian Graph}

Choose $1\le k_{\mathrm{nn}}<n$, let $N_{k_{\mathrm{nn}}}(i)$ be the $k_{\mathrm{nn}}$ nearest neighbors of $x_i$ under the base distance, and use the adaptive local scale:\par

\begin{equation*}
h_i=d_0(x_i,x_{i,k_{\mathrm{nn}}}^{+}),
\end{equation*}

Here $x_{i,k_{\mathrm{nn}}}^{+}$ denotes the $k_{\mathrm{nn}}$th neighbor ranked by \textbf{positive} $d_0$-distance, assumed to exist. Thus $h_i>0$ even in the presence of a small number of exact duplicates.\par

On the original symmetric union $k_{\mathrm{nn}}$-NN graph $G_0$, construct the adaptive kernel\par

\begin{equation*}
K_{ij}=\begin{cases}
\exp\!\left(-\dfrac{d_0(x_i,x_j)^2}{2h_ih_j}\right),&\{i,j\}\in E(G_0),\\
0,&\text{otherwise}.
\end{cases}
\end{equation*}

Let $d_i=\sum_jK_{ij}$ and $P_{ij}=K_{ij}/{d_i}$. Symmetry of $K$ gives detailed balance $\pi_iP_{ij}=\pi_jP_{ji}$ for $\pi_i=d_i / {\sum_jd_j}$. We use this stationary measure to define the diffusion proposal geometry. The sparse eigensolver normalizes each component separately, and we compare diffusion coordinates within components. To obtain a nonnegative spectrum, define the lazy operator\par

\begin{equation*}
P_L=\frac12(I+P),
\end{equation*}

which we also denote by $P$ below. The core construction uses the adaptive kernel and stationary measure defined above.\par

\subsection{Sparse Diffusion Proposals and Topology-safe Cover Contraction}

Construct an adaptive Gaussian lazy operator separately on each connected component of the original union $k_{\mathrm{nn}}$-NN graph $G_0$ from Appendix B.1. Normalize right eigenfunctions in the component's $L^2(\pi)$ space, fix an integer diffusion time $t_{\mathrm{prop}}\ge1$, and use truncated coordinates\par

\begin{equation*}
\Phi_{\mathrm{prop}}(x_i)
=\bigl(\lambda_1^{t_{\mathrm{prop}}}\psi_1(i),\ldots,
\lambda_{s_D}^{t_{\mathrm{prop}}}\psi_{s_D}(i)\bigr).
\end{equation*}

We use these coordinates to propose covers within each connected component, with $s_D$ controlling spectral resolution. Proposition D.13 establishes fixed-graph smoothing and exact-spectrum truncation error. Cap the mode budget at one fewer than the component size; numerical eigensolver error is separate. Small components serve directly as candidate regions; if the eigensolver fails, we use graph neighborhoods.\par

Select local neighbor sets in diffusion coordinates and retain the part connected to each center in $G_0$, yielding overlapping candidate regions $\mathcal U=\{U_\alpha\}$ that cover all original samples. Candidate regions and their intersections may have arbitrary topology; the reference complex is $\operatorname{Flag}(G_0)$.\par

Let adjacent current vertices $u,v$ have original preimages $B_u,B_v$. Consider edges satisfying $B_u\cup B_v\subseteq U_\alpha$ and impose the following contraction condition on the current global graph. Let\par

\begin{equation*}
L=N(u)\setminus\bigl(N(v)\cup\{v\}\bigr),\qquad
R=N(v)\setminus\bigl(N(u)\cup\{u\}\bigr).
\end{equation*}

Accept the contraction when no edge connects $L$ to $R$, equivalently when the contracted edge belongs to no induced four-cycle. This condition prevents flag completion after contraction from filling a chordless quadrilateral. Successive accepted contractions yield a coarse graph $G_c$, an original-to-coarse projection, and preimages $B_a$ satisfying\par

\begin{equation*}
\operatorname{Flag}(G_c)\simeq\operatorname{Flag}(G_0).
\end{equation*}

Edgewise safety tests establish this homotopy equivalence. Theorems D.3 and D.11--D.12 compare the filtrations.\par

For local-rank estimates and block geometry, retain each coarse vertex's original Euclidean centroid and sample mass:\par

\begin{equation*}
c_a=\frac{1}{|B_a|}\sum_{i\in B_a}x_i,\qquad
\mu_a=\frac{|B_a|}{n}.
\end{equation*}

Here $\mu_a$ is original-sample mass; $\pi$ is the stationary measure used for the proposal diffusion spectrum.\par

\subsection{Edge Births from Coarse-graph Hop Distances}

The subsequent filtration uses shortest-hop distances within the coarse graph $G_c$. Write\par

\begin{equation*}
d_H(a,b)=\begin{cases}
\operatorname{dist}_{G_c}(a,b), & a,b\text{ connected},\\
+\infty, & \text{otherwise}.
\end{cases}
\end{equation*}

Define static edge-birth weights\par

\begin{align*}
b_{ab}&=d_H(a,b),\\
\mathcal R_H(r)&=\operatorname{Flag}\bigl(V_c,
\{\{a,b\}:b_{ab}\le r\}\bigr).
\end{align*}

The distance $d_H$ is a metric within each component, with components processed separately. Original coarse-graph edges appear at $r=1$, and each simplex appears when its last edge is born:\par

\begin{equation*}
b(\sigma)=\max_{\{a,b\}\subseteq\sigma}b_{ab}.
\end{equation*}

Scan finite hop births up to terminal scale $R_{\max}$; classes with unconfirmed deaths are right-censored. Let $X_c=\{B_a\}$ denote the coarse vertex set. Pairs and blocks in Appendices B.4--B.5 use this base set, with $\delta_r(a,b)=d_H(a,b)$.\par

The full hop filtration contains $O(|V_c|^2)$ pairs in the worst case.\par

\subsection{Dynamic Persistent Clique Partition}

At scale $r$, first process all raw edge events $b_{ij}\le r$ on the coarse vertex set $X_c$. Next, compute $\mathcal P(r)$ using the rule below, and then construct $Q(r)$ as in Appendix B.5. This order handles edge insertions and block contractions consistently when their scales coincide.\par

Define the complete-linkage event for blocks $A,B$ using edge-birth weights \citep{ref1}:\par

\begin{equation*}
\ell_b(A,B)
=
\max_{x_i\in A,\ x_j\in B}b_{ij}.
\end{equation*}

Complete linkage requires a symmetric event dissimilarity; the current choice $b_{ij}=d_H(i,j)$ also satisfies the triangle inequality within each component. Threshold equivalence with the current base metric $\delta_r$ gives the required block-diameter property.\par

For a terminal scale $R_{\max}$, every pair omitted in Appendix B.3 has $b_{ij}>R_{\max}$, including pairs in different components with infinite birth times. Such pairs cannot participate in linkage at a processed scale, so computing their exact birth times is unnecessary.\par

The definition of $b_{ij}$ gives $\ell_b(A,B)\le r \Longleftrightarrow \max_{x_i\in A,\ x_j\in B} \delta_r(x_i,x_j) \le r$. Merging blocks that satisfy complete linkage at scale $r$ therefore ensures\par

\begin{equation*}
\operatorname{diam}_{\delta_r}(C)\le r.
\end{equation*}

At every distinct finite birth value $r$, apply Leitão's conservative complete-linkage tie rule \citep{ref1} to current blocks by constructing the linkage graph\par

\begin{equation*}
A\sim_r B
\Longleftrightarrow
\ell_b(A,B)\le r,
\end{equation*}

Merge a connected component in its entirety exactly when it is a clique; otherwise retain its blocks at this scale. The partition remains constant between successive critical values. This resolves ties consistently and yields a nested partition:\par

\begin{equation*}
\mathcal P(r)
=
\{C_1(r),\ldots,C_{M(r)}(r)\}.
\end{equation*}

\subsection{Quotient Flag Tower}

At scale $r$, define for two quotient blocks in $\mathcal P(r)$\par

\begin{equation*}
\delta_r^Q(C_i,C_j)
=
\min_{x\in C_i,\ y\in C_j}
\delta_r(x,y).
\end{equation*}

Here $\delta_r^Q$ is a symmetric quotient dissimilarity. Quotient edges and flag completion use the threshold $\delta_r^Q\le r$; triangle inequalities in the error analysis apply to the componentwise base metric $d_H$.\par

Insert a quotient edge when $\delta_r^Q(C_i,C_j)\le r$. Write the quotient graph as $G_Q(r)=\bigl(\mathcal P(r),E_Q(r)\bigr),$ where $E_Q(r)=\{\{C,D\}\subseteq\mathcal P(r): \exists x\in C,\ y\in D \quad\text{s.t.}\ b_{xy}\le r\}$, and define\par

\begin{equation*}
\boxed{
Q(r)
=
\operatorname{Flag}(G_Q(r)).
}
\end{equation*}

As $r$ increases:\par

\begin{itemize}
\item hop-filtration edges are inserted;
\item partitions coarsen;
\item quotient graphs undergo both edge insertion and vertex contraction.
\end{itemize}

Proposition D.2 verifies that $\{Q(r)\}$ with the simplicial maps between successive scales forms a simplicial tower, along which historical representatives are pushed forward. Partition contractions during filtration evolution can cause homology classes to die.\par

Coning converts the tower into an equivalent filtration representation. Subsequent layout uses its active quotient $1$-skeleton.\par

\subsubsection{Neighborhood State of Active Vertices}

To reuse cofaces and incidence blocks incrementally, assign each active quotient vertex $v$ a neighborhood version $\eta(v)\in\mathbb N$. Whenever $N_Q(v)$ changes, increment $\eta(v)\leftarrow\eta(v)+1$.\par

\subsection{\texorpdfstring{$H_{k \le 2}$}{Hk, k <= 2}: Exact Low-dimensional Homology and Triangle Reuse}

Compute $H_0$ from graph connectivity and $H_1$ by reducing the graph cycle space modulo triangle boundaries, adding independent directions beyond historical representatives.\par

Compute $H_2$ on a homotopy-equivalent flag subcomplex using triangles and tetrahedra over $\mathbb F_2$:\par

\begin{equation*}
H_2=\ker\partial_2/\operatorname{im}\partial_3.
\end{equation*}

Transport historical representatives through tower maps and the explicit core retraction; core chains are quotient-complex chains by inclusion. Nonzero, independent images survive. Completed algebraic tests record survival or track termination; pending certification is marked unknown.\par

\subsubsection{Closed-surface Candidates from Incidence-two Peeling}
In the quotient triangle pool $\mathcal T(r)$, maintain the number of active triangles incident to each edge. Repeatedly remove a triangle incident to an edge of degree one, updating affected degrees. Partition the remaining triangles into dual components through shared edges and retain components whose participating edges all have degree two. These components are cycle candidates and still require non-boundary and independence certification. Peeling is a candidate extractor; complete matrix computation supplies the exact $H_2$ result even when no component passes this rule.

\subsection{\texorpdfstring{$H_{k\ge3}$}{Hk, k >= 3}: Seeding with Mass-weighted Local Rank}
Seed selection uses Euclidean block centroids and masses, independently of the diffusion coordinates used for cover proposals. For $C\in\mathcal P(r)$ and the closed quotient neighborhood $\mathcal N_r[v]$, define
\begin{align*}
m_C&=\sum_{a\in C}\mu_a,\\
c^Q(C)&=m_C^{-1}\sum_{a\in C}\mu_ac_a,\\
\bar c_v&=\frac{\sum_{u\in\mathcal N_r[v]}m_uc^Q(u)}
{\sum_{u\in\mathcal N_r[v]}m_u}.
\end{align*}
The rows of the centered matrix $M_v\in\mathbb R^{n_v\times D}$ are $\sqrt{m_u}(c^Q(u)-\bar c_v)^\top$, where $n_v=|\mathcal N_r[v]|$. The matrix $M_v^\top M_v$ is the mass-weighted scatter matrix. Retain $q_v=\min\{n_v-1,D\}$ singular values $\sigma_1\ge\cdots\ge\sigma_{q_v}\ge0$, including zeros if present. Vertices with $q_v\le1$ generate no seed; otherwise also exclude $\sigma_1=0$. Otherwise set
\begin{align*}
\widehat d(v)&=\arg\max_{1\le j<q_v}\frac{\sigma_j}{\sigma_{j+1}+\varepsilon_{\mathrm{mach}}\sigma_1},\\
R_{\widehat d}(v)&=\frac{\sum_{j>\widehat d(v)}\sigma_j^2}
{\sum_{j=1}^{q_v}\sigma_j^2}.
\end{align*}
\begin{equation*}
S_k(r)=\{v:\widehat d(v)\ge k,\ R_{\widehat d}(v)\le\varepsilon_{\dim}\}.
\end{equation*}
Resolve eigengap ties by the smallest index. The tail-energy threshold is a seed-selection heuristic, not a certificate of intrinsic dimension or homology. The rule implies $q_v>k$ for a selected seed. Recompute spectra when neighborhood membership, block centroids, or masses change; otherwise reuse them. Conditional discovery results assume that the target meets this seed set.

\subsection{Incidence-guided Lazy Search over \texorpdfstring{$k$}{k}-simplices}
For $k\ge3$, enumerate all $k$-simplices containing a vertex of $S_k(r)$, using $k$-cliques in its open quotient neighborhood. Initialize a support $Z$ from each unprocessed seed simplex. Maintain face degrees at most two and open faces $\mathcal O(Z)=\{f:\deg_Z(f)=1\}$. For each open face, common-neighbor queries from Appendix A.2 give all ambient incident simplices; feasible continuations are
\begin{align*}
A_Z(f;r)=\bigl\{&\eta\in\operatorname{Inc}_{Q,k}(f;r)\setminus Z:\\
&\deg_{Z\cup\{\eta\}}(g)\le2\text{ for each}\\
& (k-1)\text{-face }g\subset\eta\bigr\}.
\end{align*}
If $\mathcal O(Z)$ is empty, return the dual-connected incidence-two candidate. Otherwise reject the branch if some $A_Z(f;r)$ is empty, propagate any unique continuation, or branch over all continuations of an open face minimizing $|A_Z(f;r)|$. Canonically deduplicate identical support states. Exhausting the budget before search completion returns \texttt{INDETERMINATE-DISCOVERY}; candidate certification is separate.

\subsection{Cycle Certification and a Fast Non-boundary Test}
For an incidence-two support $Z$, $z_Z=\sum_{\sigma\in Z}\sigma$ satisfies $\partial_kz_Z=0$ by pairwise cancellation. Define $C_r(\sigma)=\bigcap_{v\in\sigma}N_Q(v;r)$. If some $\sigma\in\operatorname{supp}(z_Z)$ has $C_r(\sigma)=\varnothing$, it has no $(k+1)$-coface and hence cannot occur in any boundary. Thus $z_Z$ is non-boundary; otherwise use complete incidence closure below.

\subsubsection{Common-neighbor Cache}
Cache $C_r(\sigma)$ with signature $\operatorname{sig}_r(\sigma)=\{(v,\eta_r(v)):v\in\sigma\}$. Reuse requires the same active vertex identities and unchanged neighborhood versions (Appendix B.5), ensuring that all common-neighbor sets are current.

\subsection{Exact Non-boundary and Independence Certification}

For cycles not resolved by the fast test in Appendix B.9, we determine boundary membership by exact algebra. We then test non-boundary cycles for independence from the classes already accepted.\par

\subsubsection{Localization to Incidence Blocks}

Let $\mathfrak J_r$ be all connected components of the $k/(k+1)$-incidence bipartite graph of $Q(r)$. Denote discovered, completely closed cached blocks by $\mathcal J_r\subseteq\mathfrak J_r$. Every complete block $J=(F_k^J,F_{k+1}^J)$ satisfies\par

\begin{align*}
\{\sigma\subset\theta:\dim\sigma=k\}&\subseteq F_k^J
\quad (\theta\in F_{k+1}^J),\\
\{\theta\supset\sigma:\dim\theta=k+1\}&\subseteq F_{k+1}^J
\quad (\sigma\in F_k^J).
\end{align*}

Define $\mathcal B_J=\operatorname{span}\{\partial_{k+1}\theta:\theta\in F_{k+1}^J\}$. Distinct complete blocks have disjoint row coordinates, hence\par

\begin{equation*}
\operatorname{im}\partial_{k+1}=\bigoplus_{J\in\mathfrak J_r}\mathcal B_J.
\end{equation*}

Starting from $\operatorname{supp}(z)$, reuse valid complete blocks or recursively apply face-coface closure to a fixed point, obtaining all relevant blocks\par

\begin{equation*}
\mathfrak J(z)=\{J\in\mathfrak J_r:\operatorname{supp}(z)\cap F_k^J\ne\varnothing\}.
\end{equation*}

By the block decomposition,\par

\begin{equation*}
z\in\operatorname{im}\partial_{k+1}
\iff z\in\bigoplus_{J\in\mathfrak J(z)}\mathcal B_J.
\end{equation*}

Exact algebra on these complete blocks decides boundary membership and, for a boundary, can retain a filling chain. A $k$-simplex without a coface is an isolated row block. Reaching the resource budget before closure returns \texttt{INDETERMINATE-BOUNDARY}.\par

\subsubsection{Candidate Independence}

Let $z_1,\ldots,z_m$ be accepted independent representatives and $z$ a new candidate. Construct a representative-block bipartite graph, connecting representative node $i$ to block node $J$ whenever $\operatorname{supp}(z_i)$ intersects $F_k^J$.\par

Take the full transitive closure starting from $\mathfrak J(z)$, with block nodes $\mathfrak J^\star(z)$ and representative nodes $\mathcal I^\star(z)$, and let\par

\begin{equation*}
\mathcal B^\star(z)=\bigoplus_{J\in\mathfrak J^\star(z)}\mathcal B_J.
\end{equation*}

Then local and global independence tests are equivalent:\par

\begin{align*}
&[z]\in\operatorname{span}\{[z_1],\ldots,[z_m]\}\\
&\quad\Longleftrightarrow\quad
z\in\mathcal B^\star(z)+\operatorname{span}\{z_i:i\in\mathcal I^\star(z)\}.
\end{align*}

Complete representative-block closure makes this equivalence exact (Corollary D.7), using the boundary subspace $\mathcal B^\star(z)\subseteq B_k(Q)$ relevant to the candidate. A boundary candidate returns a filling relation; a dependent non-boundary candidate is deduplicated. A candidate that passes both tests enters the active basis with $b_{\mathrm{cert}}(z)=r$. If the independence closure is incomplete, retain the established non-boundary result and return \texttt{INDETERMINATE-INDEPENDENCE}.\par

\subsubsection{Boundary Echelon Basis Cache}

Cache an echelon basis for every complete block, with $\operatorname{span}\mathcal E_J=\mathcal B_J$, together with common-neighbor signatures of all its $k$-simplices. Valid caches can be reused at the same scale. Across scales, reuse requires all relevant vertices to remain active and all signatures unchanged; otherwise update the closure and basis (Corollary D.6).\par

\subsection{Persistence-compatible Basis Tracking}

Let $r\prec s$ be successive processed scales. Order active representatives by first certification scale and a fixed lexicographic order. Process historical tracks before certifying new candidates.\par

\subsubsection{Pushing Cycles Forward through the Chain Map}

Let $f_{r,s}:Q(r)\to Q(s)$ be the block-parent map. For each historical cycle, compute\par

\begin{equation*}
z_j(s)=(f_{r,s})_\#z_j(r).
\end{equation*}

A $k$-simplex with repeated image vertices has zero chain image in dimension $k$. Equal images of distinct simplices add over $\mathbb F_2$. Since $\partial f_\#=f_\#\partial$, pushed chains remain cycles. If the entire $z_j(s)=0$, the image class vanishes at $s$, terminating the track.\par

\subsubsection{Boundary and Independence Tests for Mapped Representatives}

For a nonzero image, perform complete boundary reduction, then compare with earlier representatives already confirmed to survive. Writing this set as $\mathcal A_{<j}(s)$, death or loss of an independent direction is characterized by\par

\begin{equation*}
z_j(s)\in B_k(Q(s))+\operatorname{span}\{z_i(s):i\in\mathcal A_{<j}(s)\}.
\end{equation*}

Representative-block closure from Appendix B.10 localizes this criterion exactly. If the image becomes a boundary, retain an available filling relation; if it depends on earlier directions, retain a dependence relation. Both terminate the independent track at $d_{\mathrm{cert}}=s$. Boundary membership certifies class vanishing; dependence only certifies loss of an independent direction, and its image class may remain nonzero. These events need not be endpoints of the full quotient barcode.\par

An incomplete relevant boundary or independence closure returns \texttt{INDETERMINATE-TRACKING}, leaving certified survival at the last confirmed scale.\par

\subsubsection{New Candidates and Censored Output}

After historical tracks are processed, certify new candidates against the confirmed surviving span. An unknown historical track is not a certified basis direction; independence from the confirmed span alone does not certify independence from all historical images. Candidates passing both tests receive a new $b_{\mathrm{cert}}$. A track with exactly confirmed termination outputs\par

\begin{equation*}
I_{\mathrm{cert}}=[b_{\mathrm{cert}},d_{\mathrm{cert}}).
\end{equation*}

Here $b_{\mathrm{cert}}$ is first certification and $d_{\mathrm{cert}}$ is independent-track termination; these need not be true barcode birth and death. Termination at the final scale or resource budget records the last exactly tested survival scale $r_{\mathrm{surv}}$ and right-censored status.\par

The certificates describe the quotient tower. Under the stated distance-distortion, component, and scale conditions, Theorems D.11--D.12 relate them to persistence in the original Euclidean space.\par

\subsection{Topology-aware Reverse-scale Geometry and Visualization}
The layout uses the quotient $1$-skeleton, block genealogy, and certified tracks; membership maps connect them to the original samples. Algebraic certification is independent of the drawing.

\subsubsection{Selecting the Root Scale from Certified Cycles}
For processed scales $\mathcal R_{\mathrm{obs}}$, let $\mathcal A_k(r)$ contain representatives exactly confirmed active after tracking. For a target dimension with at least one such record, choose
\begin{equation*}
r_{\mathrm{root}}=\max\{r\in\mathcal R_{\mathrm{obs}}:\mathcal A_k(r)\ne\varnothing\}.
\end{equation*}
Use the induced quotient graph on selected supports and context vertices. If no certified record is available, initialize from the graph skeleton.

\subsubsection{Reverse-scale Block Refinement Forest}
For displayed scales $r_0<\cdots<r_T=r_{\mathrm{root}}$, each fine block $C$ has a unique parent $p_t(C)$ at $r_{t+1}$. Scanning backward turns mergers into splits, producing a refinement forest. At each scale, draw $H_t=G_Q(r_t)[U_t]$ on the selected support and context vertices; parent links connect these sets across frames. Generator IDs and active states are read from forward certification records.

\subsubsection{Root Layout with Crowding Control}
Within quotient hop neighborhoods, define Gaussian affinities from block centroids $c^Q$, using a bandwidth selected by target perplexity or a local distance quantile. Normalize and symmetrize to obtain $P$. With Student-$t$ affinities $q_{uv}\propto(1+\|y_u-y_v\|^2)^{-1}$, minimize
\begin{equation*}
\mathcal L_{\mathrm{root}}=\lambda_{\mathrm{nb}}D_{\mathrm{KL}}(P\|Q)
+\lambda_{\mathrm{edge}}\mathcal L_{\mathrm{edge}}
+\lambda_{\mathrm{ov}}\mathcal L_{\mathrm{overlap}}.
\end{equation*}
The additional terms constrain quotient edges and penalize overlap. Start from deterministic spectral or graph-stress coordinates. Lay out disconnected components separately and pack them by mass or bounding radius.

\subsubsection{Initializing Children during Refinement}
Unsplit blocks inherit coordinates. Initialize a child $C$ near its parent by $y_C^{(t,0)}=y_{p_t(C)}^{(t+1)}+\varepsilon_tu_C$, with $\varepsilon_t$ smaller than local parent spacing. Use external neighbors' parent positions to choose $u_C$, or a canonical radial arrangement when none exist. Initialize all simultaneous splits before relaxation.

\subsubsection{Local Subtree Layout with Parent Anchors}
Relax new children, nearby quotient neighbors, and vertices with changed adjacency, freezing the remainder. For the affected set $\Omega_t$, minimize
\begin{align*}
\mathcal L_t&=\lambda_{\mathrm{nb}}D_{\mathrm{KL}}(P_t\|Q_t)
+\lambda_{\mathrm{edge}}\mathcal L_{\mathrm{edge},t}\\
&+\lambda_{\mathrm{par}}\sum_{C\in\Omega_t}m_C
\|y_C-y_{p_t(C)}^{(t+1)}\|^2\\
&+\lambda_{\mathrm{ov}}\mathcal L_{\mathrm{overlap},t}.
\end{align*}
Expand $\Omega_t$ when changed neighborhoods extend farther. Parent anchors limit drift during refinement.

\subsubsection{Localizing and Displaying Certified \texorpdfstring{$k$}{k}-cycles}
Display edges for $H_1$, triangle supports for $H_2$, and support skeletons with dimension labels for $k\ge3$. Original-sample coloring uses $\operatorname{Supp}_X(z)$. Reverse playback reproduces the recorded active range; it does not infer new birth or death events from the drawing.

\subsubsection{Relations between Features of Different Dimensions}
For the support subcomplex $K(z_k)$ generated by a certified chain, record containment of another certified representative when $\operatorname{supp}(w_{k-1})\subseteq K(z_k)_{k-1}$. These relations form a directed acyclic graph across dimensions. Each class retains its own certificate and track; containment alone supplies no additional homology certificate.

\subsubsection{Original-sample Localization and Shared Coordinates}
Optimize a single $Y\in\mathbb R^{n\times2}$ and obtain each frame by membership averaging, $Y_t=A_tY$, where $(A_t)_{Ci}=\mathbf1\{i\in\Pi(C)\}/|\Pi(C)|$. Reverse-scale layout initializes $Y$ and supplies skeleton constraints to
\begin{align*}
\mathcal L_{\mathrm{geom}}(Y)&=D_{\mathrm{KL}}(P^{\mathrm{aff}}\|Q^{\mathrm{aff}}(Y))\\
&+\lambda_{\mathrm{metric}}\mathcal L_{\mathrm{metric}}(Y)\\
&+\lambda_{\mathrm{rev}}\mathcal L_{\mathrm{reverse}}(Y).
\end{align*}
The terms match original neighborhood affinities, control pairwise-distance distortion, and constrain centroid edge lengths across frames. All displays and metrics use the final shared $Y$, including the fixed PH correction and leaf restoration in Appendix F.

\Needspace{15\baselineskip}
\section{Algorithm Pseudocode}
The pseudocode gives the main operations; Appendix B defines the tests, maps, and resource states. All algebra is over $\mathbb F_2$. An incomplete test returns the relevant \texttt{INDETERMINATE-*} state rather than an acceptance or death decision.

\subsection{Algorithm: CHARM Pipeline}
\begin{Verbatim}[fontsize=\small]
Input: samples X, graph/diffusion parameters, scales, dimensions, budgets.
1. Build G0; propose componentwise diffusion covers; compress via C.2.
2. For each hop event, perform steps 2--5: update edges and quotient Q.
3. Push and test historical tracks via C.5 before adding new candidates.
4. Compute H0--H2 via C.3; search requested higher dimensions via C.4.
5. Certify new candidates via C.5; store supports and tracking states.
6. Reverse the block hierarchy to initialize and optimize shared Y.
7. Apply the fixed postprocessing in F.1; export Y and certified records.
\end{Verbatim}

\subsection{Algorithm: Safe Compression and Hop Births}
\begin{Verbatim}[fontsize=\small]
1. Within proposed regions, contract a current edge only if its global
   exclusive-neighbor sets have no edge between them (B.2).
2. Retain memberships, centroids, and masses after every contraction.
3. Run componentwise truncated BFS; bucket pairs by their hop distance.
\end{Verbatim}

\subsection{Algorithm: Low-dimensional Homology}
\begin{Verbatim}[fontsize=\small]
1. Compute H0 by connectivity and H1 modulo triangle boundaries.
2. Compute H2 = ker(boundary_2)/im(boundary_3) on a safe flag core;
   transport representatives through the stored chain maps.
3. Optionally extract incidence-two triangle candidates by peeling.
\end{Verbatim}

\subsection{Algorithm: Higher-dimensional Candidate Search}
\begin{Verbatim}[fontsize=\small]
1. Enumerate simplices incident to local-rank seeds; initialize supports.
2. Maintain face degrees <= 2; output a support when every degree is 2.
3. At an open face, prune if no feasible continuation exists, propagate
   a unique continuation, or branch over all feasible continuations.
\end{Verbatim}

\subsection{Algorithm: Certification and Tracking}
\begin{Verbatim}[fontsize=\small]
1. Verify the cycle equation; certify non-boundary by an exposed simplex
   or exact reduction on complete incidence blocks.
2. Test independence on the complete representative--block closure.
3. For historical images, process earlier certifications first; record
   boundary/dependence relations or survival. Insert new directions last.
4. Record confirmed endpoints; censor at the last confirmed survival
   when tracking stops. Preserve unknown status after an incomplete test.
\end{Verbatim}

\Needspace{14\baselineskip}
\section{Theoretical Statements and Proofs}

We establish the theoretical guarantees for CHARM in five steps: preservation of the base homotopy type, approximation of the coarse-graph filtration by the quotient tower, comparison with original Euclidean Rips under distance distortion, exact local algebraic certification, and discovery completeness under explicit conditions. We then analyze diffusion smoothing on a fixed graph.\par

\subsection{Theorem: Topology-safe Contraction Preserves the Base Homotopy Type}

Let $uv$ be an edge of the current graph $G$, with exclusive-neighbor sets $L,R$ as in Appendix B.2. If no edge joins $L$ to $R$, the contraction map induces
\begin{equation*}
\operatorname{Flag}(G)\simeq\operatorname{Flag}(G/uv).
\end{equation*}

\textbf{Proof.} Let $p:K=\operatorname{Flag}(G)\to K'=\operatorname{Flag}(G/uv)$ identify $u,v$ with $w$. For a target simplex $\tau$, let $K_\tau$ be the subcomplex of simplices mapped into $\tau$. If $w\notin\tau$, $K_\tau$ is a simplex. Otherwise, the vertices of $\tau\setminus\{w\}$ cannot include both a vertex in $L$ and one in $R$: the clique condition would require a forbidden edge. They are therefore all adjacent to $u$ or all to $v$. That endpoint is adjacent to every other vertex of $K_\tau$, so the flag property makes $K_\tau$ a cone. Thus every $K_\tau$ is contractible, and these subcomplexes are nested under face inclusion.

Construct a continuous inverse $g:|K'|\to|K|$ inductively over simplices, taking $g(|\tau|)\subseteq|K_\tau|$; contractibility extends each boundary map. Then $pg$ is homotopic to the identity inside each target simplex. Likewise, on each source simplex $\sigma$, both $gp$ and the identity take values in the contractible carrier $|K_{p(\sigma)}|$; extending homotopies inductively over simplex prisms makes them homotopic. Hence $p$ is a homotopy equivalence. Applying this argument after each accepted contraction proves the claim for $G_0\to G_c$. $\square$\par

\subsection{Proposition: Hop Filtration, Nested Partitions, and Tower Maps}

Within each coarse connected component, shortest-hop distance defines a metric and
\begin{align*}
r\le s&\implies\mathcal R_H(r)\subseteq\mathcal R_H(s),\\
&\mathcal P(r)\preceq\mathcal P(s),\\
&\operatorname{diam}_{d_H}(C)\le r\quad (C\in\mathcal P(r)).
\end{align*}
The block-parent maps define a simplicial tower $f_{r,s}:Q(r)\to Q(s)$.

\textbf{Proof.} Increasing the threshold adds edges. Partitions begin with singletons and only merge. At scale $r$, a clique in the complete-linkage graph has all cross-block distances at most $r$; together with the inherited within-block bound, its union has diameter at most $r$. A quotient edge's witness pair remains valid at every later scale, so its parent blocks coincide or remain adjacent. Flag completion extends this vertex map simplicially, and unique parents give $f_{s,t}f_{r,s}=f_{r,t}$. Truncated breadth-first search includes every edge birth through $R_{\max}$; infinite intercomponent distances create neither edges nor mergers. $\square$\par

\subsection{Theorem: Multiplicative 3-interleaving Relative to the Coarse Hop Filtration}

Extend the hop filtration and partition rule to all positive scales. In every homological dimension, $H_k(Q)$ and $H_k(\mathcal R_H)$ are multiplicatively $3$-interleaved, and their logarithmic bottleneck distance is at most $\log3$.

\textbf{Proof.} Block projection gives $p_r:\mathcal R_H(r)\to Q(r)$. Select $a_C\in C$ for each block. For an edge $CD$ with witnesses $x\in C,y\in D$,
\begin{equation*}
d_H(a_C,a_D)\le d_H(a_C,x)+d_H(x,y)+d_H(y,a_D)\le3r.
\end{equation*}
Flag completion therefore gives $q_r:Q(r)\to\mathcal R_H(3r)$. For $r\le s$, selections made before and after the parent map are contiguous at scale $3s$: points in the same parent block have distance at most $s$, and points in distinct parent blocks have distance at most $3s$ through a witness pair. This also proves independence of representative choices on homology.

Write $P_r=(p_r)_*$ and $S_r=(q_r)_*$. They are natural maps with shifts $1$ and $3$. Moreover, $q_rp_r$ is contiguous to the inclusion at $3r$, and $p_{3r}q_r$ is the parent map. Hence $S_rP_r$ and $P_{3r}S_r$ equal the respective structure maps $r\to3r$. Advancing $P_r$ to $Q(3r)$ yields two shift-$3$ maps whose composites are the structure maps $r\to9r$, as required. The finite complexes change at finitely many scales, so algebraic stability applies after logarithmic reparameterization \citep{ref11}. $\square$\par

This comparison concerns the coarse hop filtration; Theorems D.11--D.12 address original Euclidean Rips.

\subsection{Proposition: An Exposed Simplex Certifies Non-boundary Status}

If a cycle $z$ contains a $k$-simplex $\sigma$ without a $(k+1)$-coface, then $[z]\ne0$.

\textbf{Proof.} The $\sigma$ row of $\partial_{k+1}$ is zero, whereas $z_\sigma=1$. Thus $\partial_{k+1}a=z$ has no solution. $\square$\par

\subsection{Theorem: Exact Incidence-Block Localization}

Let $\mathfrak J_r$ contain all components of the $k/(k+1)$-incidence bipartite graph, including isolated $k$-simplices. For $J=(F_k^J,F_{k+1}^J)$, set $\mathcal B_J=\operatorname{span}\{\partial_{k+1}\theta:\theta\in F_{k+1}^J\}$, and let $\mathfrak J(z)$ contain the blocks intersecting $\operatorname{supp}(z)$. Then
\begin{align*}
\operatorname{im}\partial_{k+1}&=\bigoplus_{J\in\mathfrak J_r}\mathcal B_J,\\
z\in\operatorname{im}\partial_{k+1}&\iff
z\in\bigoplus_{J\in\mathfrak J(z)}\mathcal B_J.
\end{align*}

\textbf{Proof.} Every coface and all its boundary faces lie in one block, so the boundary matrix is block diagonal. Distinct blocks have disjoint row coordinates. A filling's components outside $\mathfrak J(z)$ have zero boundary and can be discarded; conversely, any filling using the indicated blocks is a global filling. $\square$\par

The algorithm obtains these blocks by recursive face-coface closure. Exhausting the budget before closure cannot certify non-boundary status by this test.

\subsection{Corollary: Echelon Bases and Cache Validity}

For a complete echelon basis $\mathcal E_J$ of $\mathcal B_J$,
\begin{equation*}
z|_{F_k^J}\in\mathcal B_J
\ \Longleftrightarrow\ 
\operatorname{Reduce}(z|_{F_k^J};\mathcal E_J)=0.
\end{equation*}
Blocks can be reduced independently. A cached complete block remains valid when all its vertices remain active and the neighborhood signatures of every row simplex remain unchanged.

\textbf{Proof.} The reduction criterion is exact span membership, and distinct blocks have disjoint coordinates. Unchanged signatures preserve all common-neighbor sets, hence every row's complete coface list. The closed block therefore retains the same rows, columns, and boundary relations. Changed signatures or vertex identities require revalidation. $\square$\par

\subsection{Corollary: Exact Independence Testing by Representative--Block Closure}

In the bipartite graph joining incidence blocks to accepted representatives $z_1,\ldots,z_m$ through support intersection, take the full transitive closure from $\mathfrak J(z)$. Denote its block and representative sets by $\mathfrak J^\star(z),\mathcal I^\star(z)$ and set $\mathcal B^\star(z)=\bigoplus_{J\in\mathfrak J^\star(z)}\mathcal B_J$. Then
\begin{align*}
&z\in\operatorname{im}\partial_{k+1}+\operatorname{span}\{z_1,\ldots,z_m\}\\
&\quad\Longleftrightarrow\quad
z\in\mathcal B^\star(z)+\operatorname{span}\{z_i:i\in\mathcal I^\star(z)\}.
\end{align*}

\textbf{Proof.} Project a global relation onto the closure's row coordinates. Outside boundary blocks vanish. Each representative meeting the closure is included with its entire support by transitivity; all others vanish. This gives the local relation. Conversely, a local relation has zero coordinates outside the closure and is already global. $\square$\par

\subsection{Theorem: Algebraic Soundness of Accepted Witnesses}

A verified cycle passing the exposed-simplex or complete-block non-boundary test represents a nonzero class in $H_k(Q(r);\mathbb F_2)$. Passing complete representative-block independence testing additionally certifies independence from the accepted span.

\textbf{Proof.} The tests establish $z\in\ker\partial_k$, $z\notin\operatorname{im}\partial_{k+1}$ (Proposition D.4 or Theorem D.5), and, when required, nonmembership in the boundary-plus-representative span (Corollary D.7). These exact algebraic conditions do not depend on the accuracy of proposal coordinates or local-rank estimates. $\square$\par

\subsection{Theorem: Conditional Recovery of Incidence-two Supports}

At a fixed scale, let $Z^\star$ be a finite connected incidence-two support. Assume all relevant incidence queries are exact. For $k\ge3$, assume that a selected seed lies in the target, all seed-incident simplices are enumerated, and the search branch consistent with $Z^\star$ is explored to completion. Algorithm C.4 then recovers $Z^\star$. For $k=2$, optional peeling in Algorithm C.3 recovers it under the peelability condition of Appendix A.5.

\textbf{Proof.} Peeling cannot remove the first target triangle: each of its edges still has two target cofaces. The peelability assumption then isolates the target as a degree-two component. For $k\ge3$, begin with an enumerated target simplex and maintain $Y\subseteq Z^\star$. At each open face, the other target simplex is feasible because every target face has degree two. Forced propagation selects it when unique; otherwise the appropriate branch selects it. Dual connectivity gives an open face in every nonempty proper subset of $Z^\star$, preventing premature closure. Finiteness and the exploration assumption complete the recovery. $\square$\par

If the required certification closures also complete, Theorem D.5 and Corollary D.7 determine whether the recovered cycle is non-boundary and independent. A recovered support may be a boundary. This conditional statement guarantees neither seed coverage nor an incidence-two representative for every homology class; the full $H_2$ computation instead uses $\ker\partial_2/\operatorname{im}\partial_3$.

\subsection{Corollary: Certified Ranges and Persistent Intervals}

Suppose an element accepted at $b_{\mathrm{cert}}$ has exactly verified nonzero forward images throughout $[b_{\mathrm{cert}},d_{\mathrm{cert}})$. Then some quotient-barcode interval contains this range. For a right-censored record surviving at $r_{\mathrm{surv}}$, some interval contains $[b_{\mathrm{cert}},r_{\mathrm{surv}}]$. These are containment statements: first certification need not be birth, and termination through dependence on earlier representatives need not be the death of the tracked element.\par

\textbf{Proof.} Decompose the element at certification into the finitely many interval summands active there. All are born by $b_{\mathrm{cert}}$. If every participating summand died before $d_{\mathrm{cert}}$, their images would all vanish before that endpoint. Hence at least one contains the asserted half-open range. Nonzero image at $r_{\mathrm{surv}}$ likewise requires a participating summand whose death is strictly later. $\square$\par

Theorem D.3 implies that a confirmed range with positive start and termination/start ratio greater than $9$, or last-survival/start ratio greater than $9$, forces a nonzero interval in the coarse-hop filtration: the containing quotient interval is too long to match the diagonal within logarithmic cost $\log3$. For original Euclidean Rips, use Theorem D.12 with rescaled certification start $\ell b_{\mathrm{cert}}\ge r_{\min}$ and ratio greater than $c_Q^2$. If the containing quotient bar was born earlier than $r_{\min}$, first truncate its birth to $r_{\min}$; it still contains the confirmed range.\par

\subsection{Theorem: Additive Error between Original Euclidean Rips and the Coarse Graph}

Assume $G_0$ is connected, $q:X\twoheadrightarrow V_c$ is the membership projection, and $\widetilde d_H=\ell d_H$ for $\ell>0$. Define
\begin{equation*}
\varepsilon_\ell=\max_{i,j}\left|\|x_i-x_j\|-\widetilde d_H(q(i),q(j))\right|.
\end{equation*}
For $\varepsilon\ge\varepsilon_\ell$ and every $k\ge0$, $H_k(\mathcal R_X)$ and $H_k(\widetilde{\mathcal R}_H)$ are additively $\varepsilon$-interleaved, so their bottleneck distance is at most $\varepsilon$. The bound in (2) satisfies $\varepsilon_\ell\le2\rho+\eta_\ell$.\par

\textbf{Proof.} Choose a fixed section $s:V_c\to X$ with $s(a)\in B_a$. The distortion bound gives simplicial maps
\begin{align*}
q_r&:\mathcal R_X(r)\longrightarrow
\widetilde{\mathcal R}_H(r+\varepsilon),\\
s_r&:\widetilde{\mathcal R}_H(r)\longrightarrow
\mathcal R_X(r+\varepsilon).
\end{align*}
Their fixed vertex maps commute with inclusions. Since $qs=\mathrm{id}$, the coarse-side composite is the structure map. For $\sigma\in\mathcal R_X(r)$ and $u,v\in\sigma\cup s(q(\sigma))$,
\begin{equation*}
\|u-v\|\le\widetilde d_H(q(u),q(v))+\varepsilon\le r+2\varepsilon.
\end{equation*}
Thus the original-side composite is contiguous to inclusion into $\mathcal R_X(r+2\varepsilon)$ and induces the same homology map. Algebraic stability gives the bottleneck bound \citep{ref11,ref20}. Finally, for $i\in B_a,j\in B_b$,
\begin{equation*}
\left|\|x_i-x_j\|-\|c_a-c_b\|\right|
\le\|x_i-c_a\|+\|x_j-c_b\|\le2\rho.
\end{equation*}
Adding the centroid--hop distortion $\eta_\ell$ proves (2), including $a=b$. $\square$\par

\textbf{Computing the bound.} Computing $\rho$ costs $O(nD)$. For $m=|V_c|$, all-pairs BFS and centroid distances compute $\eta_\ell$ in $O(m(m+|E_c|)+m^2D)$ time; a proven upper bound covering all coarse pairs also suffices.\par

\subsection{Theorem: Composite Guarantee for Original Euclidean Rips and the Cover Quotient Tower}

Under Theorem D.11's assumptions, extend $\widetilde Q(r)=Q(r/\ell)$ using the same partition rule at every $r\ge0$. Then $H_k(\mathcal R_X)$ and $H_k(\widetilde Q)$ are interleaved by $\phi(r)=3r+\varepsilon$: the two composites are structure maps to $\phi^2(r)=9r+4\varepsilon$. On $[r_{\min},\infty)$, $r_{\min}>0$, the original/coarse and original/quotient modules are multiplicatively interleaved with
\begin{equation*}
c_H=1+\varepsilon/r_{\min},\qquad c_Q=3+\varepsilon/r_{\min},
\end{equation*}
respectively. The corresponding logarithmic bottleneck bounds are $\log c_H$ and $\log c_Q$.\par

\textbf{Proof.} Compose Theorem D.11's projection with block projection to obtain
\begin{equation*}
F_r:\mathcal R_X(r)\longrightarrow\widetilde Q(r+\varepsilon).
\end{equation*}
For each block $C$ of $\widetilde Q(r)$, choose an original member $u_C\in\Pi(C)$. Distinct blocks $C,D$ in a quotient simplex have a witness edge of length at most $r$, and each block has $\widetilde d_H$-diameter at most $r$. Consequently, for arbitrary $u\in\Pi(C),v\in\Pi(D)$,
\begin{equation*}
\widetilde d_H(q(u),q(v))\le3r,\qquad \|u-v\|\le3r+\varepsilon.
\end{equation*}
The same-block bound is $r+\varepsilon$. Representative selection therefore defines $G_r:\widetilde Q(r)\to\mathcal R_X(\phi(r))$. For $r\le t$, selecting representatives before or after taking parent blocks gives contiguous maps into $\mathcal R_X(\phi(t))$, by the same diameter bound. Thus $G$ is natural on homology; $F$ is natural by the parent relations.\par

Advance $F_r$ to $\widetilde Q(\phi(r))$. The original-side composite chooses representatives of the source vertices' blocks at $\phi(r)$. Its images and source simplex together have diameter at most $3\phi(r)+\varepsilon=\phi^2(r)$, proving contiguity to inclusion. The quotient-side composite sends each block to its unique parent at $\phi^2(r)$, exactly the structure map. This proves the $\phi$-interleaving.\par

For $r\ge r_{\min}$, $r+\varepsilon\le c_Hr$ and $\phi(r)\le c_Qr$. Advancing the maps to these larger targets preserves the composite identities by naturality. Logarithmic reparameterization and algebraic stability yield the stated bottleneck bounds \citep{ref11}. Restriction replaces each earlier birth by $r_{\min}$. $\square$\par

\textbf{Interval interpretation.} Let $[b,d)$ be a finite bar in the \emph{restricted} rescaled quotient module. If $d/b>c_Q^2$, it cannot match the diagonal within cost $\log c_Q$ and must match an original-Rips bar $[b_X,d_X)$ satisfying
\begin{equation*}
c_Q^{-1}b\le b_X\le c_Qb,\qquad c_Q^{-1}d\le d_X\le c_Qd.
\end{equation*}
These are bounds on restricted endpoints. If $b_X=r_{\min}$, the original birth may be earlier; if $b>c_Qr_{\min}$, both matched births are untruncated. Essential bars match essential bars, with the same finite-birth bound. A certified range starting at or after $r_{\min}$ with ratio above $c_Q^2$ lies in a sufficiently long restricted bar by Corollary D.10. Certification alone does not identify that bar's exact endpoints.\par

\textbf{Scope.} These barcode bounds concern full extended filtrations. For disconnected $G_0$, they apply to the disjoint union of the componentwise Euclidean Rips filtrations. This union equals full Euclidean Rips only below the minimum intercomponent Euclidean distance $g$. A finite run, or a comparison below $g$, certifies only maps whose source, target, and composite scales are available. For example, a confirmed quotient element from $b\ge r_{\min}$ through $s>c_Q^2b$ forces a nonzero original-Rips map from $c_Qb$ to $s/c_Q$, provided all required scales are available and below $g$ when needed: the nonzero quotient map from $b$ to $s$ factors through this map by the interleaving identities. Such local information does not imply a global barcode matching or uncensored endpoint bounds.\par

The positive lower cutoff is essential: contracting two points at distance $\delta>0$ removes their finite $H_0$ bar while preserving the contractible base flag complex. A multiplicative $c$-interleaving at all positive scales would make the rank-two original map from $r$ to $c^2r<\delta$ factor through the rank-one coarse $H_0$, which is impossible.\par

\subsection{Proposition: Diffusion Smoothing and Truncation on a Fixed Graph}

On a finite connected component with $N$ vertices, let $P_L$ be a reversible lazy Markov operator with stationary distribution $\pi_i>0$, eigenvalues $1=\lambda_0>\lambda_1\ge\cdots\ge\lambda_{N-1}\ge0$, and right eigenfunctions $\psi_a$ orthonormal in $L^2(\pi)$, with $\psi_0=1$. For integer $t\ge1$,
\begin{align*}
D_t^2(i,j)&:=\sum_u\frac{(P_L^t(i,u)-P_L^t(j,u))^2}{\pi_u}\\
&=\sum_{a=1}^{N-1}\lambda_a^{2t}
(\psi_a(i)-\psi_a(j))^2.
\end{align*}
Retaining $0\le s<N-1$ nontrivial modes gives
\begin{equation*}
0\le D_t^2(i,j)-\|\Phi_t^{(s)}(i)-\Phi_t^{(s)}(j)\|^2
\le\lambda_{s+1}^{2t}(\pi_i^{-1}+\pi_j^{-1}).
\end{equation*}
If all modes are retained, the error is zero. For $s<N-1$, a signal $f$ supported only on discarded modes satisfies $\|P_L^tf\|_{L^2(\pi)}\le\lambda_{s+1}^t\|f\|_{L^2(\pi)}$.\par

\textbf{Proof.} Reversibility gives $P_L^t(i,u)/\pi_u=\sum_a\lambda_a^t\psi_a(i)\psi_a(u)$. Substitution and orthogonality give the spectral identity \citep{ref2,ref22}. For $i\ne j$, completeness yields $\sum_{a\ge1}(\psi_a(i)-\psi_a(j))^2=\pi_i^{-1}+\pi_j^{-1}$; for $i=j$ it is zero. Bounding each tail eigenvalue by $\lambda_{s+1}$ proves truncation. Parseval's identity gives signal attenuation. $\square$\par

These statements concern exact eigenpairs of the fixed graph. They do not bound numerical eigensolver error or guarantee robustness to arbitrary input noise; attenuation helps proposals when noise is concentrated in discarded modes.\par

\section{Complexity Analysis}

Let $n$ be the number of original samples, $D$ their dimension, and $m_G=O(nk_{\mathrm{nn}})$ the number of original graph edges. Write $n_c,m_c$ for the coarse-graph vertex and edge counts, and $m_R$ for the number of hop pairs within the terminal scale. At scale $r$, let $M(r),m_Q(r),T_Q(r),U_Q(r)$ count quotient vertices, edges, triangles, and tetrahedra. The bounds below account for the structures constructed, visited, and processed to completion.\par

\subsection{Original Graph, Diffusion Proposals, and Safe Contraction}

Let $\mathcal C_{\mathrm{kNN}}$ denote the cost of constructing the original neighbor graph. Given adjacency and distances, computing the Gaussian kernel and its normalization takes $O(m_G)$ time and space. With $I$ iterations, the matrix multiplications in sparse spectral computation cost $O(I m_G s_D)$, in addition to orthogonalization and eigenpair-solution costs. Storing the retained coordinates takes $O(ns_D)$ space. The proposal mode budget $s_D$ is selected independently of the requested homological dimension.\par

Diffusion-space neighbor queries, connected-region extraction, and cover storage depend on actual candidate sizes. With adjacency hash sets supporting expected constant-time membership queries, the $e$th safety test checks for edges between $L_e,R_e$ in expected $O(\sum_{u\in L_e}\min\{|N(u)|,|R_e|\})$ time by iterating over smaller sets. Additional costs include constructing exclusive-neighbor sets, testing preimage containment, and updating adjacency. Candidate overlap and repeated tests contribute further work. Sufficient statistics for original centroids take $O(nD)$ to aggregate.\par

\subsection{Coarse-graph Hop Edge Births}

Run truncated BFS from each coarse vertex. If $V_a^{(R)},E_a^{(R)}$ are the vertices visited and edges scanned from $a$ within radius $R_{\max}$, total traversal cost is\par

\begin{equation*}
O\!\left(\sum_{a\in V_c}(|V_a^{(R)}|+|E_a^{(R)}|)\right),
\end{equation*}

The full-range worst case is $O(n_c(n_c+m_c))$. Storing returned pairs takes $O(m_R)$ space, with $m_R=\Theta(n_c^2)$ in the worst case. Integer births can be bucketed by hop value; comparison sorting costs $O(m_R\log m_R)$.\par

\subsection{Persistent Partitions and Quotient Towers}

The costs of complete-linkage updates, linkage-component tests, and quotient-adjacency maintenance depend on the returned pairs, merge history, and cumulative edge and contraction events. We quantify compression through the quotient vertex and simplex counts at each scale, together with the cumulative update work.\par

\subsection{Exact \texorpdfstring{$H_0,H_1,H_2$}{H0, H1, H2}}

Dynamic connectivity information supports the computation of $H_0$. Computing $H_1$ uses graph cycles and triangle boundaries, while $H_2$ requires triangles, tetrahedra, and exact kernel and image computations for $\partial_2,\partial_3$. Homotopy cores reduce the computational input; local clique density determines $T_Q(r),U_Q(r)$. Across scales, cost is measured by cumulative simplex visits, tower maps, and elimination work.\par

For direct enumeration, a complete graph on $v$ vertices contains $\sum_{j=1}^{\min(k+2,v)}\binom{v}{j}$ simplices through dimension $(k+1)$; for example, $H_2$ boundary tests may involve $\binom{v}{4}$ tetrahedra. Coarsening and quotienting reduce actual work through active-vertex counts and relevant incidence blocks.\par

For a materialized triangle pool, incidence updates in one free-face peeling pass take $O(T_Q(r))$ work; an implicit pool additionally incurs enumeration costs.\par

\subsection{Mass-weighted Local Rank}

For a vertex with closed-neighborhood size $n_v$, constructing the weighted centered matrix takes $O(n_vD)$ time. A conservative bound for direct thin SVD is\par

\begin{equation*}
O\!\left(n_vD\min\{n_v,D\}\right).
\end{equation*}

This step provides seeds for $k\ge3$ searches and uses the original Euclidean dimension $D$. Sufficient statistics update block masses and centroids in $O(D)$ time per merge. Local matrices are recomputed when their adjacency or member centroids change; spectra are reused for unchanged neighborhoods.\par

\subsection{Higher-dimensional Search and Exact Local Certification}

Let $N_{k,j}^{\mathrm{vis}}$ be simplex-state visits in search $j$, and $\mathcal C_{\cap,j}$ the associated common-neighbor query cost. The principal query work is\par

\begin{equation*}
O\!\left(\sum_j N_{k,j}^{\mathrm{vis}}(k+1)\mathcal C_{\cap,j}\right),
\end{equation*}

Additional work includes seed-clique and continuation enumeration, state copying, deduplication, and incidence updates. Persistent ambiguity can produce exponentially many search branches.\par

For a complete certification-block matrix with $R_j$ rows and $C_j$ columns, a conservative bound for naive dense elimination is $O(R_jC_j\min\{R_j,C_j\})$ time and $O(R_jC_j)$ space. Sparse and bitset implementations can reduce the practical cost. Independence tests add columns for relevant representatives, and valid echelon caches can be reused. In the worst case, closure reaches the full complex. If the budget is reached, the algorithm returns the appropriate unknown state for discovery, boundary, independence, or tracking.\par

\subsection{Reverse-scale Layout and Shared Coordinates}

The reverse-scale layout cost depends on the number of displayed frames, the active-graph sizes, and the local optimization iterations. Sparse membership-averaging matrices $A_t$ aggregate through preimages in $O(n)$ time per frame for a complete partition. The shared coordinates $Y$ require $O(n)$ space. Dense all-pairs affinity or repulsion calculations cost $O(n^2)$ per iteration. Pairwise distance computation and PH postprocessing contribute to the cost of the complete pipeline.\par

\section{Complete Experimental Protocol and Supplementary Results}

This appendix documents the archived results in \texttt{paper/evaluation} as of \textbf{November 6, 2026, 17:58:40 (Beijing time)}, using the evaluation overview, per-dataset report, and homology support viewer. The main text highlights the principal results across three evaluation categories. This appendix reports all 15 sample-weighted metrics for the full collection and for the nonsynthetic and synthetic groups. Appendix G contains 110 method-result rows across 22 datasets, each with 15 metrics.\par

\subsection{Datasets and Training Protocol}

The evaluation includes 22 datasets and 41,637 samples: 15,637 samples in 9 nonsynthetic datasets and 26,000 samples in 13 synthetic datasets. All methods use the archived inputs in their original row order. The following table gives evaluated sample counts.\par

\begin{center}
\small\setlength{\tabcolsep}{4pt}
\begin{tabularx}{\linewidth}{lXX}
\toprule
Dataset & Group & Samples \\
\midrule
digits & Nonsynthetic & 1,797 \\
mnist & Nonsynthetic & 2,000 \\
fashion\_mnist & Nonsynthetic & 2,000 \\
coil20 & Nonsynthetic & 1,440 \\
olivetti & Nonsynthetic & 400 \\
coil100 & Nonsynthetic & 2,000 \\
mouse\_scrna & Nonsynthetic & 2,000 \\
googlenews & Nonsynthetic & 2,000 \\
flow\_cytometry & Nonsynthetic & 2,000 \\
intersecting\_spheres & Synthetic & 2,000 \\
klein\_bottle & Synthetic & 2,000 \\
multiple\_spheres & Synthetic & 2,000 \\
nonuniform\_sphere & Synthetic & 2,000 \\
orthogonal\_group\_3 & Synthetic & 2,000 \\
projective\_plane & Synthetic & 2,000 \\
s\_curve & Synthetic & 2,000 \\
sphere & Synthetic & 2,000 \\
sphere\_and\_filled\_ball & Synthetic & 2,000 \\
sphere\_times\_circle & Synthetic & 2,000 \\
swiss\_roll & Synthetic & 2,000 \\
torus2 & Synthetic & 2,000 \\
torus3 & Synthetic & 2,000 \\
\bottomrule
\end{tabularx}
\end{center}

The table below specifies each training configuration. All methods use random seed 0. Evaluation sampling uses separate fixed seeds shared across methods.\par

\begin{center}
\small\setlength{\tabcolsep}{4pt}
\begin{tabularx}{\linewidth}{lX}
\toprule
Method & Training configuration \\
\midrule
CHARM & \texttt{knn\_diffusion\_cover}; graph $k=50$; cover diffusion time 4, mode budget 32, region-size cap 512; single-scale $P=50$; at most 600 shared geometric optimization iterations; Euclidean training-distance constraints; at most 3 PH correction steps and leaf restoration \\
t-SNE & $P=50$, at most 600 iterations \\
UMAP & $k=50$, 600 epochs, \texttt{min\_dist=0.1} \\
Isomap & $k=50$, Dijkstra shortest paths, dense eigensolver \\
PCA & Full SVD; no neighborhood-scale parameter; shares the PCA fit with the k15 branch of the same experiment \\
\bottomrule
\end{tabularx}
\end{center}

Isomap encounters disconnected training graphs for \texttt{coil20}, \texttt{multiple\_spheres}, \texttt{orthogonal\_group\_3}, and \texttt{sphere\_and\_filled\_ball} and uses its default component-connection mechanism. G-family evaluation below uses an independently constructed original k15 graph without added edges and excludes unreachable pairs.\par

\subsection{Evaluation Conventions and the 15 Metrics}

For every method, local neighborhood scores compare original-space and two-dimensional Euclidean distances over all input samples, with $k=50$. G-family metrics use the original all-sample union k15 graph with Euclidean edge weights. Evaluation samples pairs/triplets from all rows and computes the required exact graph shortest paths. The evaluator uses 50,000 pairs and 5 sets of 20,000 triplets each.\par

PH evaluation uses Euclidean VR complexes on the full point clouds over $\mathbb F_2$, computing $H_0,H_1$. The original cloud and every embedding are independently centered and RMS-normalized:\par

\begin{align*}
\widetilde x_i&=\frac{x_i-\bar x}{\sqrt{n^{-1}\sum_j\|x_j-\bar x\|^2}},\\
\widetilde y_i&=\frac{y_i-\bar y}{\sqrt{n^{-1}\sum_j\|y_j-\bar y\|^2}}.
\end{align*}

These PH scores compare shapes after normalizing the overall scale of each cloud. Bottleneck and order-2 Wasserstein distances use $L^\infty$ pointwise costs and diagonal matching; $W_2$ is not divided by the diagram size. CHARM's 15 metrics, monochrome scatterplots, and support colors use the same final $Y$.\par

\begin{center}
\footnotesize\setlength{\tabcolsep}{2pt}\renewcommand{\arraystretch}{0.92}
\begin{tabularx}{\linewidth}{@{}l c X@{}}
\toprule
Metric & Direction & Meaning \\
\midrule
T50 & $\uparrow$ & Trustworthiness: penalizes false neighbors. \\
C50 & $\uparrow$ & Continuity: penalizes neighbors lost in 2D. \\
Q50 & $\uparrow$ & Fraction of retained 50-neighbor pairs. \\
RNX50 & $\uparrow$ & Chance-corrected 50-neighbor overlap. \\
RNX\_AUC\_log & $\uparrow$ & Log-weighted AUC of adjusted overlap. \\
RNX\_global & $\uparrow$ & Adjusted overlap at relative sizes 10--50\%. \\
RTA\_G & $\uparrow$ & Graph-geodesic triplet-order accuracy. \\
rho\_G & $\uparrow$ & Spearman correlation of geodesic distances. \\
pearson\_G & $\uparrow$ & Pearson correlation of geodesic distances. \\
stress\_G & $\downarrow$ & Relative stress after global scale alignment. \\
RV\_G & $\downarrow$ & Geodesic residual variance: $1-\mathrm{pearson}_G^2$. \\
H0\_dB & $\downarrow$ & $H_0$ persistence-diagram bottleneck distance. \\
H0\_W2 & $\downarrow$ & $H_0$ diagram order-2 Wasserstein distance. \\
H1\_dB & $\downarrow$ & $H_1$ persistence-diagram bottleneck distance. \\
H1\_W2 & $\downarrow$ & $H_1$ diagram order-2 Wasserstein distance. \\
\bottomrule
\end{tabularx}
\end{center}

Neighborhood overlap and its correction for random overlap are defined as\par

\begin{align*}
Q_{NX}(k)&=\frac{1}{nk}\sum_i|N_k^X(i)\cap N_k^Y(i)|,\\
R_{NX}(k)&=\frac{(n-1)Q_{NX}(k)-k}{n-1-k}.
\end{align*}

\texttt{RNX\_AUC\_log} aggregates $R_{NX}(k)$ over $k=1,\ldots,n-2$ with weights $1/k$. \texttt{RNX\_global} averages values over $k/(n-1)\in[0.1,0.5]$. For each dataset, \texttt{RV\_G} equals $1-\mathrm{pearson}_G^2$; group-level RV and Pearson values are sample-weighted separately.\par

Group aggregation evaluates each dataset independently and then weights by its actual sample count:\par

\begin{equation*}
\overline M_{\mathcal D}(m)=
\frac{\sum_{d\in\mathcal D}n_d M_d(m)}{\sum_{d\in\mathcal D}n_d}.
\end{equation*}

Values are displayed to six decimal places; \textbf{bold indicates the best} and \underline{underlining the second best}. Rankings use unrounded values and tolerance $10^{-10}$: rank equals one plus the number of strictly better methods, and numerical ties share a rank.\par

\subsection{Sample-weighted Group Results}

\subsubsection{All Data: 22 Datasets, 41,637 Samples}

\begin{center}
\small\setlength{\tabcolsep}{4pt}
\begin{adjustbox}{max width=\linewidth}
\begin{tabular}{@{}lrrrrr@{}}
\toprule
Metric & CHARM & t-SNE & UMAP & Isomap & PCA \\
\midrule
T50 \(\uparrow\) & 0.961093 & \textbf{0.967636} & \underline{0.962655} & 0.853403 & 0.851939 \\
C50 \(\uparrow\) & \textbf{0.955756} & 0.950764 & \underline{0.954248} & 0.950800 & 0.942362 \\
Q50 \(\uparrow\) & \underline{0.650620} & \textbf{0.664755} & 0.638602 & 0.385501 & 0.368305 \\
RNX50 \(\uparrow\) & \underline{0.641030} & \textbf{0.655549} & 0.628742 & 0.368916 & 0.351245 \\
RNX\_AUC\_log \(\uparrow\) & \underline{0.561491} & \textbf{0.607781} & 0.516560 & 0.369980 & 0.361207 \\
RNX\_global \(\uparrow\) & \underline{0.563502} & 0.526877 & 0.536464 & 0.550061 & \textbf{0.564919} \\
\midrule
RTA\_G \(\uparrow\) & \underline{0.775532} & 0.767427 & 0.773681 & \textbf{0.783225} & 0.763374 \\
rho\_G \(\uparrow\) & \underline{0.700730} & 0.678865 & 0.684133 & \textbf{0.710378} & 0.686749 \\
pearson\_G \(\uparrow\) & \underline{0.699129} & 0.678484 & 0.674917 & \textbf{0.714548} & 0.692952 \\
stress\_G \(\downarrow\) & \textbf{0.308088} & 0.325245 & 0.334540 & \underline{0.312149} & 0.327113 \\
RV\_G \(\downarrow\) & 0.481438 & 0.514002 & 0.513109 & \textbf{0.441757} & \underline{0.475654} \\
\midrule
H0\_dB \(\downarrow\) & \textbf{0.325450} & 0.379915 & 0.386242 & 0.355021 & \underline{0.341532} \\
H0\_W2 \(\downarrow\) & \textbf{6.099576} & 6.245110 & 6.262154 & 6.161252 & \underline{6.146908} \\
H1\_dB \(\downarrow\) & \underline{0.195031} & 0.208424 & 0.218139 & 0.198971 & \textbf{0.189897} \\
H1\_W2 \(\downarrow\) & \textbf{0.728950} & 0.762859 & 0.766165 & 0.755962 & \underline{0.753322} \\
\bottomrule
\end{tabular}
\end{adjustbox}
\end{center}

\subsubsection{Nonsynthetic Data: 9 Datasets, 15,637 Samples}

\begin{center}
\small\setlength{\tabcolsep}{4pt}
\begin{adjustbox}{max width=\linewidth}
\begin{tabular}{@{}lrrrrr@{}}
\toprule
Metric & CHARM & t-SNE & UMAP & Isomap & PCA \\
\midrule
T50 \(\uparrow\) & \textbf{0.932820} & \underline{0.931321} & 0.927676 & 0.839815 & 0.823711 \\
C50 \(\uparrow\) & \textbf{0.940222} & \underline{0.935928} & 0.934815 & 0.923977 & 0.899779 \\
Q50 \(\uparrow\) & \underline{0.533032} & \textbf{0.534062} & 0.514725 & 0.298785 & 0.274950 \\
RNX50 \(\uparrow\) & \underline{0.519382} & \textbf{0.520494} & 0.500709 & 0.278611 & 0.254076 \\
RNX\_AUC\_log \(\uparrow\) & \underline{0.478862} & \textbf{0.479139} & 0.368367 & 0.270208 & 0.259412 \\
RNX\_global \(\uparrow\) & \underline{0.495896} & 0.473621 & 0.444586 & \textbf{0.503305} & 0.489068 \\
\midrule
RTA\_G \(\uparrow\) & \underline{0.767870} & 0.754928 & 0.749070 & \textbf{0.781931} & 0.729116 \\
rho\_G \(\uparrow\) & \underline{0.716945} & 0.680635 & 0.664285 & \textbf{0.735979} & 0.639076 \\
pearson\_G \(\uparrow\) & \underline{0.724352} & 0.678457 & 0.660385 & \textbf{0.744504} & 0.655294 \\
stress\_G \(\downarrow\) & \textbf{0.301757} & 0.335772 & 0.372863 & \underline{0.321520} & 0.379699 \\
RV\_G \(\downarrow\) & \underline{0.455981} & 0.522396 & 0.546374 & \textbf{0.421839} & 0.533218 \\
\midrule
H0\_dB \(\downarrow\) & \textbf{0.668314} & 0.753144 & 0.723239 & 0.749747 & \underline{0.713917} \\
H0\_W2 \(\downarrow\) & \textbf{13.266261} & 13.354466 & 13.331322 & 13.334842 & \underline{13.300012} \\
H1\_dB \(\downarrow\) & \textbf{0.102801} & 0.105936 & 0.114668 & 0.105515 & \underline{0.104980} \\
H1\_W2 \(\downarrow\) & \textbf{0.753020} & 0.776184 & \underline{0.770062} & 0.780641 & 0.787859 \\
\bottomrule
\end{tabular}
\end{adjustbox}
\end{center}

\subsubsection{Synthetic Data: 13 Datasets, 26,000 Samples}

\begin{center}
\small\setlength{\tabcolsep}{4pt}
\begin{adjustbox}{max width=\linewidth}
\begin{tabular}{@{}lrrrrr@{}}
\toprule
Metric & CHARM & t-SNE & UMAP & Isomap & PCA \\
\midrule
T50 \(\uparrow\) & 0.978097 & \textbf{0.989477} & \underline{0.983692} & 0.861575 & 0.868916 \\
C50 \(\uparrow\) & 0.965099 & 0.959687 & 0.965936 & \underline{0.966932} & \textbf{0.967973} \\
Q50 \(\uparrow\) & \underline{0.721340} & \textbf{0.743358} & 0.713105 & 0.437654 & 0.424450 \\
RNX50 \(\uparrow\) & \underline{0.714191} & \textbf{0.736774} & 0.705745 & 0.423227 & 0.409685 \\
RNX\_AUC\_log \(\uparrow\) & \underline{0.611187} & \textbf{0.685150} & 0.605687 & 0.429985 & 0.422428 \\
RNX\_global \(\uparrow\) & \underline{0.604162} & 0.558906 & 0.591721 & 0.578182 & \textbf{0.610537} \\
\midrule
RTA\_G \(\uparrow\) & 0.780140 & 0.774944 & \textbf{0.788482} & \underline{0.784003} & 0.783978 \\
rho\_G \(\uparrow\) & 0.690978 & 0.677801 & \underline{0.696071} & 0.694981 & \textbf{0.715421} \\
pearson\_G \(\uparrow\) & 0.683960 & 0.678500 & 0.683658 & \underline{0.696531} & \textbf{0.715600} \\
stress\_G \(\downarrow\) & 0.311896 & 0.318913 & 0.311491 & \underline{0.306513} & \textbf{0.295487} \\
RV\_G \(\downarrow\) & 0.496749 & 0.508953 & 0.493102 & \underline{0.453735} & \textbf{0.441033} \\
\midrule
H0\_dB \(\downarrow\) & 0.119244 & 0.155447 & 0.183564 & \underline{0.117624} & \textbf{0.117571} \\
H0\_W2 \(\downarrow\) & \textbf{1.789366} & 1.969379 & 2.010593 & 1.846890 & \underline{1.844867} \\
H1\_dB \(\downarrow\) & \underline{0.250500} & 0.270063 & 0.280369 & 0.255177 & \textbf{0.240968} \\
H1\_W2 \(\downarrow\) & \textbf{0.714474} & 0.754846 & 0.763822 & 0.741119 & \underline{0.732550} \\
\bottomrule
\end{tabular}
\end{adjustbox}
\end{center}

\subsubsection{Interpretation}

CHARM ranks first on seven and second on eight of the 15 sample-weighted metrics for nonsynthetic data. Across all datasets, it ranks among the top two on 13 metrics. The table below reports first- and second-place counts for every method and group.\par

\begin{center}
\small\setlength{\tabcolsep}{4pt}
\begin{adjustbox}{max width=\linewidth}
\begin{adjustbox}{max width=\linewidth}
\begin{tabular}{lrrr}
\toprule
Method & All & Nonsynthetic & Synthetic \\
\midrule
CHARM & 5 / 8 & 7 / 8 & 2 / 5 \\
t-SNE & 4 / 0 & 3 / 2 & 4 / 0 \\
UMAP & 0 / 2 & 0 / 1 & 1 / 2 \\
Isomap & 4 / 1 & 5 / 1 & 0 / 6 \\
PCA & 2 / 4 & 0 / 3 & 8 / 2 \\
\bottomrule
\end{tabular}
\end{adjustbox}
\end{adjustbox}
\end{center}

Sections 4.2--4.3 discuss group results, and Appendix G provides per-dataset results.\par

\subsection{Two-dimensional Point-cloud Comparisons}

Each comparison grid contains six columns: CHARM + supports, CHARM, t-SNE, UMAP, Isomap, and PCA. For display, each embedding is centered and isotropically RMS-normalized; each row shares the t-SNE axis limits from column three. Out-of-view counts are listed in Appendix G; evaluation uses all samples.\par

\begin{figure}[p]

\centering

\includegraphics[width=\linewidth]{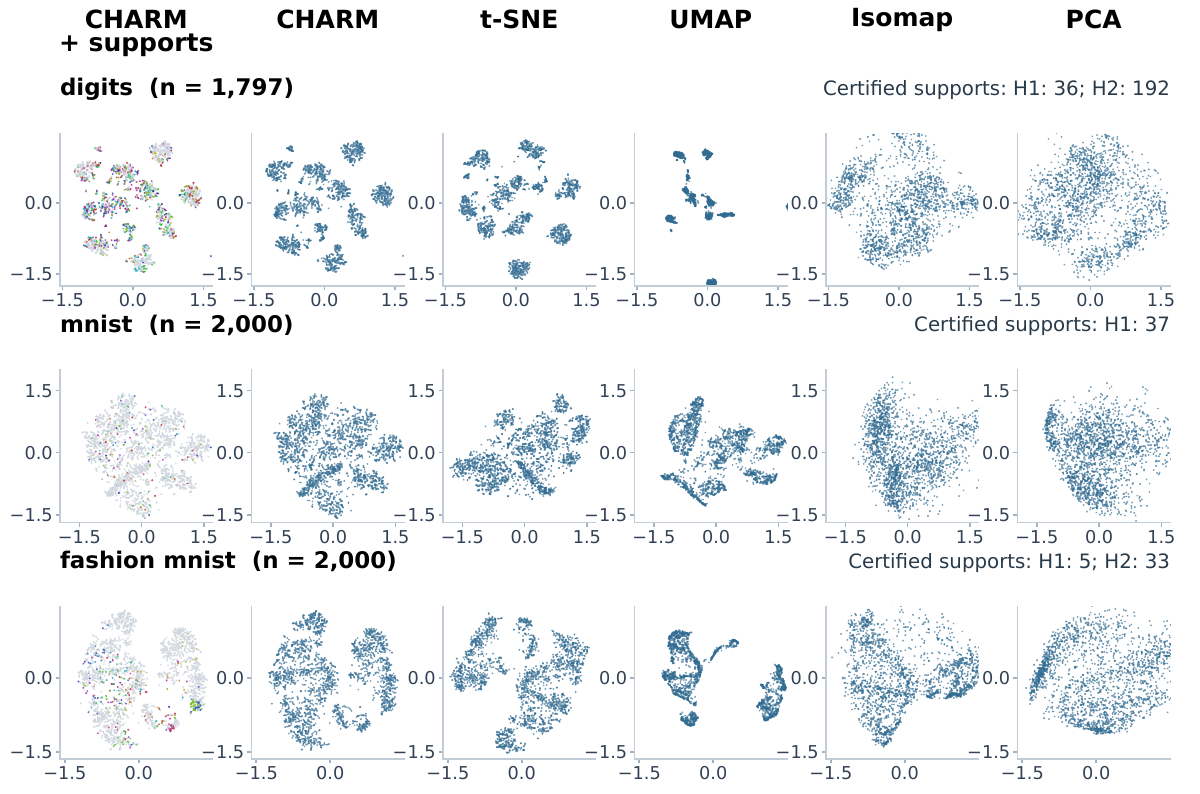}

\caption*{Figure F.1. Nine nonsynthetic datasets, with six views per row. The first two columns share identical CHARM coordinates. \href{figures/pointcloud_non_synthetic.svg}{Vector version}. Part 1/3.}

\end{figure}

\begin{figure}[p]

\centering

\includegraphics[width=\linewidth]{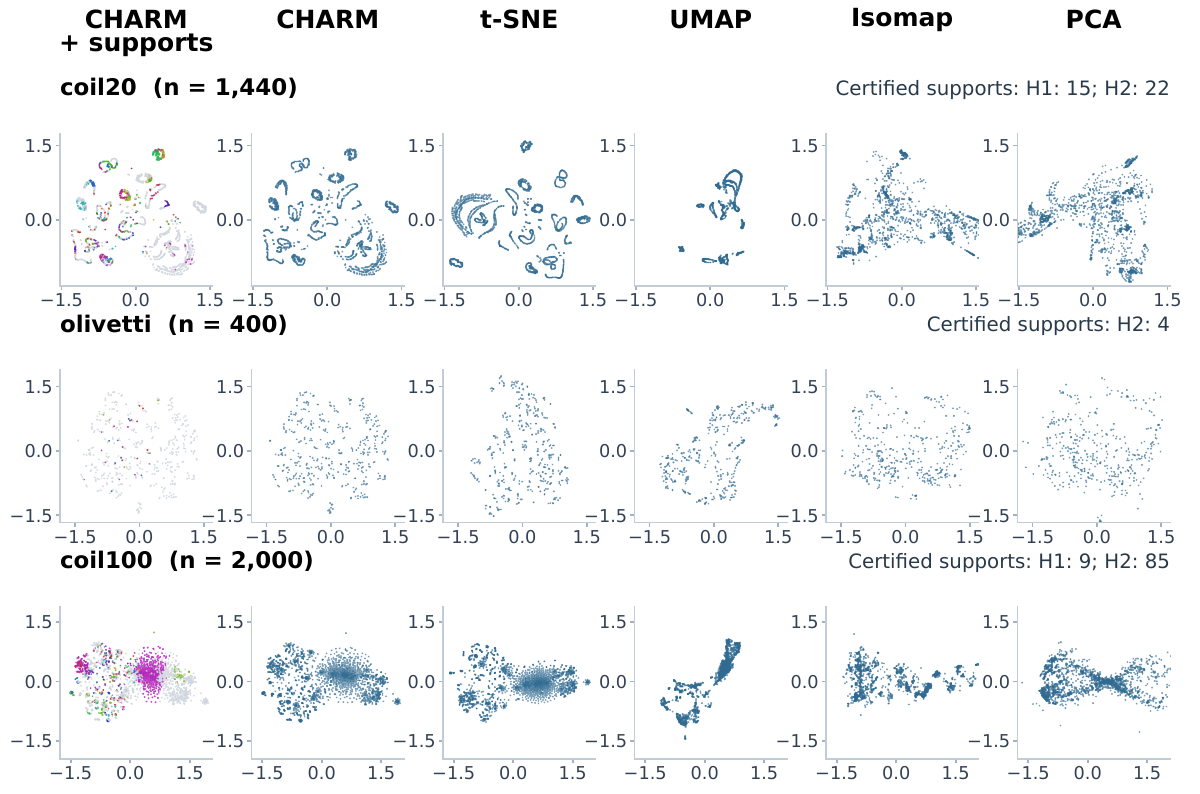}

\caption*{Figure F.1. Nine nonsynthetic datasets, with six views per row. The first two columns share identical CHARM coordinates. \href{figures/pointcloud_non_synthetic.svg}{Vector version}. Part 2/3.}

\end{figure}

\begin{figure}[p]

\centering

\includegraphics[width=\linewidth]{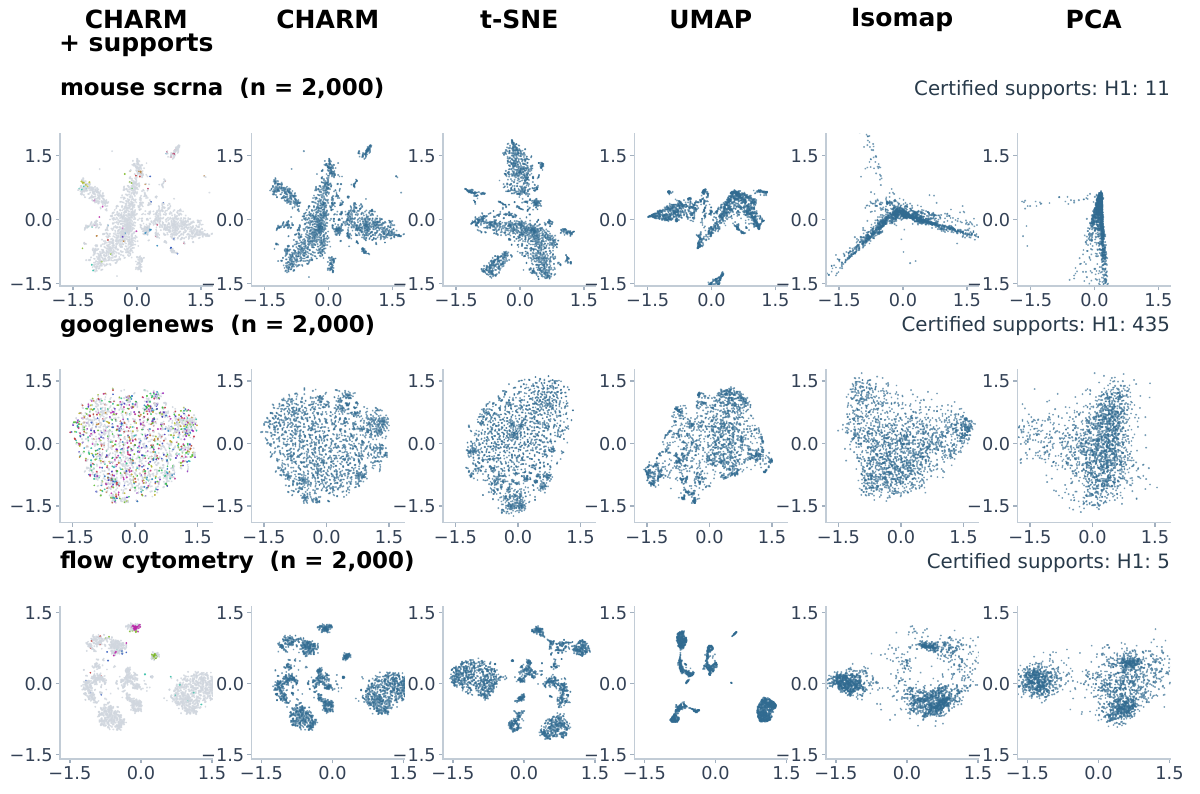}

\caption*{Figure F.1. Nine nonsynthetic datasets, with six views per row. The first two columns share identical CHARM coordinates. \href{figures/pointcloud_non_synthetic.svg}{Vector version}. Part 3/3.}

\end{figure}

\clearpage

\begin{figure}[p]

\centering

\includegraphics[width=\linewidth]{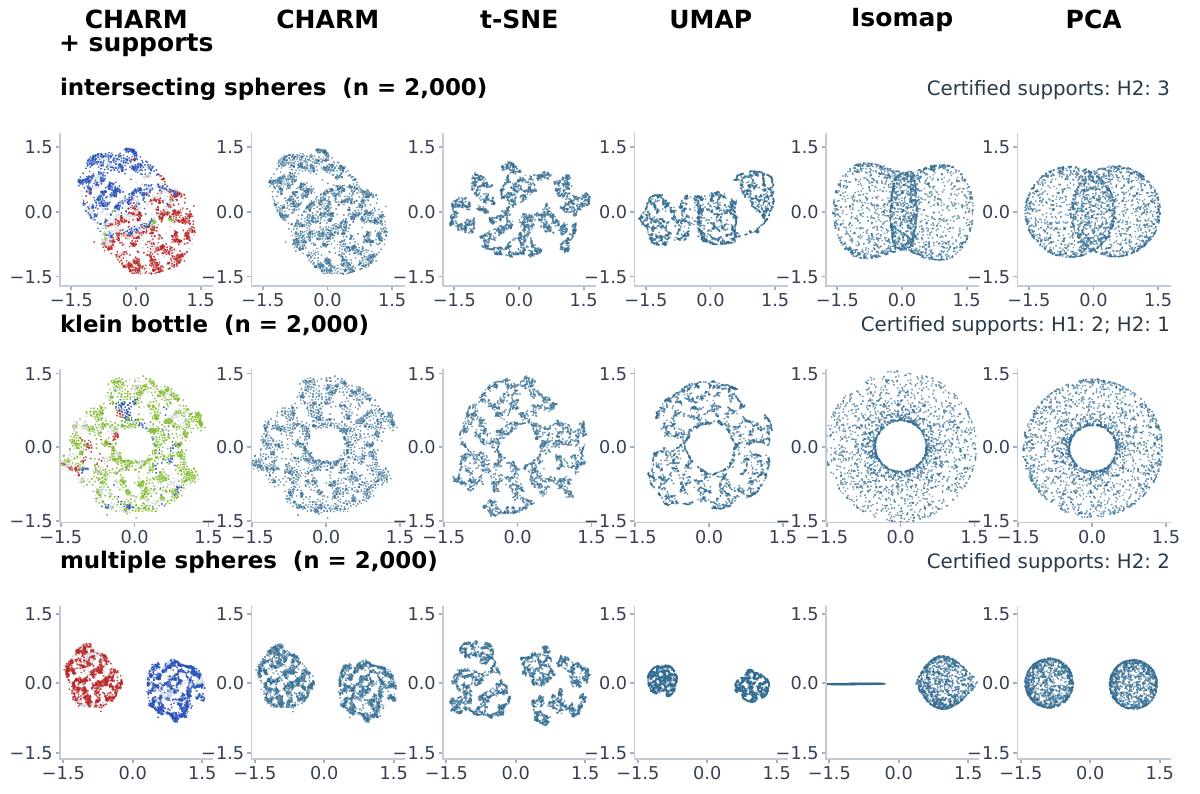}

\caption*{Figure F.2. Thirteen synthetic datasets, using the column order and normalization of Figure F.1. \href{figures/pointcloud_synthetic.svg}{Vector version}. Part 1/5.}

\end{figure}

\begin{figure}[p]

\centering

\includegraphics[width=\linewidth]{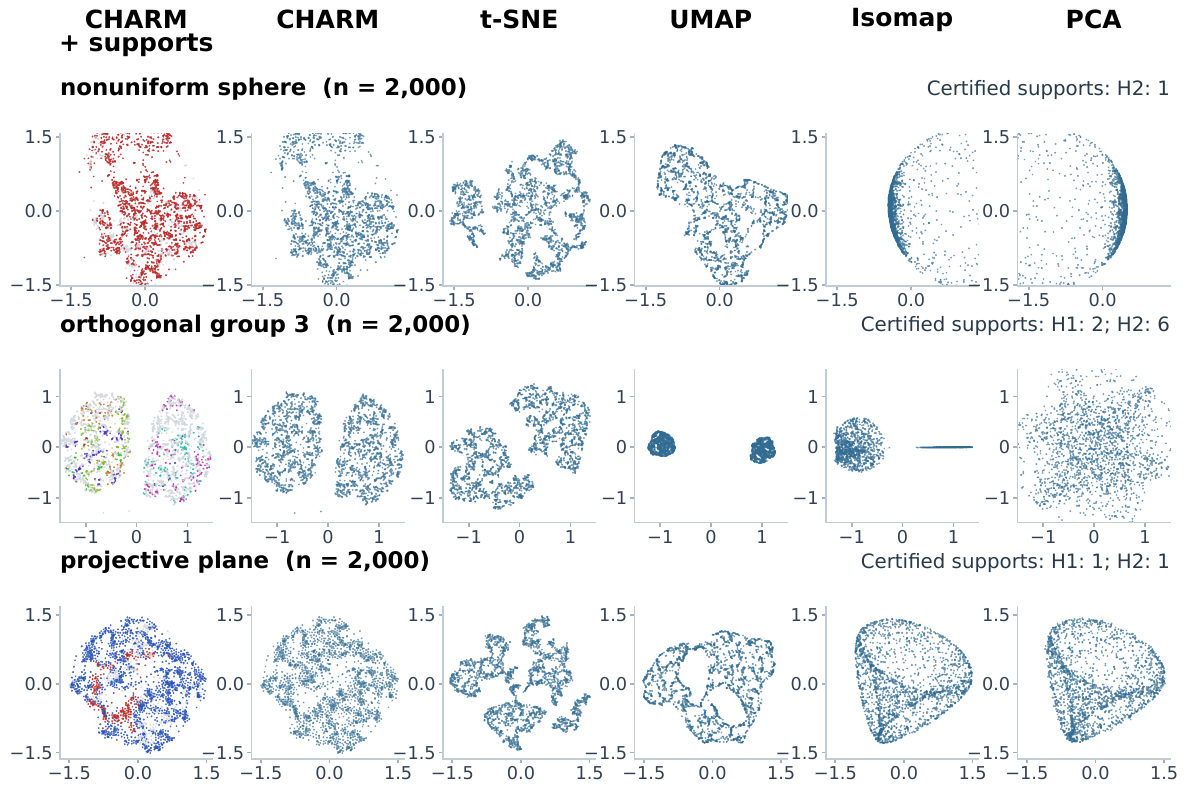}

\caption*{Figure F.2. Thirteen synthetic datasets, using the column order and normalization of Figure F.1. \href{figures/pointcloud_synthetic.svg}{Vector version}. Part 2/5.}

\end{figure}

\begin{figure}[p]

\centering

\includegraphics[width=\linewidth]{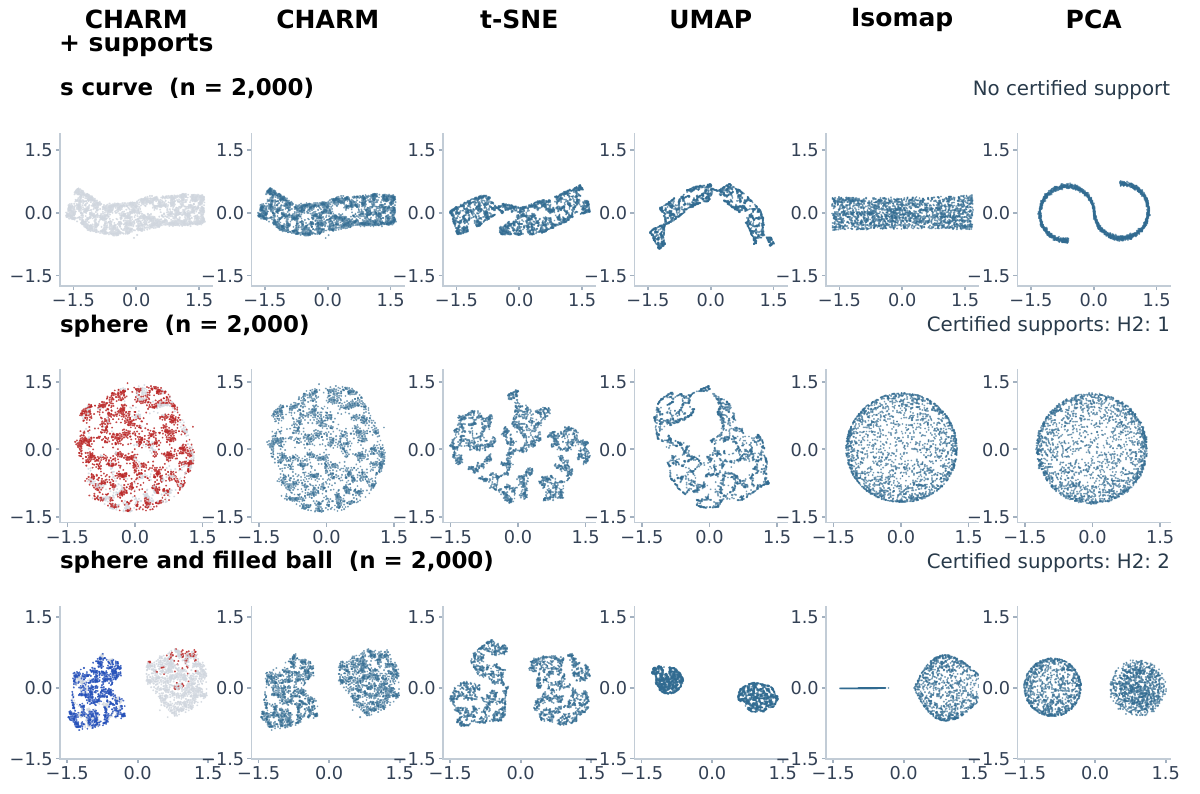}

\caption*{Figure F.2. Thirteen synthetic datasets, using the column order and normalization of Figure F.1. \href{figures/pointcloud_synthetic.svg}{Vector version}. Part 3/5.}

\end{figure}

\begin{figure}[p]

\centering

\includegraphics[width=\linewidth]{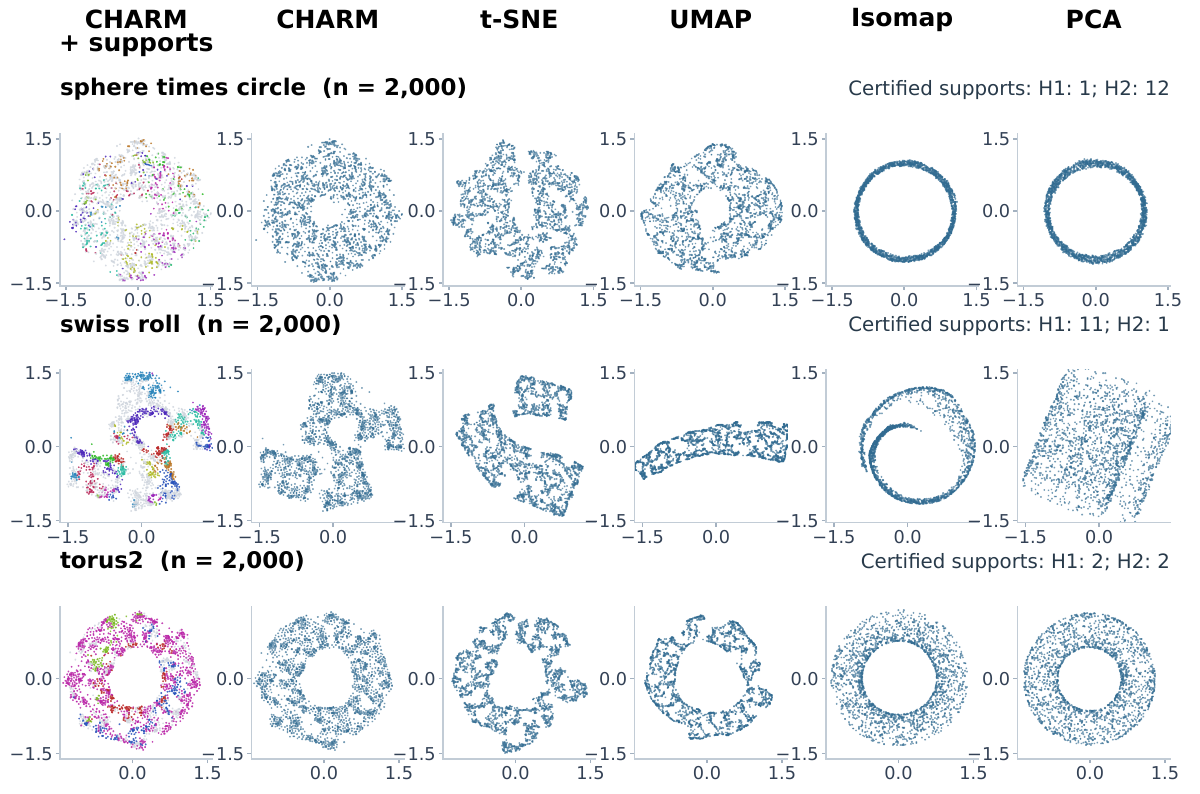}

\caption*{Figure F.2. Thirteen synthetic datasets, using the column order and normalization of Figure F.1. \href{figures/pointcloud_synthetic.svg}{Vector version}. Part 4/5.}

\end{figure}

\begin{figure}[p]

\centering

\includegraphics[width=\linewidth]{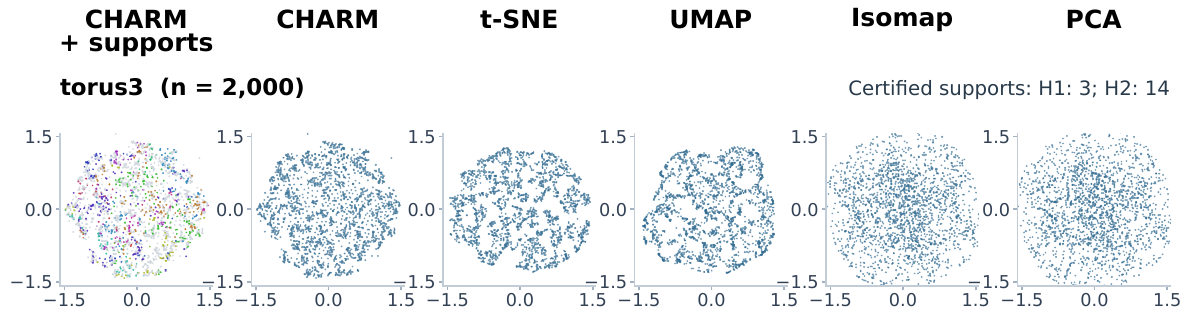}

\caption*{Figure F.2. Thirteen synthetic datasets, using the column order and normalization of Figure F.1. \href{figures/pointcloud_synthetic.svg}{Vector version}. Part 5/5.}

\end{figure}

\clearpage

\subsection{Homology Supports and Member Localization}

The support viewer stores 957 support records across 22 datasets, with complete sample memberships. The table reports dimensions, coverage, and right-censoring. Covered samples belong to at least one stored support; overlapping samples belong to two or more.\par

\begin{center}
\small\setlength{\tabcolsep}{4pt}
\begin{adjustbox}{max width=\linewidth}
\begin{adjustbox}{max width=\linewidth}
\begin{tabular}{lrrrrr}
\toprule
Dataset & H1 supports & H2 supports & Covered samples & Overlapping samples & Right-censored supports \\
\midrule
digits & 36 & 192 & 800 & 589 & 192 \\
mnist & 37 & 0 & 130 & 28 & 0 \\
fashion\_mnist & 5 & 33 & 269 & 120 & 33 \\
coil20 & 15 & 22 & 536 & 332 & 0 \\
olivetti & 0 & 4 & 32 & 9 & 0 \\
coil100 & 9 & 85 & 884 & 720 & 0 \\
mouse\_scrna & 11 & 0 & 78 & 21 & 0 \\
googlenews & 435 & 0 & 771 & 290 & 0 \\
flow\_cytometry & 5 & 0 & 108 & 40 & 0 \\
intersecting\_spheres & 0 & 3 & 1603 & 906 & 0 \\
klein\_bottle & 2 & 1 & 1618 & 239 & 0 \\
multiple\_spheres & 0 & 2 & 1598 & 0 & 0 \\
nonuniform\_sphere & 0 & 1 & 1623 & 0 & 0 \\
orthogonal\_group\_3 & 2 & 6 & 583 & 228 & 6 \\
projective\_plane & 1 & 1 & 1744 & 249 & 0 \\
s\_curve & 0 & 0 & 0 & 0 & 0 \\
sphere & 0 & 1 & 1578 & 0 & 0 \\
sphere\_and\_filled\_ball & 0 & 2 & 920 & 0 & 0 \\
sphere\_times\_circle & 1 & 12 & 601 & 434 & 12 \\
swiss\_roll & 11 & 1 & 1121 & 605 & 0 \\
torus2 & 2 & 2 & 1619 & 442 & 0 \\
torus3 & 3 & 14 & 757 & 636 & 14 \\
\bottomrule
\end{tabular}
\end{adjustbox}
\end{adjustbox}
\end{center}

The counts summarize stored cross-scale tracks and their supports. For a right-censored record, the reported endpoint is the last scale at which survival was confirmed.\par

For the two visually intersecting shells in \texttt{intersecting\_spheres}, three certified $H_2$ records collectively cover 1,603 original samples, and 906 samples occur in more than one stored preimage. These are representative-dependent, nonexclusive memberships: the two visible geometric shells should not be conflated with the number of tracked quotient classes. Unlike a bare circular outline, a CHARM support record exposes the exact input-row membership and any overlap. For overlapping points, the grids prioritize colors of supports with longer confirmed survival. Ties are resolved by member count, dimension, and ID. The interactive support viewer displays complete memberships and overlaps record by record.\par

\subsection{Future Research Directions}
Future work should test noise and parameter robustness, higher-dimensional discovery recall and certification delay, support localization against known memberships, and search cost against uncompressed Rips complexes. Multiple seeds, module ablations, and comparisons with topology-aware reduction methods would clarify which components account for the observed gains.

\subsection{Result Provenance and Code Correspondence}

All tables and figures use the archived inputs and final two-dimensional coordinates recorded in the frozen HTML reports. Weighted and per-dataset tables come from \texttt{overview.html} and \texttt{detailed\_evaluation.html}, respectively; support statistics aggregate the embedded memberships in \texttt{supports.html}. Figure 1 combines four rows from the two grids in Appendix F.4, ordered as \texttt{intersecting\_spheres}, \texttt{multiple\_spheres}, \texttt{coil20}, and \texttt{coil100}. Its support colors use the memberships summarized in Appendix F.5; the two intersecting geometric shells have three tracked $H_2$ records, whose preimages can overlap. Source HTML cells retain full precision; the manuscript displays six decimal places and preserves original rankings.\par

The core implementation comprises cover proposals and diffusion spectra (\texttt{diffusion\_cover.py}), safe contraction and hop births (\texttt{flat\_cover.py}), quotient towers (\texttt{quotient.py}), homology tracking (\texttt{tracking.py}), complete H2 computation (\texttt{h2.py}), shared geometric optimization (\texttt{geometry\_layout.py}), reverse-scale shared coordinates (\texttt{reverse\_geometry.py}), PH correction (\texttt{persistence\_refinement.py}), and leaf restoration (\texttt{persistence\_connectivity.py}). Evaluation and generation entry points are the shared evaluator (\texttt{coordinate\_evaluation.py}), k15/k50 experimental protocol (\texttt{run\_fair\_scales.py}), and final k50 report generator (\texttt{report\_final\_k50.py}).\par

\clearpage
\section{Complete Per-dataset Evaluation Results}

The tables retain all 22 datasets, 5 methods, and 15 metrics from the archived report. Rows are grouped into neighborhood, graph-geodesic, and Euclidean persistence-diagram metrics. \textbf{Bold} and \underline{underlining} mark first and second place using unrounded values. The listed out-of-view counts affect display only; evaluation uses every sample.

\noindent\begin{minipage}{\linewidth}

\subsection{digits}

1,797 samples; nonsynthetic data. 228 stored support records; 589 samples belong to multiple supports. Points outside the view: CHARM 1, t-SNE 0, UMAP 158, Isomap 62, PCA 6.\par

\begin{center}
\small\setlength{\tabcolsep}{4pt}
\begin{adjustbox}{max width=\linewidth}
\begin{tabular}{@{}lrrrrr@{}}
\toprule
Metric & CHARM & t-SNE & UMAP & Isomap & PCA \\
\midrule
T50 \(\uparrow\) & 0.978686 & \underline{0.979550} & \textbf{0.979580} & 0.851097 & 0.832963 \\
C50 \(\uparrow\) & \underline{0.967094} & \textbf{0.967877} & 0.963183 & 0.952912 & 0.928777 \\
Q50 \(\uparrow\) & \underline{0.655526} & \textbf{0.656628} & 0.636105 & 0.363740 & 0.294213 \\
RNX50 \(\uparrow\) & \underline{0.645661} & \textbf{0.646795} & 0.625684 & 0.345519 & 0.274001 \\
RNX\_AUC\_log \(\uparrow\) & \underline{0.502141} & \textbf{0.514453} & 0.402955 & 0.269891 & 0.233380 \\
RNX\_global \(\uparrow\) & 0.414487 & 0.441766 & 0.372812 & \textbf{0.455923} & \underline{0.450966} \\
\midrule
RTA\_G \(\uparrow\) & 0.731885 & \textbf{0.741165} & 0.704214 & \underline{0.734685} & 0.694634 \\
rho\_G \(\uparrow\) & \underline{0.631348} & \textbf{0.655994} & 0.560977 & 0.629806 & 0.523644 \\
pearson\_G \(\uparrow\) & \underline{0.682793} & \textbf{0.695379} & 0.584063 & 0.657579 & 0.546930 \\
stress\_G \(\downarrow\) & \underline{0.318269} & \textbf{0.316552} & 0.431351 & 0.334114 & 0.366767 \\
RV\_G \(\downarrow\) & \underline{0.533794} & \textbf{0.516449} & 0.658870 & 0.567590 & 0.700867 \\
\midrule
H0\_dB \(\downarrow\) & \textbf{0.415577} & \textbf{0.415577} & \textbf{0.415577} & 0.463171 & 0.463171 \\
H0\_W2 \(\downarrow\) & \textbf{10.651745} & 10.668080 & \underline{10.656811} & 10.700891 & 10.706154 \\
H1\_dB \(\downarrow\) & \textbf{0.125232} & \textbf{0.125232} & \textbf{0.125232} & \textbf{0.125232} & \textbf{0.125232} \\
H1\_W2 \(\downarrow\) & \textbf{1.025675} & 1.036269 & \underline{1.026733} & 1.064655 & 1.066171 \\
\bottomrule
\end{tabular}
\end{adjustbox}
\end{center}

\end{minipage}

\medskip

\noindent\begin{minipage}{\linewidth}

\subsection{mnist}

2,000 samples; nonsynthetic data. 37 stored support records; 28 samples belong to multiple supports. Points outside the view: CHARM 2, t-SNE 0, UMAP 0, Isomap 47, PCA 55.\par

\begin{center}
\small\setlength{\tabcolsep}{4pt}
\begin{adjustbox}{max width=\linewidth}
\begin{tabular}{@{}lrrrrr@{}}
\toprule
Metric & CHARM & t-SNE & UMAP & Isomap & PCA \\
\midrule
T50 \(\uparrow\) & \textbf{0.928197} & 0.924353 & \underline{0.927196} & 0.745046 & 0.734211 \\
C50 \(\uparrow\) & \textbf{0.919998} & 0.900362 & \underline{0.913782} & 0.896210 & 0.872182 \\
Q50 \(\uparrow\) & \textbf{0.497020} & \underline{0.495820} & 0.492720 & 0.194320 & 0.188980 \\
RNX50 \(\uparrow\) & \textbf{0.484116} & \underline{0.482886} & 0.479706 & 0.173651 & 0.168174 \\
RNX\_AUC\_log \(\uparrow\) & \textbf{0.445676} & \underline{0.430496} & 0.325204 & 0.166463 & 0.160316 \\
RNX\_global \(\uparrow\) & \textbf{0.372633} & 0.319324 & 0.328274 & \underline{0.335768} & 0.316183 \\
\midrule
RTA\_G \(\uparrow\) & \textbf{0.715690} & 0.678400 & 0.692190 & \underline{0.709270} & 0.669340 \\
rho\_G \(\uparrow\) & \textbf{0.623689} & 0.522563 & 0.543375 & \underline{0.609588} & 0.500984 \\
pearson\_G \(\uparrow\) & \textbf{0.631045} & 0.531161 & 0.570070 & \underline{0.621366} & 0.518440 \\
stress\_G \(\downarrow\) & \textbf{0.336712} & 0.395931 & 0.367562 & \underline{0.364117} & 0.396012 \\
RV\_G \(\downarrow\) & \textbf{0.601782} & 0.717868 & 0.675020 & \underline{0.613904} & 0.731220 \\
\midrule
H0\_dB \(\downarrow\) & \textbf{0.616366} & \underline{0.628763} & \underline{0.628763} & \underline{0.628763} & \underline{0.628763} \\
H0\_W2 \(\downarrow\) & \textbf{17.043744} & 17.067096 & \underline{17.063535} & 17.067043 & 17.065951 \\
H1\_dB \(\downarrow\) & \textbf{0.087025} & \textbf{0.087025} & \textbf{0.087025} & \textbf{0.087025} & \textbf{0.087025} \\
H1\_W2 \(\downarrow\) & \textbf{1.175869} & 1.192224 & \underline{1.188694} & 1.206256 & 1.207412 \\
\bottomrule
\end{tabular}
\end{adjustbox}
\end{center}

\end{minipage}

\noindent\begin{minipage}{\linewidth}

\subsection{fashion\_mnist}

2,000 samples; nonsynthetic data. 38 stored support records; 120 samples belong to multiple supports. Points outside the view: CHARM 2, t-SNE 0, UMAP 0, Isomap 25, PCA 20.\par

\begin{center}
\small\setlength{\tabcolsep}{4pt}
\begin{adjustbox}{max width=\linewidth}
\begin{tabular}{@{}lrrrrr@{}}
\toprule
Metric & CHARM & t-SNE & UMAP & Isomap & PCA \\
\midrule
T50 \(\uparrow\) & \underline{0.975354} & \textbf{0.975640} & 0.972445 & 0.924270 & 0.916085 \\
C50 \(\uparrow\) & \textbf{0.972760} & \underline{0.971361} & 0.968302 & 0.969815 & 0.964078 \\
Q50 \(\uparrow\) & \underline{0.582080} & \textbf{0.585710} & 0.558440 & 0.375440 & 0.345840 \\
RNX50 \(\uparrow\) & \underline{0.571359} & \textbf{0.575082} & 0.547112 & 0.359417 & 0.329058 \\
RNX\_AUC\_log \(\uparrow\) & \textbf{0.515705} & \underline{0.514218} & 0.415797 & 0.334508 & 0.334773 \\
RNX\_global \(\uparrow\) & 0.625262 & 0.605466 & 0.569151 & \underline{0.654176} & \textbf{0.693491} \\
\midrule
RTA\_G \(\uparrow\) & \underline{0.829880} & 0.822470 & 0.812080 & \textbf{0.859520} & 0.798480 \\
rho\_G \(\uparrow\) & \underline{0.850210} & 0.833092 & 0.792017 & \textbf{0.877125} & 0.777914 \\
pearson\_G \(\uparrow\) & \underline{0.853706} & 0.836692 & 0.786117 & \textbf{0.875588} & 0.773627 \\
stress\_G \(\downarrow\) & \textbf{0.231567} & 0.245469 & 0.318218 & \underline{0.234563} & 0.278353 \\
RV\_G \(\downarrow\) & \underline{0.271187} & 0.299946 & 0.382020 & \textbf{0.233346} & 0.401501 \\
\midrule
H0\_dB \(\downarrow\) & \textbf{0.530902} & 0.565296 & \underline{0.545346} & 0.565296 & 0.565296 \\
H0\_W2 \(\downarrow\) & \textbf{12.688887} & 12.716394 & \underline{12.696900} & 12.720741 & 12.726218 \\
H1\_dB \(\downarrow\) & \textbf{0.067061} & \textbf{0.067061} & \textbf{0.067061} & \textbf{0.067061} & 0.067488 \\
H1\_W2 \(\downarrow\) & \textbf{0.583387} & 0.604143 & \underline{0.589153} & 0.631017 & 0.639998 \\
\bottomrule
\end{tabular}
\end{adjustbox}
\end{center}

\end{minipage}

\medskip

\noindent\begin{minipage}{\linewidth}

\subsection{coil20}

1,440 samples; nonsynthetic data. 37 stored support records; 332 samples belong to multiple supports. Points outside the view: CHARM 2, t-SNE 0, UMAP 216, Isomap 37, PCA 72.\par

\begin{center}
\small\setlength{\tabcolsep}{4pt}
\begin{adjustbox}{max width=\linewidth}
\begin{tabular}{@{}lrrrrr@{}}
\toprule
Metric & CHARM & t-SNE & UMAP & Isomap & PCA \\
\midrule
T50 \(\uparrow\) & 0.967253 & \underline{0.967751} & \textbf{0.974252} & 0.902401 & 0.881750 \\
C50 \(\uparrow\) & \textbf{0.976072} & \underline{0.974116} & 0.967190 & 0.968009 & 0.948878 \\
Q50 \(\uparrow\) & 0.725486 & \textbf{0.735528} & \underline{0.732417} & 0.494500 & 0.443889 \\
RNX50 \(\uparrow\) & 0.715604 & \textbf{0.726008} & \underline{0.722784} & 0.476303 & 0.423870 \\
RNX\_AUC\_log \(\uparrow\) & \underline{0.635695} & \textbf{0.648467} & 0.467285 & 0.394464 & 0.397816 \\
RNX\_global \(\uparrow\) & 0.572998 & 0.546313 & 0.393609 & \underline{0.574453} & \textbf{0.584282} \\
\midrule
RTA\_G \(\uparrow\) & 0.722249 & \underline{0.750350} & \textbf{0.751544} & 0.747860 & 0.685874 \\
rho\_G \(\uparrow\) & 0.632660 & \textbf{0.681076} & 0.675128 & \underline{0.675226} & 0.525357 \\
pearson\_G \(\uparrow\) & 0.642847 & \textbf{0.695511} & 0.647885 & \underline{0.686031} & 0.539859 \\
stress\_G \(\downarrow\) & \underline{0.346325} & \textbf{0.329411} & 0.397964 & 0.362584 & 0.413106 \\
RV\_G \(\downarrow\) & 0.586748 & \textbf{0.516264} & 0.580245 & \underline{0.529362} & 0.708552 \\
\midrule
H0\_dB \(\downarrow\) & \textbf{0.484320} & 0.540468 & \textbf{0.484320} & 0.540468 & 0.540468 \\
H0\_W2 \(\downarrow\) & \textbf{7.277120} & 7.316782 & \underline{7.292898} & 7.324517 & 7.326965 \\
H1\_dB \(\downarrow\) & \textbf{0.271557} & \textbf{0.271557} & \textbf{0.271557} & \textbf{0.271557} & \textbf{0.271557} \\
H1\_W2 \(\downarrow\) & \textbf{0.901798} & 0.912725 & \underline{0.906362} & 0.910421 & 0.912723 \\
\bottomrule
\end{tabular}
\end{adjustbox}
\end{center}

\end{minipage}

\noindent\begin{minipage}{\linewidth}

\subsection{olivetti}

400 samples; nonsynthetic data. 4 stored support records; 9 samples belong to multiple supports. Points outside the view: CHARM 2, t-SNE 0, UMAP 0, Isomap 0, PCA 0.\par

\begin{center}
\small\setlength{\tabcolsep}{4pt}
\begin{adjustbox}{max width=\linewidth}
\begin{tabular}{@{}lrrrrr@{}}
\toprule
Metric & CHARM & t-SNE & UMAP & Isomap & PCA \\
\midrule
T50 \(\uparrow\) & 0.900425 & \underline{0.905299} & \textbf{0.918369} & 0.870514 & 0.869447 \\
C50 \(\uparrow\) & \underline{0.914067} & 0.899718 & 0.909603 & \textbf{0.917845} & 0.914056 \\
Q50 \(\uparrow\) & 0.571050 & \underline{0.586400} & \textbf{0.605250} & 0.506200 & 0.500800 \\
RNX50 \(\uparrow\) & 0.509596 & \underline{0.527145} & \textbf{0.548696} & 0.435455 & 0.429281 \\
RNX\_AUC\_log \(\uparrow\) & \underline{0.627906} & \textbf{0.640421} & 0.498009 & 0.357150 & 0.367021 \\
RNX\_global \(\uparrow\) & \textbf{0.545732} & 0.509996 & 0.520901 & 0.525707 & \underline{0.532476} \\
\midrule
RTA\_G \(\uparrow\) & \textbf{0.788620} & 0.758100 & 0.769410 & \underline{0.784680} & 0.772400 \\
rho\_G \(\uparrow\) & \underline{0.772311} & 0.711194 & 0.729202 & \textbf{0.776262} & 0.740510 \\
pearson\_G \(\uparrow\) & \underline{0.776983} & 0.702440 & 0.733603 & \textbf{0.778775} & 0.749853 \\
stress\_G \(\downarrow\) & \textbf{0.275687} & 0.327819 & 0.321758 & \underline{0.288979} & 0.301557 \\
RV\_G \(\downarrow\) & \underline{0.396297} & 0.506578 & 0.461827 & \textbf{0.393509} & 0.437721 \\
\midrule
H0\_dB \(\downarrow\) & \textbf{0.539860} & \underline{0.591677} & \underline{0.591677} & \underline{0.591677} & \underline{0.591677} \\
H0\_W2 \(\downarrow\) & \textbf{7.390852} & 7.440982 & \underline{7.431255} & 7.439837 & 7.442044 \\
H1\_dB \(\downarrow\) & \textbf{0.082383} & \underline{0.101176} & 0.151404 & 0.118751 & 0.165446 \\
H1\_W2 \(\downarrow\) & \textbf{0.328130} & \underline{0.348344} & 0.370285 & 0.386848 & 0.424361 \\
\bottomrule
\end{tabular}
\end{adjustbox}
\end{center}

\end{minipage}

\medskip

\noindent\begin{minipage}{\linewidth}

\subsection{coil100}

2,000 samples; nonsynthetic data. 94 stored support records; 720 samples belong to multiple supports. Points outside the view: CHARM 2, t-SNE 0, UMAP 92, Isomap 6, PCA 43.\par

\begin{center}
\small\setlength{\tabcolsep}{4pt}
\begin{adjustbox}{max width=\linewidth}
\begin{tabular}{@{}lrrrrr@{}}
\toprule
Metric & CHARM & t-SNE & UMAP & Isomap & PCA \\
\midrule
T50 \(\uparrow\) & \textbf{0.910891} & \underline{0.908805} & 0.897094 & 0.874340 & 0.900080 \\
C50 \(\uparrow\) & \textbf{0.951252} & \underline{0.950340} & 0.930772 & 0.926109 & 0.935401 \\
Q50 \(\uparrow\) & \underline{0.412620} & \textbf{0.413480} & 0.380520 & 0.229730 & 0.308010 \\
RNX50 \(\uparrow\) & \underline{0.397551} & \textbf{0.398433} & 0.364628 & 0.209969 & 0.290258 \\
RNX\_AUC\_log \(\uparrow\) & \underline{0.438767} & \textbf{0.447532} & 0.354441 & 0.309744 & 0.356624 \\
RNX\_global \(\uparrow\) & 0.606342 & 0.599034 & 0.577371 & \underline{0.616748} & \textbf{0.649083} \\
\midrule
RTA\_G \(\uparrow\) & 0.833367 & 0.836940 & 0.839282 & \textbf{0.856901} & \underline{0.840830} \\
rho\_G \(\uparrow\) & 0.853937 & 0.849778 & 0.847915 & \textbf{0.892466} & \underline{0.885324} \\
pearson\_G \(\uparrow\) & 0.854113 & 0.851023 & 0.846125 & \underline{0.898364} & \textbf{0.903347} \\
stress\_G \(\downarrow\) & \textbf{0.253766} & 0.267033 & 0.328266 & \underline{0.261275} & 0.267696 \\
RV\_G \(\downarrow\) & 0.270492 & 0.275761 & 0.284072 & \underline{0.192943} & \textbf{0.183963} \\
\midrule
H0\_dB \(\downarrow\) & \underline{0.821277} & 0.828886 & \textbf{0.818045} & 0.828886 & 0.828886 \\
H0\_W2 \(\downarrow\) & \underline{16.336835} & 16.382195 & \textbf{16.323279} & 16.375240 & 16.370206 \\
H1\_dB \(\downarrow\) & \textbf{0.054245} & 0.079115 & 0.133225 & \textbf{0.054245} & \textbf{0.054245} \\
H1\_W2 \(\downarrow\) & \textbf{0.186628} & 0.235587 & 0.223424 & \underline{0.204097} & 0.239421 \\
\bottomrule
\end{tabular}
\end{adjustbox}
\end{center}

\end{minipage}

\noindent\begin{minipage}{\linewidth}

\subsection{mouse\_scrna}

2,000 samples; nonsynthetic data. 11 stored support records; 21 samples belong to multiple supports. Points outside the view: CHARM 15, t-SNE 0, UMAP 137, Isomap 106, PCA 120.\par

\begin{center}
\small\setlength{\tabcolsep}{4pt}
\begin{adjustbox}{max width=\linewidth}
\begin{tabular}{@{}lrrrrr@{}}
\toprule
Metric & CHARM & t-SNE & UMAP & Isomap & PCA \\
\midrule
T50 \(\uparrow\) & \textbf{0.915550} & \underline{0.910873} & 0.897113 & 0.836908 & 0.748862 \\
C50 \(\uparrow\) & \textbf{0.953100} & \underline{0.951415} & 0.946004 & 0.926432 & 0.852065 \\
Q50 \(\uparrow\) & \textbf{0.498720} & \underline{0.492970} & 0.464460 & 0.265340 & 0.168390 \\
RNX50 \(\uparrow\) & \textbf{0.485860} & \underline{0.479963} & 0.450721 & 0.246493 & 0.147056 \\
RNX\_AUC\_log \(\uparrow\) & \textbf{0.427537} & \underline{0.418595} & 0.337446 & 0.258822 & 0.168041 \\
RNX\_global \(\uparrow\) & \underline{0.493641} & 0.465229 & 0.417789 & \textbf{0.526511} & 0.328534 \\
\midrule
RTA\_G \(\uparrow\) & \underline{0.794200} & 0.765390 & 0.734360 & \textbf{0.809910} & 0.687310 \\
rho\_G \(\uparrow\) & \underline{0.765767} & 0.651800 & 0.609373 & \textbf{0.799499} & 0.687211 \\
pearson\_G \(\uparrow\) & \underline{0.758276} & 0.589439 & 0.548620 & \textbf{0.824419} & 0.732471 \\
stress\_G \(\downarrow\) & \textbf{0.340438} & 0.432420 & 0.468533 & \underline{0.392809} & 0.585716 \\
RV\_G \(\downarrow\) & \underline{0.425018} & 0.652562 & 0.699016 & \textbf{0.320334} & 0.463486 \\
\midrule
H0\_dB \(\downarrow\) & \textbf{1.452301} & 1.775619 & 1.775619 & 1.706292 & \underline{1.618363} \\
H0\_W2 \(\downarrow\) & \textbf{11.446340} & 11.916176 & 11.886872 & 11.770177 & \underline{11.508535} \\
H1\_dB \(\downarrow\) & \textbf{0.086458} & \textbf{0.086458} & \textbf{0.086458} & \textbf{0.086458} & \textbf{0.086458} \\
H1\_W2 \(\downarrow\) & \textbf{0.253292} & 0.297187 & 0.288770 & 0.276368 & \underline{0.270937} \\
\bottomrule
\end{tabular}
\end{adjustbox}
\end{center}

\end{minipage}

\medskip

\noindent\begin{minipage}{\linewidth}

\subsection{googlenews}

2,000 samples; nonsynthetic data. 435 stored support records; 290 samples belong to multiple supports. Points outside the view: CHARM 2, t-SNE 0, UMAP 0, Isomap 0, PCA 105.\par

\begin{center}
\small\setlength{\tabcolsep}{4pt}
\begin{adjustbox}{max width=\linewidth}
\begin{tabular}{@{}lrrrrr@{}}
\toprule
Metric & CHARM & t-SNE & UMAP & Isomap & PCA \\
\midrule
T50 \(\uparrow\) & \textbf{0.827079} & \underline{0.823493} & 0.817953 & 0.682757 & 0.673174 \\
C50 \(\uparrow\) & \underline{0.822757} & 0.819531 & \textbf{0.829700} & 0.808586 & 0.745009 \\
Q50 \(\uparrow\) & \textbf{0.339700} & \underline{0.336290} & 0.330780 & 0.132840 & 0.112770 \\
RNX50 \(\uparrow\) & \textbf{0.322761} & \underline{0.319263} & 0.313612 & 0.110594 & 0.090009 \\
RNX\_AUC\_log \(\uparrow\) & \underline{0.346991} & \textbf{0.354568} & 0.213063 & 0.101574 & 0.086672 \\
RNX\_global \(\uparrow\) & \underline{0.211679} & 0.209382 & 0.208814 & \textbf{0.215528} & 0.190713 \\
\midrule
RTA\_G \(\uparrow\) & 0.653330 & \textbf{0.658520} & 0.655670 & \underline{0.657250} & 0.598230 \\
rho\_G \(\uparrow\) & 0.442089 & \textbf{0.473066} & 0.441947 & \underline{0.445141} & 0.270399 \\
pearson\_G \(\uparrow\) & 0.450087 & \textbf{0.469807} & \underline{0.466111} & 0.449522 & 0.300312 \\
stress\_G \(\downarrow\) & \textbf{0.384142} & 0.397684 & \underline{0.396595} & 0.409235 & 0.476098 \\
RV\_G \(\downarrow\) & 0.797422 & \textbf{0.779281} & \underline{0.782741} & 0.797930 & 0.909813 \\
\midrule
H0\_dB \(\downarrow\) & \textbf{0.637886} & \underline{0.661052} & \underline{0.661052} & \underline{0.661052} & \underline{0.661052} \\
H0\_W2 \(\downarrow\) & \textbf{24.424922} & 24.444178 & \underline{24.439574} & 24.444336 & 24.447686 \\
H1\_dB \(\downarrow\) & \textbf{0.093239} & \textbf{0.093239} & \textbf{0.093239} & \textbf{0.093239} & \textbf{0.093239} \\
H1\_W2 \(\downarrow\) & \textbf{1.496776} & \underline{1.512774} & 1.516727 & 1.521336 & 1.523714 \\
\bottomrule
\end{tabular}
\end{adjustbox}
\end{center}

\end{minipage}

\noindent\begin{minipage}{\linewidth}

\subsection{flow\_cytometry}

2,000 samples; nonsynthetic data. 5 stored support records; 40 samples belong to multiple supports. Points outside the view: CHARM 40, t-SNE 0, UMAP 0, Isomap 20, PCA 47.\par

\begin{center}
\small\setlength{\tabcolsep}{4pt}
\begin{adjustbox}{max width=\linewidth}
\begin{tabular}{@{}lrrrrr@{}}
\toprule
Metric & CHARM & t-SNE & UMAP & Isomap & PCA \\
\midrule
T50 \(\uparrow\) & \underline{0.980326} & \textbf{0.980406} & 0.975946 & 0.914231 & 0.910609 \\
C50 \(\uparrow\) & \textbf{0.976739} & 0.973602 & \underline{0.976579} & 0.960237 & 0.965681 \\
Q50 \(\uparrow\) & \underline{0.611820} & \textbf{0.614450} & 0.577530 & 0.354280 & 0.341600 \\
RNX50 \(\uparrow\) & \underline{0.601862} & \textbf{0.604559} & 0.566692 & 0.337715 & 0.324709 \\
RNX\_AUC\_log \(\uparrow\) & \textbf{0.534850} & \underline{0.523522} & 0.436024 & 0.343573 & 0.352266 \\
RNX\_global \(\uparrow\) & \underline{0.673482} & 0.612301 & 0.652050 & 0.657963 & \textbf{0.713405} \\
\midrule
RTA\_G \(\uparrow\) & \underline{0.841786} & 0.782879 & 0.795296 & \textbf{0.865164} & 0.833962 \\
rho\_G \(\uparrow\) & \underline{0.892497} & 0.769225 & 0.823111 & \textbf{0.923136} & 0.877932 \\
pearson\_G \(\uparrow\) & \underline{0.884387} & 0.760343 & 0.808197 & \textbf{0.911116} & 0.865130 \\
stress\_G \(\downarrow\) & \textbf{0.222205} & 0.299534 & 0.297597 & \underline{0.232750} & 0.277513 \\
RV\_G \(\downarrow\) & \underline{0.217859} & 0.421878 & 0.346817 & \textbf{0.169868} & 0.251550 \\
\midrule
H0\_dB \(\downarrow\) & \textbf{0.336404} & 0.547974 & 0.385378 & 0.547974 & \underline{0.355765} \\
H0\_W2 \(\downarrow\) & \underline{5.493245} & 5.544302 & 5.508501 & 5.504556 & \textbf{5.484245} \\
H1\_dB \(\downarrow\) & \underline{0.091203} & \textbf{0.087086} & \underline{0.091203} & 0.105150 & \underline{0.091203} \\
H1\_W2 \(\downarrow\) & \textbf{0.555042} & 0.568760 & \underline{0.564805} & 0.574903 & 0.578404 \\
\bottomrule
\end{tabular}
\end{adjustbox}
\end{center}

\end{minipage}

\medskip

\noindent\begin{minipage}{\linewidth}

\subsection{intersecting\_spheres}

2,000 samples; synthetic data. 3 stored support records; 906 samples belong to multiple supports. Points outside the view: CHARM 0, t-SNE 0, UMAP 0, Isomap 0, PCA 0.\par

\begin{center}
\small\setlength{\tabcolsep}{4pt}
\begin{adjustbox}{max width=\linewidth}
\begin{tabular}{@{}lrrrrr@{}}
\toprule
Metric & CHARM & t-SNE & UMAP & Isomap & PCA \\
\midrule
T50 \(\uparrow\) & 0.973282 & \textbf{0.991676} & \underline{0.991256} & 0.870513 & 0.869959 \\
C50 \(\uparrow\) & 0.971218 & 0.961812 & 0.963978 & \textbf{0.986837} & \underline{0.986753} \\
Q50 \(\uparrow\) & 0.694940 & \textbf{0.738710} & \underline{0.732880} & 0.479670 & 0.474790 \\
RNX50 \(\uparrow\) & 0.687114 & \textbf{0.732007} & \underline{0.726027} & 0.466321 & 0.461316 \\
RNX\_AUC\_log \(\uparrow\) & 0.597447 & \textbf{0.704599} & \underline{0.628319} & 0.468216 & 0.468540 \\
RNX\_global \(\uparrow\) & 0.555413 & 0.553439 & 0.527845 & \underline{0.593885} & \textbf{0.604197} \\
\midrule
RTA\_G \(\uparrow\) & 0.775690 & 0.765950 & 0.753590 & \textbf{0.809820} & \underline{0.805870} \\
rho\_G \(\uparrow\) & 0.737940 & 0.700962 & 0.665194 & \textbf{0.795679} & \underline{0.793376} \\
pearson\_G \(\uparrow\) & 0.764416 & 0.736623 & 0.669295 & \textbf{0.816366} & \underline{0.814811} \\
stress\_G \(\downarrow\) & 0.287721 & 0.302478 & 0.365683 & \underline{0.256752} & \textbf{0.254207} \\
RV\_G \(\downarrow\) & 0.415669 & 0.457386 & 0.552045 & \textbf{0.333547} & \underline{0.336083} \\
\midrule
H0\_dB \(\downarrow\) & \textbf{0.049392} & \underline{0.069023} & 0.077199 & 0.069698 & 0.072447 \\
H0\_W2 \(\downarrow\) & \underline{1.223435} & 1.453337 & 1.465262 & \textbf{1.211395} & 1.256819 \\
H1\_dB \(\downarrow\) & \textbf{0.058532} & 0.136797 & 0.137703 & \underline{0.059935} & 0.095383 \\
H1\_W2 \(\downarrow\) & \textbf{0.501313} & 0.587299 & 0.581040 & \underline{0.509535} & 0.563493 \\
\bottomrule
\end{tabular}
\end{adjustbox}
\end{center}

\end{minipage}

\noindent\begin{minipage}{\linewidth}

\subsection{klein\_bottle}

2,000 samples; synthetic data. 3 stored support records; 239 samples belong to multiple supports. Points outside the view: CHARM 0, t-SNE 0, UMAP 0, Isomap 0, PCA 0.\par

\begin{center}
\small\setlength{\tabcolsep}{4pt}
\begin{adjustbox}{max width=\linewidth}
\begin{tabular}{@{}lrrrrr@{}}
\toprule
Metric & CHARM & t-SNE & UMAP & Isomap & PCA \\
\midrule
T50 \(\uparrow\) & 0.987248 & \textbf{0.994608} & \underline{0.990597} & 0.958178 & 0.959607 \\
C50 \(\uparrow\) & 0.979601 & 0.976013 & 0.978637 & \underline{0.993309} & \textbf{0.993656} \\
Q50 \(\uparrow\) & \underline{0.736290} & \textbf{0.755180} & 0.727930 & 0.574920 & 0.577940 \\
RNX50 \(\uparrow\) & \underline{0.729525} & \textbf{0.748899} & 0.720950 & 0.564015 & 0.567112 \\
RNX\_AUC\_log \(\uparrow\) & 0.674700 & \textbf{0.762828} & \underline{0.680353} & 0.602163 & 0.604761 \\
RNX\_global \(\uparrow\) & 0.766076 & 0.735204 & 0.738912 & \underline{0.829066} & \textbf{0.849391} \\
\midrule
RTA\_G \(\uparrow\) & 0.882190 & 0.862050 & 0.866000 & \underline{0.915770} & \textbf{0.918070} \\
rho\_G \(\uparrow\) & 0.918109 & 0.886930 & 0.890352 & \underline{0.955313} & \textbf{0.958903} \\
pearson\_G \(\uparrow\) & 0.920830 & 0.888342 & 0.889743 & \underline{0.957213} & \textbf{0.960704} \\
stress\_G \(\downarrow\) & 0.166087 & 0.195624 & 0.194246 & \underline{0.127661} & \textbf{0.121450} \\
RV\_G \(\downarrow\) & 0.152072 & 0.210849 & 0.208358 & \underline{0.083744} & \textbf{0.077049} \\
\midrule
H0\_dB \(\downarrow\) & \textbf{0.046425} & 0.065729 & 0.067914 & 0.059226 & \underline{0.056998} \\
H0\_W2 \(\downarrow\) & \underline{0.989078} & 1.251074 & 1.255383 & \textbf{0.949983} & 1.045846 \\
H1\_dB \(\downarrow\) & \textbf{0.325899} & \textbf{0.325899} & \textbf{0.325899} & \textbf{0.325899} & \textbf{0.325899} \\
H1\_W2 \(\downarrow\) & \underline{0.576713} & 0.593130 & \textbf{0.557692} & 0.577017 & 0.584464 \\
\bottomrule
\end{tabular}
\end{adjustbox}
\end{center}

\end{minipage}

\medskip

\noindent\begin{minipage}{\linewidth}

\subsection{multiple\_spheres}

2,000 samples; synthetic data. 2 stored support records; 0 samples belong to multiple supports. Points outside the view: CHARM 0, t-SNE 0, UMAP 0, Isomap 7, PCA 0.\par

\begin{center}
\small\setlength{\tabcolsep}{4pt}
\begin{adjustbox}{max width=\linewidth}
\begin{tabular}{@{}lrrrrr@{}}
\toprule
Metric & CHARM & t-SNE & UMAP & Isomap & PCA \\
\midrule
T50 \(\uparrow\) & \underline{0.985455} & \textbf{0.996703} & 0.980573 & 0.891863 & 0.927798 \\
C50 \(\uparrow\) & 0.981169 & 0.974418 & \underline{0.985256} & 0.969773 & \textbf{0.991405} \\
Q50 \(\uparrow\) & \underline{0.758180} & \textbf{0.787040} & 0.719780 & 0.386980 & 0.536630 \\
RNX50 \(\uparrow\) & \underline{0.751976} & \textbf{0.781577} & 0.712591 & 0.371253 & 0.524743 \\
RNX\_AUC\_log \(\uparrow\) & \underline{0.674296} & \textbf{0.764330} & 0.660506 & 0.438625 & 0.567248 \\
RNX\_global \(\uparrow\) & 0.722758 & 0.663180 & \underline{0.754389} & 0.692829 & \textbf{0.770441} \\
\midrule
RTA\_G \(\uparrow\) & 0.764739 & 0.755377 & \underline{0.785606} & 0.732071 & \textbf{0.804167} \\
rho\_G \(\uparrow\) & 0.669952 & 0.643922 & \underline{0.701720} & 0.550496 & \textbf{0.754513} \\
pearson\_G \(\uparrow\) & 0.681551 & 0.654961 & \underline{0.709848} & 0.568426 & \textbf{0.755928} \\
stress\_G \(\downarrow\) & 0.328457 & 0.346613 & \underline{0.312737} & 0.433620 & \textbf{0.285389} \\
RV\_G \(\downarrow\) & 0.535488 & 0.571026 & \underline{0.496116} & 0.676892 & \textbf{0.428573} \\
\midrule
H0\_dB \(\downarrow\) & 0.206669 & 0.322157 & 0.510356 & \underline{0.092307} & \textbf{0.032228} \\
H0\_W2 \(\downarrow\) & \textbf{0.610728} & 0.730429 & 0.980156 & 0.769948 & \underline{0.697151} \\
H1\_dB \(\downarrow\) & \textbf{0.026864} & 0.079003 & 0.062991 & 0.062991 & \underline{0.060980} \\
H1\_W2 \(\downarrow\) & \textbf{0.244936} & \underline{0.326692} & 0.345011 & 0.333613 & 0.337293 \\
\bottomrule
\end{tabular}
\end{adjustbox}
\end{center}

\end{minipage}

\noindent\begin{minipage}{\linewidth}

\subsection{nonuniform\_sphere}

2,000 samples; synthetic data. 1 stored support record; 0 samples belong to multiple supports. Points outside the view: CHARM 102, t-SNE 0, UMAP 21, Isomap 210, PCA 185.\par

\begin{center}
\small\setlength{\tabcolsep}{4pt}
\begin{adjustbox}{max width=\linewidth}
\begin{tabular}{@{}lrrrrr@{}}
\toprule
Metric & CHARM & t-SNE & UMAP & Isomap & PCA \\
\midrule
T50 \(\uparrow\) & 0.991813 & \underline{0.992631} & \textbf{0.994930} & 0.857418 & 0.853098 \\
C50 \(\uparrow\) & \textbf{0.982365} & 0.975542 & \underline{0.976135} & 0.972449 & 0.974343 \\
Q50 \(\uparrow\) & \textbf{0.804900} & 0.782370 & \underline{0.798800} & 0.422140 & 0.426390 \\
RNX50 \(\uparrow\) & \textbf{0.799895} & 0.776787 & \underline{0.793638} & 0.407315 & 0.411675 \\
RNX\_AUC\_log \(\uparrow\) & \textbf{0.719949} & \underline{0.718181} & 0.669234 & 0.435142 & 0.437780 \\
RNX\_global \(\uparrow\) & \textbf{0.746667} & 0.569286 & \underline{0.593975} & 0.562064 & 0.565012 \\
\midrule
RTA\_G \(\uparrow\) & \textbf{0.850700} & 0.744250 & 0.716740 & 0.831460 & \underline{0.833210} \\
rho\_G \(\uparrow\) & 0.836316 & 0.597721 & 0.496099 & \underline{0.912113} & \textbf{0.914655} \\
pearson\_G \(\uparrow\) & 0.748078 & 0.526527 & 0.371364 & \textbf{0.951632} & \underline{0.950751} \\
stress\_G \(\downarrow\) & 0.412827 & 0.528621 & 0.595025 & \underline{0.210525} & \textbf{0.200815} \\
RV\_G \(\downarrow\) & 0.440380 & 0.722770 & 0.862089 & \textbf{0.094397} & \underline{0.096072} \\
\midrule
H0\_dB \(\downarrow\) & \underline{0.127078} & 0.161137 & 0.178659 & \textbf{0.069364} & 0.137368 \\
H0\_W2 \(\downarrow\) & 1.482161 & 1.662320 & 1.634134 & \textbf{1.156621} & \underline{1.288280} \\
H1\_dB \(\downarrow\) & 0.150961 & 0.168346 & 0.173758 & \textbf{0.103837} & \underline{0.145474} \\
H1\_W2 \(\downarrow\) & 0.527785 & 0.567853 & 0.605494 & \textbf{0.467918} & \underline{0.521247} \\
\bottomrule
\end{tabular}
\end{adjustbox}
\end{center}

\end{minipage}

\medskip

\noindent\begin{minipage}{\linewidth}

\subsection{orthogonal\_group\_3}

2,000 samples; synthetic data. 8 stored support records; 228 samples belong to multiple supports. Points outside the view: CHARM 0, t-SNE 0, UMAP 0, Isomap 0, PCA 157.\par

\begin{center}
\small\setlength{\tabcolsep}{4pt}
\begin{adjustbox}{max width=\linewidth}
\begin{tabular}{@{}lrrrrr@{}}
\toprule
Metric & CHARM & t-SNE & UMAP & Isomap & PCA \\
\midrule
T50 \(\uparrow\) & \underline{0.947079} & \textbf{0.965584} & 0.930075 & 0.646270 & 0.716387 \\
C50 \(\uparrow\) & \underline{0.930534} & 0.922778 & \textbf{0.945536} & 0.910379 & 0.875715 \\
Q50 \(\uparrow\) & \underline{0.578810} & \textbf{0.603340} & 0.558520 & 0.207370 & 0.217570 \\
RNX50 \(\uparrow\) & \underline{0.568005} & \textbf{0.593164} & 0.547194 & 0.187036 & 0.197497 \\
RNX\_AUC\_log \(\uparrow\) & \underline{0.421346} & \textbf{0.450501} & 0.344657 & 0.105753 & 0.163196 \\
RNX\_global \(\uparrow\) & \underline{0.080778} & 0.052584 & 0.057796 & 0.029366 & \textbf{0.226304} \\
\midrule
RTA\_G \(\uparrow\) & \underline{0.638391} & 0.624095 & \textbf{0.643940} & 0.603551 & 0.637759 \\
rho\_G \(\uparrow\) & \underline{0.396643} & 0.350959 & \textbf{0.404693} & 0.248357 & 0.364689 \\
pearson\_G \(\uparrow\) & \underline{0.417553} & 0.384768 & \textbf{0.433439} & 0.291284 & 0.400568 \\
stress\_G \(\downarrow\) & \underline{0.413036} & 0.416113 & \textbf{0.392150} & 0.516536 & 0.428313 \\
RV\_G \(\downarrow\) & \underline{0.825650} & 0.851954 & \textbf{0.812131} & 0.915153 & 0.839545 \\
\midrule
H0\_dB \(\downarrow\) & \underline{0.577566} & \underline{0.577566} & \textbf{0.317101} & \underline{0.577566} & \underline{0.577566} \\
H0\_W2 \(\downarrow\) & 5.255765 & 5.278999 & \underline{5.249716} & 5.267503 & \textbf{5.235632} \\
H1\_dB \(\downarrow\) & \textbf{0.425496} & \textbf{0.425496} & \textbf{0.425496} & \textbf{0.425496} & \textbf{0.425496} \\
H1\_W2 \(\downarrow\) & \textbf{1.740718} & 1.754778 & \underline{1.742644} & 1.743988 & 1.766222 \\
\bottomrule
\end{tabular}
\end{adjustbox}
\end{center}

\end{minipage}

\noindent\begin{minipage}{\linewidth}

\subsection{projective\_plane}

2,000 samples; synthetic data. 2 stored support records; 249 samples belong to multiple supports. Points outside the view: CHARM 0, t-SNE 0, UMAP 0, Isomap 0, PCA 0.\par

\begin{center}
\small\setlength{\tabcolsep}{4pt}
\begin{adjustbox}{max width=\linewidth}
\begin{tabular}{@{}lrrrrr@{}}
\toprule
Metric & CHARM & t-SNE & UMAP & Isomap & PCA \\
\midrule
T50 \(\uparrow\) & 0.987652 & \underline{0.996476} & \textbf{0.997078} & 0.836224 & 0.835330 \\
C50 \(\uparrow\) & 0.960851 & 0.956712 & 0.957032 & \textbf{0.976098} & \underline{0.975380} \\
Q50 \(\uparrow\) & 0.781750 & \textbf{0.795610} & \underline{0.786180} & 0.431960 & 0.428380 \\
RNX50 \(\uparrow\) & 0.776151 & \textbf{0.790367} & \underline{0.780695} & 0.417387 & 0.413716 \\
RNX\_AUC\_log \(\uparrow\) & 0.600865 & \textbf{0.704412} & \underline{0.653712} & 0.383384 & 0.354556 \\
RNX\_global \(\uparrow\) & \underline{0.581337} & 0.531160 & \textbf{0.585794} & 0.449449 & 0.448880 \\
\midrule
RTA\_G \(\uparrow\) & \textbf{0.717590} & \underline{0.713300} & 0.712530 & 0.709100 & 0.708940 \\
rho\_G \(\uparrow\) & \textbf{0.574428} & 0.536921 & 0.549197 & \underline{0.549627} & 0.547514 \\
pearson\_G \(\uparrow\) & \textbf{0.585913} & 0.569879 & 0.561704 & \underline{0.580661} & 0.578274 \\
stress\_G \(\downarrow\) & \textbf{0.355611} & \underline{0.365508} & 0.368426 & 0.367140 & 0.368839 \\
RV\_G \(\downarrow\) & \textbf{0.656706} & 0.675238 & 0.684489 & \underline{0.662833} & 0.665599 \\
\midrule
H0\_dB \(\downarrow\) & \textbf{0.049486} & 0.097873 & 0.067597 & 0.056841 & \underline{0.053306} \\
H0\_W2 \(\downarrow\) & \textbf{1.366765} & 1.642626 & 1.614349 & \underline{1.511190} & 1.513882 \\
H1\_dB \(\downarrow\) & \textbf{0.670943} & \textbf{0.670943} & \textbf{0.670943} & \textbf{0.670943} & \textbf{0.670943} \\
H1\_W2 \(\downarrow\) & \textbf{0.852165} & 0.917920 & 0.939516 & 0.878072 & \underline{0.874163} \\
\bottomrule
\end{tabular}
\end{adjustbox}
\end{center}

\end{minipage}

\medskip

\noindent\begin{minipage}{\linewidth}

\subsection{s\_curve}

2,000 samples; synthetic data. 0 stored support records; 0 samples belong to multiple supports. Points outside the view: CHARM 0, t-SNE 0, UMAP 0, Isomap 0, PCA 0.\par

\begin{center}
\small\setlength{\tabcolsep}{4pt}
\begin{adjustbox}{max width=\linewidth}
\begin{tabular}{@{}lrrrrr@{}}
\toprule
Metric & CHARM & t-SNE & UMAP & Isomap & PCA \\
\midrule
T50 \(\uparrow\) & \underline{0.999275} & 0.999056 & 0.997350 & \textbf{0.999909} & 0.943687 \\
C50 \(\uparrow\) & \underline{0.999009} & 0.996251 & 0.993858 & \textbf{0.999882} & 0.982859 \\
Q50 \(\uparrow\) & \underline{0.884260} & 0.873280 & 0.797440 & \textbf{0.960450} & 0.381670 \\
RNX50 \(\uparrow\) & \underline{0.881291} & 0.870029 & 0.792243 & \textbf{0.959435} & 0.365807 \\
RNX\_AUC\_log \(\uparrow\) & 0.737991 & \underline{0.819404} & 0.709334 & \textbf{0.880624} & 0.412789 \\
RNX\_global \(\uparrow\) & \underline{0.872858} & 0.840186 & 0.796983 & \textbf{0.892754} & 0.812197 \\
\midrule
RTA\_G \(\uparrow\) & \underline{0.977190} & 0.966440 & 0.948630 & \textbf{0.995260} & 0.851530 \\
rho\_G \(\uparrow\) & \underline{0.995938} & 0.990649 & 0.982164 & \textbf{0.999792} & 0.897762 \\
pearson\_G \(\uparrow\) & \underline{0.995659} & 0.991789 & 0.968673 & \textbf{0.999878} & 0.865716 \\
stress\_G \(\downarrow\) & \underline{0.051429} & 0.069346 & 0.135429 & \textbf{0.008427} & 0.269506 \\
RV\_G \(\downarrow\) & \underline{0.008663} & 0.016354 & 0.061673 & \textbf{0.000245} & 0.250537 \\
\midrule
H0\_dB \(\downarrow\) & \textbf{0.035181} & 0.053771 & \underline{0.047294} & 0.055631 & 0.062313 \\
H0\_W2 \(\downarrow\) & \textbf{0.751004} & 0.906295 & 0.958991 & \underline{0.802724} & 1.005550 \\
H1\_dB \(\downarrow\) & \underline{0.147966} & \underline{0.147966} & \underline{0.147966} & \underline{0.147966} & \textbf{0.048284} \\
H1\_W2 \(\downarrow\) & \underline{0.337815} & 0.349946 & 0.349090 & 0.377053 & \textbf{0.313260} \\
\bottomrule
\end{tabular}
\end{adjustbox}
\end{center}

\end{minipage}

\noindent\begin{minipage}{\linewidth}

\subsection{sphere}

2,000 samples; synthetic data. 1 stored support record; 0 samples belong to multiple supports. Points outside the view: CHARM 0, t-SNE 0, UMAP 0, Isomap 0, PCA 0.\par

\begin{center}
\small\setlength{\tabcolsep}{4pt}
\begin{adjustbox}{max width=\linewidth}
\begin{tabular}{@{}lrrrrr@{}}
\toprule
Metric & CHARM & t-SNE & UMAP & Isomap & PCA \\
\midrule
T50 \(\uparrow\) & 0.968653 & \underline{0.994909} & \textbf{0.996766} & 0.848871 & 0.848136 \\
C50 \(\uparrow\) & 0.978763 & 0.968978 & 0.969467 & \textbf{0.989486} & \underline{0.989315} \\
Q50 \(\uparrow\) & 0.748280 & \textbf{0.790560} & \underline{0.783810} & 0.504120 & 0.500540 \\
RNX50 \(\uparrow\) & 0.741822 & \textbf{0.785187} & \underline{0.778264} & 0.491399 & 0.487727 \\
RNX\_AUC\_log \(\uparrow\) & 0.619476 & \textbf{0.723281} & \underline{0.678413} & 0.484425 & 0.470371 \\
RNX\_global \(\uparrow\) & 0.581493 & 0.564139 & \textbf{0.664933} & 0.593464 & \underline{0.593485} \\
\midrule
RTA\_G \(\uparrow\) & 0.776010 & 0.763600 & 0.782900 & \textbf{0.813150} & \underline{0.813140} \\
rho\_G \(\uparrow\) & 0.684457 & 0.672398 & 0.687002 & \underline{0.772351} & \textbf{0.772720} \\
pearson\_G \(\uparrow\) & 0.695752 & 0.684241 & 0.659896 & \underline{0.772504} & \textbf{0.772748} \\
stress\_G \(\downarrow\) & 0.317802 & 0.330832 & 0.343035 & \underline{0.274503} & \textbf{0.274379} \\
RV\_G \(\downarrow\) & 0.515929 & 0.531815 & 0.564537 & \underline{0.403238} & \textbf{0.402860} \\
\midrule
H0\_dB \(\downarrow\) & \textbf{0.025663} & 0.048753 & 0.046324 & \underline{0.035522} & 0.050676 \\
H0\_W2 \(\downarrow\) & \textbf{0.776054} & 1.119874 & 1.118941 & \underline{0.942005} & 0.949264 \\
H1\_dB \(\downarrow\) & \textbf{0.042550} & 0.189652 & 0.278938 & \underline{0.065405} & 0.065514 \\
H1\_W2 \(\downarrow\) & \textbf{0.329969} & 0.507090 & 0.491609 & 0.426145 & \underline{0.422550} \\
\bottomrule
\end{tabular}
\end{adjustbox}
\end{center}

\end{minipage}

\medskip

\noindent\begin{minipage}{\linewidth}

\subsection{sphere\_and\_filled\_ball}

2,000 samples; synthetic data. 2 stored support records; 0 samples belong to multiple supports. Points outside the view: CHARM 0, t-SNE 0, UMAP 0, Isomap 31, PCA 0.\par

\begin{center}
\small\setlength{\tabcolsep}{4pt}
\begin{adjustbox}{max width=\linewidth}
\begin{tabular}{@{}lrrrrr@{}}
\toprule
Metric & CHARM & t-SNE & UMAP & Isomap & PCA \\
\midrule
T50 \(\uparrow\) & 0.987025 & \underline{0.989662} & \textbf{0.990975} & 0.885297 & 0.926263 \\
C50 \(\uparrow\) & 0.977119 & 0.969141 & \underline{0.982789} & 0.959624 & \textbf{0.988459} \\
Q50 \(\uparrow\) & \textbf{0.720120} & 0.714020 & \underline{0.715670} & 0.364100 & 0.487550 \\
RNX50 \(\uparrow\) & \textbf{0.712940} & 0.706683 & \underline{0.708376} & 0.347787 & 0.474404 \\
RNX\_AUC\_log \(\uparrow\) & \underline{0.651113} & \textbf{0.677757} & 0.646036 & 0.412969 & 0.504269 \\
RNX\_global \(\uparrow\) & 0.743106 & 0.663966 & \textbf{0.773873} & 0.685088 & \underline{0.768092} \\
\midrule
RTA\_G \(\uparrow\) & 0.780742 & 0.744787 & \underline{0.795725} & 0.730905 & \textbf{0.801151} \\
rho\_G \(\uparrow\) & 0.677617 & 0.627195 & \underline{0.725048} & 0.662881 & \textbf{0.773456} \\
pearson\_G \(\uparrow\) & 0.642686 & 0.619570 & 0.664006 & \underline{0.715015} & \textbf{0.764604} \\
stress\_G \(\downarrow\) & 0.355381 & 0.371982 & \underline{0.347577} & 0.418372 & \textbf{0.290528} \\
RV\_G \(\downarrow\) & 0.586955 & 0.616133 & 0.559096 & \underline{0.488753} & \textbf{0.415381} \\
\midrule
H0\_dB \(\downarrow\) & \underline{0.056361} & 0.084964 & 0.601200 & \textbf{0.052086} & 0.056840 \\
H0\_W2 \(\downarrow\) & \textbf{1.034599} & 1.132896 & 1.378640 & 1.144673 & \underline{1.121275} \\
H1\_dB \(\downarrow\) & \textbf{0.038780} & \underline{0.043381} & 0.052586 & 0.051692 & 0.052586 \\
H1\_W2 \(\downarrow\) & \textbf{0.360273} & \underline{0.376427} & 0.406983 & 0.383550 & 0.402409 \\
\bottomrule
\end{tabular}
\end{adjustbox}
\end{center}

\end{minipage}

\noindent\begin{minipage}{\linewidth}

\subsection{sphere\_times\_circle}

2,000 samples; synthetic data. 13 stored support records; 434 samples belong to multiple supports. Points outside the view: CHARM 1, t-SNE 0, UMAP 0, Isomap 0, PCA 0.\par

\begin{center}
\small\setlength{\tabcolsep}{4pt}
\begin{adjustbox}{max width=\linewidth}
\begin{tabular}{@{}lrrrrr@{}}
\toprule
Metric & CHARM & t-SNE & UMAP & Isomap & PCA \\
\midrule
T50 \(\uparrow\) & 0.966890 & \textbf{0.986277} & \underline{0.969434} & 0.841735 & 0.849748 \\
C50 \(\uparrow\) & \textbf{0.938800} & 0.925407 & \underline{0.936437} & 0.927172 & 0.927417 \\
Q50 \(\uparrow\) & \underline{0.567640} & \textbf{0.618750} & 0.565710 & 0.170290 & 0.180710 \\
RNX50 \(\uparrow\) & \underline{0.556548} & \textbf{0.608969} & 0.554569 & 0.149004 & 0.159692 \\
RNX\_AUC\_log \(\uparrow\) & \underline{0.486671} & \textbf{0.561855} & 0.465040 & 0.214214 & 0.215709 \\
RNX\_global \(\uparrow\) & 0.510795 & 0.482140 & \textbf{0.515639} & 0.511852 & \underline{0.512052} \\
\midrule
RTA\_G \(\uparrow\) & 0.747760 & 0.731640 & 0.736580 & \textbf{0.792310} & \underline{0.791180} \\
rho\_G \(\uparrow\) & 0.642615 & 0.605878 & 0.618604 & \textbf{0.779609} & \underline{0.777357} \\
pearson\_G \(\uparrow\) & 0.644859 & 0.615680 & 0.622478 & \textbf{0.738881} & \underline{0.738656} \\
stress\_G \(\downarrow\) & 0.323195 & 0.334540 & 0.334687 & \underline{0.291821} & \textbf{0.291574} \\
RV\_G \(\downarrow\) & 0.584157 & 0.620938 & 0.612521 & \textbf{0.454054} & \underline{0.454388} \\
\midrule
H0\_dB \(\downarrow\) & \textbf{0.127066} & \underline{0.142548} & \underline{0.142548} & \underline{0.142548} & \underline{0.142548} \\
H0\_W2 \(\downarrow\) & \textbf{3.586879} & 3.677134 & \underline{3.660367} & 3.677064 & 3.678821 \\
H1\_dB \(\downarrow\) & 0.506919 & 0.506919 & 0.506919 & \underline{0.412453} & \textbf{0.380436} \\
H1\_W2 \(\downarrow\) & 1.261966 & 1.274079 & 1.306715 & \underline{1.222461} & \textbf{1.213065} \\
\bottomrule
\end{tabular}
\end{adjustbox}
\end{center}

\end{minipage}

\medskip

\noindent\begin{minipage}{\linewidth}

\subsection{swiss\_roll}

2,000 samples; synthetic data. 12 stored support records; 605 samples belong to multiple supports. Points outside the view: CHARM 0, t-SNE 0, UMAP 42, Isomap 0, PCA 38.\par

\begin{center}
\small\setlength{\tabcolsep}{4pt}
\begin{adjustbox}{max width=\linewidth}
\begin{tabular}{@{}lrrrrr@{}}
\toprule
Metric & CHARM & t-SNE & UMAP & Isomap & PCA \\
\midrule
T50 \(\uparrow\) & 0.996442 & \textbf{0.998643} & \underline{0.997849} & 0.858443 & 0.858902 \\
C50 \(\uparrow\) & 0.974426 & \underline{0.989143} & \textbf{0.993552} & 0.974062 & 0.985956 \\
Q50 \(\uparrow\) & \underline{0.841620} & \textbf{0.864310} & 0.826070 & 0.375150 & 0.472100 \\
RNX50 \(\uparrow\) & \underline{0.837557} & \textbf{0.860829} & 0.821608 & 0.359120 & 0.458557 \\
RNX\_AUC\_log \(\uparrow\) & 0.620418 & \textbf{0.717139} & \underline{0.634361} & 0.352001 & 0.468233 \\
RNX\_global \(\uparrow\) & \underline{0.507919} & 0.470376 & 0.476336 & 0.476438 & \textbf{0.571991} \\
\midrule
RTA\_G \(\uparrow\) & 0.660610 & \underline{0.858270} & \textbf{0.959780} & 0.655960 & 0.628340 \\
rho\_G \(\uparrow\) & 0.453508 & \underline{0.847975} & \textbf{0.980543} & 0.393451 & 0.336526 \\
pearson\_G \(\uparrow\) & 0.368641 & \underline{0.772979} & \textbf{0.981050} & 0.219785 & 0.258412 \\
stress\_G \(\downarrow\) & 0.532227 & \underline{0.337054} & \textbf{0.104016} & 0.589523 & 0.567779 \\
RV\_G \(\downarrow\) & 0.864104 & \underline{0.402503} & \textbf{0.037541} & 0.951695 & 0.933223 \\
\midrule
H0\_dB \(\downarrow\) & \textbf{0.052683} & 0.166427 & 0.102740 & 0.102740 & \underline{0.057259} \\
H0\_W2 \(\downarrow\) & \underline{1.127679} & 1.338970 & 1.406484 & 1.385379 & \textbf{1.005743} \\
H1\_dB \(\downarrow\) & \underline{0.256127} & \textbf{0.242701} & \underline{0.256127} & 0.385216 & \underline{0.256127} \\
H1\_W2 \(\downarrow\) & 0.535290 & \underline{0.523076} & 0.540171 & 0.615174 & \textbf{0.470537} \\
\bottomrule
\end{tabular}
\end{adjustbox}
\end{center}

\end{minipage}

\noindent\begin{minipage}{\linewidth}

\subsection{torus2}

2,000 samples; synthetic data. 4 stored support records; 442 samples belong to multiple supports. Points outside the view: CHARM 0, t-SNE 0, UMAP 0, Isomap 0, PCA 0.\par

\begin{center}
\small\setlength{\tabcolsep}{4pt}
\begin{adjustbox}{max width=\linewidth}
\begin{tabular}{@{}lrrrrr@{}}
\toprule
Metric & CHARM & t-SNE & UMAP & Isomap & PCA \\
\midrule
T50 \(\uparrow\) & 0.990242 & \textbf{0.995786} & \underline{0.991170} & 0.973275 & 0.974670 \\
C50 \(\uparrow\) & 0.987002 & 0.982137 & 0.984096 & \underline{0.992663} & \textbf{0.994416} \\
Q50 \(\uparrow\) & \underline{0.740250} & \textbf{0.774970} & 0.721960 & 0.577000 & 0.599420 \\
RNX50 \(\uparrow\) & \underline{0.733586} & \textbf{0.769197} & 0.714827 & 0.566148 & 0.589143 \\
RNX\_AUC\_log \(\uparrow\) & \underline{0.693068} & \textbf{0.786519} & 0.689526 & 0.619543 & 0.633090 \\
RNX\_global \(\uparrow\) & 0.853968 & 0.813602 & 0.823707 & \underline{0.896184} & \textbf{0.912725} \\
\midrule
RTA\_G \(\uparrow\) & 0.913430 & 0.895730 & 0.896750 & \textbf{0.941770} & \underline{0.937890} \\
rho\_G \(\uparrow\) & 0.953438 & 0.932158 & 0.931421 & \textbf{0.978856} & \underline{0.976077} \\
pearson\_G \(\uparrow\) & 0.949364 & 0.927438 & 0.923817 & \underline{0.971120} & \textbf{0.972537} \\
stress\_G \(\downarrow\) & 0.132894 & 0.159064 & 0.163380 & \underline{0.101737} & \textbf{0.098784} \\
RV\_G \(\downarrow\) & 0.098709 & 0.139858 & 0.146562 & \underline{0.056925} & \textbf{0.054172} \\
\midrule
H0\_dB \(\downarrow\) & \textbf{0.033129} & 0.052786 & 0.049327 & \underline{0.037509} & 0.050806 \\
H0\_W2 \(\downarrow\) & \textbf{0.794602} & 1.085908 & 1.107184 & 0.917035 & \underline{0.915246} \\
H1\_dB \(\downarrow\) & \underline{0.231801} & \textbf{0.200047} & \underline{0.231801} & \underline{0.231801} & \underline{0.231801} \\
H1\_W2 \(\downarrow\) & \underline{0.475127} & \textbf{0.467470} & 0.484375 & 0.529216 & 0.481292 \\
\bottomrule
\end{tabular}
\end{adjustbox}
\end{center}

\end{minipage}

\medskip

\noindent\begin{minipage}{\linewidth}

\subsection{torus3}

2,000 samples; synthetic data. 17 stored support records; 636 samples belong to multiple supports. Points outside the view: CHARM 0, t-SNE 0, UMAP 0, Isomap 12, PCA 11.\par

\begin{center}
\small\setlength{\tabcolsep}{4pt}
\begin{adjustbox}{max width=\linewidth}
\begin{tabular}{@{}lrrrrr@{}}
\toprule
Metric & CHARM & t-SNE & UMAP & Isomap & PCA \\
\midrule
T50 \(\uparrow\) & 0.934209 & \textbf{0.961190} & \underline{0.959947} & 0.732480 & 0.732319 \\
C50 \(\uparrow\) & 0.885431 & 0.877594 & 0.890388 & \textbf{0.918380} & \underline{0.917974} \\
Q50 \(\uparrow\) & 0.520380 & \textbf{0.565510} & \underline{0.535610} & 0.235350 & 0.234160 \\
RNX50 \(\uparrow\) & 0.508076 & \textbf{0.554364} & \underline{0.523696} & 0.215734 & 0.214513 \\
RNX\_AUC\_log \(\uparrow\) & \underline{0.448087} & \textbf{0.516141} & 0.414444 & 0.192749 & 0.191025 \\
RNX\_global \(\uparrow\) & \underline{0.330930} & 0.326512 & \textbf{0.382191} & 0.303928 & 0.302212 \\
\midrule
RTA\_G \(\uparrow\) & 0.656780 & 0.648780 & 0.651490 & \textbf{0.660910} & \underline{0.660470} \\
rho\_G \(\uparrow\) & \textbf{0.441755} & 0.417746 & 0.416887 & \underline{0.436227} & 0.432924 \\
pearson\_G \(\uparrow\) & \textbf{0.476176} & 0.447698 & 0.432237 & \underline{0.472138} & 0.469094 \\
stress\_G \(\downarrow\) & \textbf{0.377977} & 0.388101 & 0.392997 & \underline{0.388053} & 0.389768 \\
RV\_G \(\downarrow\) & \textbf{0.773256} & 0.799567 & 0.813171 & \underline{0.777085} & 0.779951 \\
\midrule
H0\_dB \(\downarrow\) & \textbf{0.163471} & \underline{0.178072} & \underline{0.178072} & \underline{0.178072} & \underline{0.178072} \\
H0\_W2 \(\downarrow\) & \textbf{4.263013} & 4.322063 & 4.308095 & 4.274048 & \underline{4.269761} \\
H1\_dB \(\downarrow\) & \textbf{0.373667} & \textbf{0.373667} & \textbf{0.373667} & \textbf{0.373667} & \textbf{0.373667} \\
H1\_W2 \(\downarrow\) & \textbf{1.544095} & \underline{1.567234} & 1.579339 & 1.570798 & 1.573161 \\
\bottomrule
\end{tabular}
\end{adjustbox}
\end{center}

\end{minipage}

\section{Additional Main-text Context and Discussion}
\subsection{Full Introduction and Contributions}

Dimensionality reduction (DR) is widely used to explore image collections, single-cell measurements, word embeddings, and other high-dimensional data, but a visually coherent two-dimensional layout does not by itself certify the topology of the input. Common objectives emphasize neighborhood similarities or pairwise distances \citep{ref12,ref13,ref14}; geometric quality alone leaves open which structures are nontrivial in homology and which samples support them. Crowding and the limits of planar representation further complicate this interpretation, while directly computing higher-order homology incurs the cost of growing simplicial complexes \citep{ref5,ref6}. These limitations motivate a method that combines geometric exploration, tractable homology certification, and sample-level localization. We focus on $H_k$, $k\ge1$, including one-dimensional loops ($H_1$) and void structures represented by closed two-dimensional surfaces ($H_2$).\par

\textbf{Neighborhood optimization and global structure.} t-SNE organizes points by matching neighborhood probabilities in high and low dimensions \citep{ref12}; initialization and parameter choices substantially influence its global layout \citep{ref17,ref18}. Damrich et al. explain t-SNE and UMAP within a contrastive-learning framework, relating negative sampling and normalization to the attraction-repulsion balance, cluster compactness, and local and global structure \citep{ref23}. Several methods introduce complementary geometric objectives: PCA models linear variance \citep{ref19}, Isomap preserves graph-geodesic distances \citep{ref14}, UMAP combines a fuzzy neighborhood graph with spectral initialization \citep{ref13}, and GLoMAP estimates local-global distances \citep{ref24}. To identify homological dimensions and localize the samples supporting them, a method must also use the boundary relations among simplices.\par

\textbf{From crowding to topological representation.} The crowding problem arises when a low-dimensional space must accommodate high-dimensional neighborhoods; t-SNE addresses this geometric difficulty with a heavy-tailed distribution \citep{ref12}. Topological representation introduces a further distinction between intrinsic dimension and embedding dimension. The sphere $S^2$ has intrinsic dimension two and $H_2(S^2;\mathbb F_2)\cong\mathbb F_2$, but a topology-preserving Euclidean embedding requires at least three dimensions. By invariance of domain, no continuous injection $S^2\to\mathbb R^2$ exists \citep[Theorem~2B.3 and Corollary~2B.4]{ref25}. Optimizing neighborhood KL divergence or fuzzy-edge cross-entropy controls similarities expressed by the coordinates \citep{ref12,ref13,ref23}; homological evidence supplies the structural information needed to interpret those coordinates. For example, a sphere and a solid ball can both project orthogonally onto a disk, although their $H_2$ groups are nonzero and trivial, respectively. Combining point locations, dimension labels, and representative-chain supports allows a planar display to convey the shell's closed surface and enclosed void.\par

\textbf{Homology preservation and structural localization.} Topology-aware dimensionality reduction integrates structure into geometric modeling in several ways. Topological Autoencoders incorporate topological signatures into representation learning \citep{ref15}; manifold landmarking and tearing organize low-dimensional representations around loops \citep{ref27}; and AdaHIsomap uses persistence diagrams to guide Mapper cover refinement and landmark selection \citep{ref30}. We make homological evidence an explicit part of the two-dimensional output. Dimension labels identify structural types, representative chains and original-sample memberships specify their locations, and cross-scale records describe their persistence. Figure 1 illustrates this combination: colored supports with $H_2$ labels identify structures in separate and intersecting spherical shells, while colored members in the COIL datasets locate certified structures within the image collections.\par

\textbf{Computing higher-order homology at scale.} TopoMap \citep{ref16} preserves $H_0$ between the original and two-dimensional Rips filtrations by organizing connectivity events through a minimum spanning tree. TopoMap++ improves the layout and efficiency of these projections \citep{ref21}. Computing $H_1,H_2$, and higher-dimensional homology requires higher-order simplices and their boundary matrices. A clique on $m$ vertices contains $2^m-1$ nonempty simplices. Computation through $H_k$ requires simplices through dimension $(k+1)$, giving a worst-case candidate count of $\sum_{j=1}^{k+2}\binom{m}{j}$ \citep{ref5,ref6}. This count is a high-degree polynomial in $m$ for fixed $k$ and can become exponential as the dimension grows. Complex compression, exact certification of representative chains, and reuse across scales therefore form the computational foundation of our approach.\par

We propose \textbf{CHARM, a two-dimensional reduction method with certified homology supports}. Its four stages connect geometric modeling, structural computation, and visualization. First, diffusion coordinates smooth the geometry used to propose covers, following the multistep propagation of Diffusion Maps \citep{ref2,ref22} and the landmark diffusion perspective of Roseland \citep{ref26}. Graph-based safety tests then determine which contractions to accept. Second, a coarse graph and a Cover Quotient Tower \citep{ref1} compress multiscale computation, building on the smaller computational domains used by witness complexes \citep{ref9}, graph-induced complexes \citep{ref28}, and Sparse Rips \citep{ref6}. Interleaving bounds relate the tower to the coarse graph and original Rips filtration. Third, representative chains provide localizable supports \citep{ref29}. Exact algebra on complete incidence blocks certifies these chains, tests their independence, and tracks them along tower maps. Fourth, original-sample memberships and reverse-scale layout place the resulting records in shared two-dimensional coordinates. Across 22 datasets, this joint output provides 957 support records; on nonsynthetic data, CHARM also ranks among the top two on all 15 sample-weighted evaluation metrics.\par

Our contributions are as follows:\par

1. \textbf{Diffusion-guided proposals and traceable compression.} We combine diffusion spectral smoothing, safe graph contraction, and a Cover Quotient Tower. Diffusion geometry guides cover proposals, edgewise safety tests preserve the base homotopy type, and membership maps connect compressed structures to original samples.\par

2. \textbf{Error guarantees from original Rips to the quotient tower.} We establish a 3-interleaving with the coarse-graph hop filtration. A bound on distortion between original Euclidean distances and coarse distances yields additive interleavings, multiplicative $c$-interleavings on positive scales, and persistence-diagram error bounds.\par

3. \textbf{Certified homology with sample-level localization.} Exact algebra on complete incidence blocks certifies representatives and supports cross-scale tracking. Membership maps and reverse-scale layout attach homological dimensions, supports, and survival states to shared two-dimensional coordinates.\par

4. \textbf{Geometric quality and interpretable structural output.} A comparison with four common reduction methods on 22 datasets evaluates 15 metrics alongside support records. CHARM achieves top-two results on all 15 sample-weighted metrics for nonsynthetic data, while spherical-shell and image examples show how the same view conveys geometry, homological dimension, and sample location.\par

\subsection{Extended Related Work}

\textbf{Geometry- and neighborhood-driven reduction.} t-SNE uses a Student-$t$ distribution to alleviate crowding \citep{ref12}, and UMAP optimizes edge relations in the 1-skeleton of a fuzzy simplicial set \citep[Sections~2--3]{ref13}. Contrastive-learning analysis \citep{ref23} clarifies their connections through negative sampling, normalization, and attraction-repulsion mechanisms. PCA's linear reconstruction \citep{ref19}, Isomap's graph-geodesic distances \citep{ref14}, and GLoMAP's local-global distance estimates \citep{ref24} provide different objectives for global geometry. CHARM combines neighborhood probabilities, distance relations, and multiscale skeleton constraints in a shared layout, complemented by homology records specifying structural types and locations.\par

\textbf{Topology-aware representation learning and visualization.} Topological Autoencoders \citep{ref15} construct differentiable losses from topological signatures of the input and latent spaces. TopoMap \citep{ref16} preserves $H_0$ persistence diagrams and connected-component correspondences; TopoMap++ \citep{ref21} improves computational and spatial efficiency. Yan et al. \citep{ref27} use manifold landmarking and tearing to handle loops. AdaMapper/AdaHIsomap \citep{ref30} refine covers and select homology-informed landmarks using persistence diagrams, adding random anchors to balance distance and homology preservation. CHARM organizes $H_1,H_2$ and requested higher-dimensional structures into records with representative chains and membership maps, \textbf{incorporating higher-order homological evidence directly into two-dimensional displays}.\par

\textbf{Diffusion geometry and landmark compression.} Diffusion Maps \citep{ref2,ref22} convert multistep transition probabilities into spectral coordinates. Roseland \citep{ref26} organizes diffusion through landmarks and studies robustness and scalability. Witness complexes \citep{ref9} and graph-induced complexes \citep{ref28} organize topological computation on smaller vertex sets; the latter induce a complex on subsamples using the original sample graph. These studies motivate connecting robust geometry with compressed representations: diffusion coordinates propose covers, safe contractions preserve the base homotopy type, and membership maps retain the route from coarse structures to original samples.\par

\textbf{Persistent homology, representative chains, and quotient towers.} Persistent homology \citep{ref3,ref4} describes structural evolution across scales. Ripser \citep{ref5} and Sparse Rips \citep{ref6} reduce computational demands through algebraic implementation and complex approximation, respectively. Čufar and Virk \citep{ref29} use persistent cohomology to accelerate representative-chain extraction, highlighting the role of representatives in localizing holes. Cover Quotient \citep{ref1} constructs towers from nested clique partitions and gives a multiplicative 3-approximation relative to the input metric; related methods \citep{ref7,ref8} support simplicial towers containing contractions. CHARM combines this framework with complete incidence-block certification and chain tracking, explicitly establishing the maps induced by coarse projection and within-block representatives. Comparison with the original Euclidean Rips filtration uses distance distortion, consistent with the correspondence-based approach to stability of geometric complexes \citep{ref20}.\par

\subsection{Extended Interpretation and Limitations}

\textbf{Joint performance on nonsynthetic data.} Tables 1--2 show that CHARM's clearest advantage is the combination of geometric fidelity and structural information on the tested nonsynthetic datasets. Across the 15 sample-weighted metrics in Appendix F.3, CHARM places first seven times and second eight times. It improves on UMAP and PCA on every one of these aggregate metrics, and on t-SNE on 12 of 15. Across all 22 datasets, it retains the best continuity and lowest stress, H0\_dB, H0\_W2, and H1\_W2, with top-two rankings on 13 of 15 metrics. These criteria cover neighborhood preservation, distance relationships, and persistence-diagram agreement, so the result supports a broad balance across evaluation objectives. This balance is useful for exploratory analysis of image collections, single-cell measurements, word embeddings, and cytometry data, where analysts may need several views of structure within one representation. The comparisons describe the full pipeline under the fixed configuration of Section 4.1.\par

\textbf{Local fidelity with global distance control.} On nonsynthetic data, CHARM attains the best trustworthiness and continuity, addressing both spurious embedded neighbors and original neighbors that become separated. Its neighborhood overlap and log-weighted multiscale score are within 0.0011 and 0.0003, respectively, of t-SNE's best values. At this closely matched level of neighborhood preservation, CHARM reduces graph-geodesic stress by 10.1\% relative to t-SNE, 19.1\% relative to UMAP, 6.1\% relative to Isomap, and 20.5\% relative to PCA. Its second-place triplet accuracy further indicates that relative distance orderings remain competitive. Thus the benefit includes both local organization and control of distance distortion in the same coordinates. The geodesic scores use an independently constructed 15-neighbor graph, whereas CHARM uses 50 neighbors for training, providing an additional evaluation of geometry at a different graph scale.\par

\textbf{Consistency across data modalities.} Appendix G shows that the aggregate stress advantage extends to seven of the nine nonsynthetic datasets. Relative to t-SNE, stress decreases by 15.0\% on MNIST, 21.3\% on mouse single-cell data, and 25.8\% on flow cytometry. CHARM also attains the highest neighborhood overlap on MNIST, mouse single-cell data, and Google News embeddings. On the latter two datasets, its Q50 values are 0.498720 and 0.339700, compared with 0.168390 and 0.112770 for PCA, respectively. These examples connect the group-level result to both biological measurements and language representations, in addition to images. They support the practical value of preserving local relationships in data whose organization is poorly captured by a single linear projection, while controlling broader distance distortion.\par

\textbf{Agreement across topological discrepancy measures.} Table 2 adds a complementary advantage: CHARM achieves the smallest nonsynthetic aggregate values for all four $H_0,H_1$ diagram distances. Against the best baseline for each metric, the reductions are approximately 6.4\% for H0\_dB, 0.25\% for H0\_W2, 2.1\% for H1\_dB, and 2.2\% for H1\_W2. The per-dataset results strengthen this picture: H1\_W2 is smallest on all nine nonsynthetic datasets, and H0\_W2 is smallest on seven. For example, H1\_W2 decreases by 14.8\% on mouse single-cell data and 20.8\% on COIL-100 relative to t-SNE. Bottleneck and Wasserstein distances emphasize the largest matching deviation and the cumulative matching discrepancy, respectively. Their agreement on the nonsynthetic aggregates therefore supports preservation of both aspects of the input diagrams. These are results for the final coordinates, including PH correction against the input diagrams.\par

\textbf{Connecting visual organization to quantitative evidence.} The two-dimensional comparisons help locate the geometric effects summarized by the tables. In Figure 1, the COIL-20 layout resolves numerous curved groups, with Q50 of 0.725486 compared with 0.494500 for Isomap and 0.443889 for PCA. CHARM also achieves the highest continuity on this dataset, supporting the interpretation that the displayed organization retains many original neighborhood relationships. In Figure F.1, mouse single-cell data occupy an extended configuration with several branches under CHARM, while PCA concentrates many samples in a narrow region; the corresponding Q50 and stress results favor CHARM. Google News provides a different case: the point clouds appear broadly continuous, yet the numerical comparisons favor CHARM's neighborhood overlap and stress. Thus useful preservation can appear as either clearer local groups or a more faithful continuous organization. The plots and metrics together offer evidence beyond visual separation alone.\par

\textbf{Locating structure that coordinates alone cannot specify.} The support overlays provide information additional to geometric quality. The first two columns of Figure 1 use identical CHARM coordinates, making the contribution of the annotations directly visible. In \texttt{multiple\_spheres}, two colored $H_2$ supports identify 796 and 802 original samples with no shared members. In \texttt{intersecting\_spheres}, three $H_2$ records cover 1,603 samples, of which 906 belong to multiple supports (Appendix F.5). These memberships distinguish separated and overlapping structures within the same display. The spherical-shell example further identifies 1,578 samples supporting a certified $H_2$ class. A planar point cloud can suggest groups, gaps, or curves; the dimension labels, representative chains, and membership maps supply the algebraic type and participating samples. In particular, higher-order information is carried by certified records on the quotient tower, complementing the $H_0,H_1$ coordinate-level comparisons in Table 2.\par

\textbf{Structural detail within real-data layouts.} COIL-20 and COIL-100 illustrate the same benefit in image collections: their 37 and 94 stored cross-scale records cover 536 and 884 samples, respectively. Of these, 332 and 720 participate in multiple supports. Selecting a record in the support viewer exposes its original members and confirmed survival information, allowing the analyst to inspect how different structures share samples. Across the benchmark, 957 records provide this route from compressed algebraic computation to input rows. The counts describe stored tracks across scales; the useful output is the combination of each representative, its dimension, its membership, and its survival state. This makes the display suitable for investigating overlapping structures and selecting sample sets for subsequent domain analysis.\par

\textbf{Interpreting the synthetic comparisons.} Controlled synthetic data help separate the objectives that coincide more strongly on the nonsynthetic benchmark. For instance, PCA attains lower graph-geodesic stress on \texttt{sphere} (0.274379 versus 0.317802 for CHARM), while CHARM retains greater neighborhood overlap (0.748280 versus 0.500540). A linear projection can therefore capture broad distance variation effectively on this regular configuration while a neighborhood-oriented representation retains more local relationships. Across all synthetic datasets, CHARM ranks second in Q50 and RNX\_AUC\_log and minimizes both Wasserstein distances. Its H0\_W2 and H1\_W2 improve on PCA by 3.0\% and 2.5\%, even though PCA attains the smaller bottleneck distances. This combination is consistent with stronger cumulative diagram agreement alongside a different worst-case matching error. Synthetic samples constitute 62.4\% of the pooled benchmark, making the separate nonsynthetic results especially informative for the intended applications. Taken together, the comparisons support CHARM's particular value for the tested real-world distributions, where its joint geometric and structural output is accompanied by stronger overall metric rankings.\par

\textbf{Interpretable compression and scale selection.} The theoretical analysis explains how structural information remains traceable after reducing the computational domain. Safe contractions preserve the base homotopy type; complete incidence blocks support exact algebraic decisions; and representative tracking carries certificates across scales. Coarse graphs and quotient towers aggregate vertices, while distance-distortion bounds connect coarse metrics and persistence intervals to the original distances under the stated assumptions. The ratio $\varepsilon/r_{\min}$ expresses compression error relative to the analysis scale, giving a quantitative basis for selecting scales and interpreting sufficiently persistent structures. Membership maps and reverse-scale layout then connect those structural records to the samples shown in two dimensions. Together, these components make compression, certification, and visual inspection parts of one traceable analysis.\par

\textbf{Further research.} Future work will examine the effects of diffusion parameters and noise strength on localization, analyze homology recovery under sampling models \citep{ref10}, and extend higher-dimensional discovery, cross-field analysis, and structural computational efficiency. Appendix F.6 details these directions.\par

\section{Runtime Results and Measurement Caveats}
\label{app:runtime}
The supplied runtime summary gives the following arithmetic means across 22 datasets. It reports 95\% Student-$t$ confidence intervals across datasets but does not establish identical execution conditions for all baselines. Thus these values are descriptive; a statistically controlled speed comparison would require matched hardware, thread counts, preprocessing inclusions, and timing procedures.
\begin{center}
\small\setlength{\tabcolsep}{6pt}
\begin{adjustbox}{max width=\linewidth}
\begin{tabular}{lr}
\toprule
Method & Mean elapsed time (s)\\
\midrule
CHARM & 17.96\\
t-SNE & 15.94\\
UMAP & 16.27\\
Isomap & 8.17\\
PCA & 1.07\\
\bottomrule
\end{tabular}
\end{adjustbox}
\end{center}